\documentclass[review,authoryear]{elsarticle}
\usepackage{url,hyperref,lineno,microtype,subcaption}
\usepackage[onehalfspacing]{setspace}
\usepackage[thinc]{esdiff} 
\usepackage{amssymb}
\usepackage{amsmath}
\usepackage{bm}
\usepackage{tikz}
\usepackage{geometry}
\usepackage[ruled,vlined]{algorithm2e}
\usepackage{algorithmic}
\usepackage{booktabs} 
\hypersetup{pdfauthor={Name}}
\definecolor{green}{rgb}{0,0.6,0}
\newcommand{\norm}[1]{\left\lVert#1\right\rVert} 
\newcommand{\vect}[1]{\bm{#1}} 
\newcommand{\blackline}{\raisebox{0.8pt}{\textbf{---}}}
\newcommand{\greenline}{\raisebox{0.8pt}{\textcolor{green}{\textbf{---}}}}
\newcommand{\redline}{\raisebox{0.8pt}{\textcolor{red}{\textbf{---}}}}
\newcommand{\blueline}{\raisebox{0.8pt}{\textcolor{blue}{\textbf{---}}}}
\usepackage{siunitx}
\DeclareMathOperator{\sech}{sech}
\modulolinenumbers[5]

\journal{Neural Networks}
    \newgeometry{vmargin={20mm}, hmargin={20mm,20mm}}   

\usepackage{numcompress}

\begin{document}

\begin{frontmatter}
\title{Deep neural network enabled corrective source term approach to hybrid analysis and modeling} 



\author[sindresaddress]{Sindre Stenen Blakseth}
\ead{sindre.blakseth@sintef.no}

\author[adilsaddress,adiltrondaddress]{Adil Rasheed\corref{mycorrespondingauthor}}
\cortext[mycorrespondingauthor]{Adil Rasheed}
\ead{adil.rasheed@ntnu.no}

\author[trondsaddress,adiltrondaddress]{Trond Kvamsdal}
\ead{trond.kvamsdal@ntnu.no}

\author[omersaddress]{Omer San}
\ead{osan@okstate.edu }

\address[sindresaddress]{Department of Gas Technology, SINTEF Energy Research)}
\address[adilsaddress]{Department of Engineering Cybernetics, Norwegian University of Science and Technology}
\address[trondsaddress]{Department of Mathematical Sciences, Norwegian University of Science and Technology}
\address[adiltrondaddress]{Mathematics and Cybernetics, SINTEF Digital}
\address[omersaddress]{School of Mechanical and Aerospace Engineering, Oklahoma State University}

\begin{abstract}
In this work, we introduce, justify and demonstrate the Corrective Source Term Approach (CoSTA) -- a novel approach to Hybrid Analysis and Modeling (HAM). The objective of HAM is to combine physics-based modeling (PBM) and data-driven modeling (DDM) to create generalizable, trustworthy, accurate, computationally efficient and self-evolving models. CoSTA achieves this objective by augmenting the governing equation of a PBM model with a corrective source term generated using a deep neural network. In a series of numerical experiments on one-dimensional heat diffusion, CoSTA is found to outperform comparable DDM and PBM models in terms of accuracy -- often reducing predictive errors by several orders of magnitude -- while also generalizing better than pure DDM. Due to its flexible but solid theoretical foundation, CoSTA provides a modular framework for leveraging novel developments within both PBM and DDM. Its theoretical foundation also ensures that CoSTA can be used to model any system governed by (deterministic) partial differential equations. Moreover, CoSTA facilitates interpretation of the DNN-generated source term within the context of PBM, which results in improved explainability of the DNN. These factors make CoSTA a potential door-opener for data-driven techniques to enter high-stakes applications previously reserved for pure PBM.
\end{abstract}

\end{frontmatter}

\section{Introduction}
\label{sec:introduction}
The recent wave of digitalization has given a push to emerging technologies like digital twins. A digital twin is defined as a virtual representation of a physical asset enabled through data and simulators for real-time prediction, optimization, monitoring, controlling, and improved decision making \citep{rasheed2020dtv}. Paramount to digital twins' success is the level of physical realism that can be instilled into them. In this regard, as noticed by \cite{rasheed2020dtv}, modeling plays an important role. The exact requirements of the modeling depend on the digital twin's capability level, which is generally categorized on a scale from 0 to 5 (0-standalone, 1-descriptive, 2-diagnostic, 3-predictive, 4-prescriptive, 5-autonomous) (see Figure~\ref{fig:scales}). 

Although a digital twin offers huge potential in many industries, adaptation of the digital twin technology has been stagnated since its inception, mainly due to the lack of methodological works. To leverage asset-twin technologies, \cite{kapteyn2020probabilistic} proposed a unifying mathematical foundation that draws from probabilistic graphical models and dynamical system theory. The role of surrogate models in the development of digital twin has also been emphasized by \cite{hartmann2018model} and \cite{chakraborty2021role}. There is, however, more work to be done to bring physical realism into digital twins, as many industrial and scientific applications steadily migrate
from sparse data to big data regimes. With this in mind, \cite{san2021hybrid} identified that there are at least four modeling  characteristics of utmost importance; \textit{generalizability, trustworthiness, computational efficiency and accuracy,} and \textit{self-adaptation}. A model's generalizability refers to its ability to solve a wide variety of problems without any problem-specific fine-tuning. Trustworthiness refers to the extent to which a model is explainable, while computational efficiency and accuracy refers to the model's ability to make real-time predictions that match ground truth as closely as possible. Lastly, a model is self-adapting if it can learn and evolve when new situations are encountered. Until recently, most modeling approaches could be categorized as either physics-based modeling (PBM) or data-driven modeling (DDM). These categories are briefly explained below:

\begin{figure*}
	\centering
	\includegraphics[width=\textwidth]{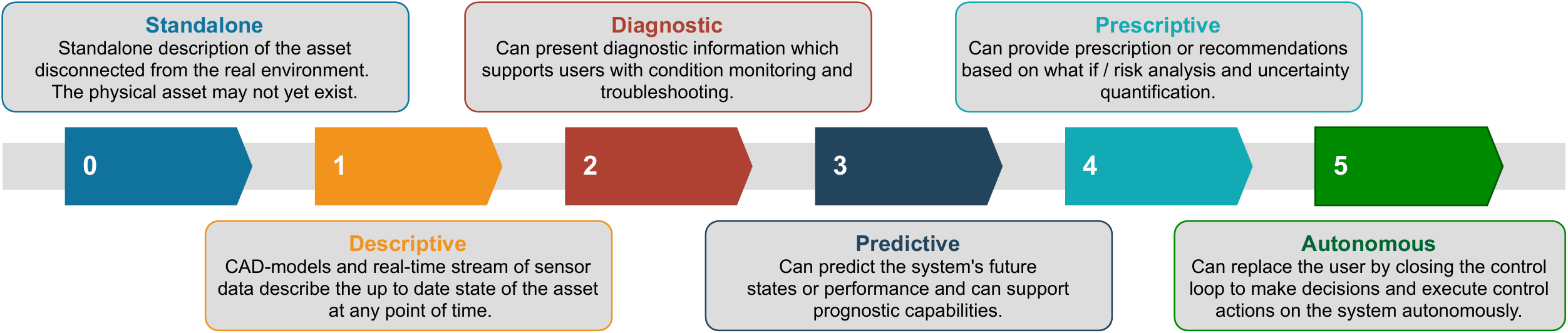}
	\caption{The capability levels of DTs on a scale from 0 to 5. Adapted from \cite{san2021hybrid}.} 
	\label{fig:scales}
\end{figure*}

\begin{figure}[tbh]
	\begin{subfigure}{\linewidth}
		\centering 
		\includegraphics[width=\textwidth]{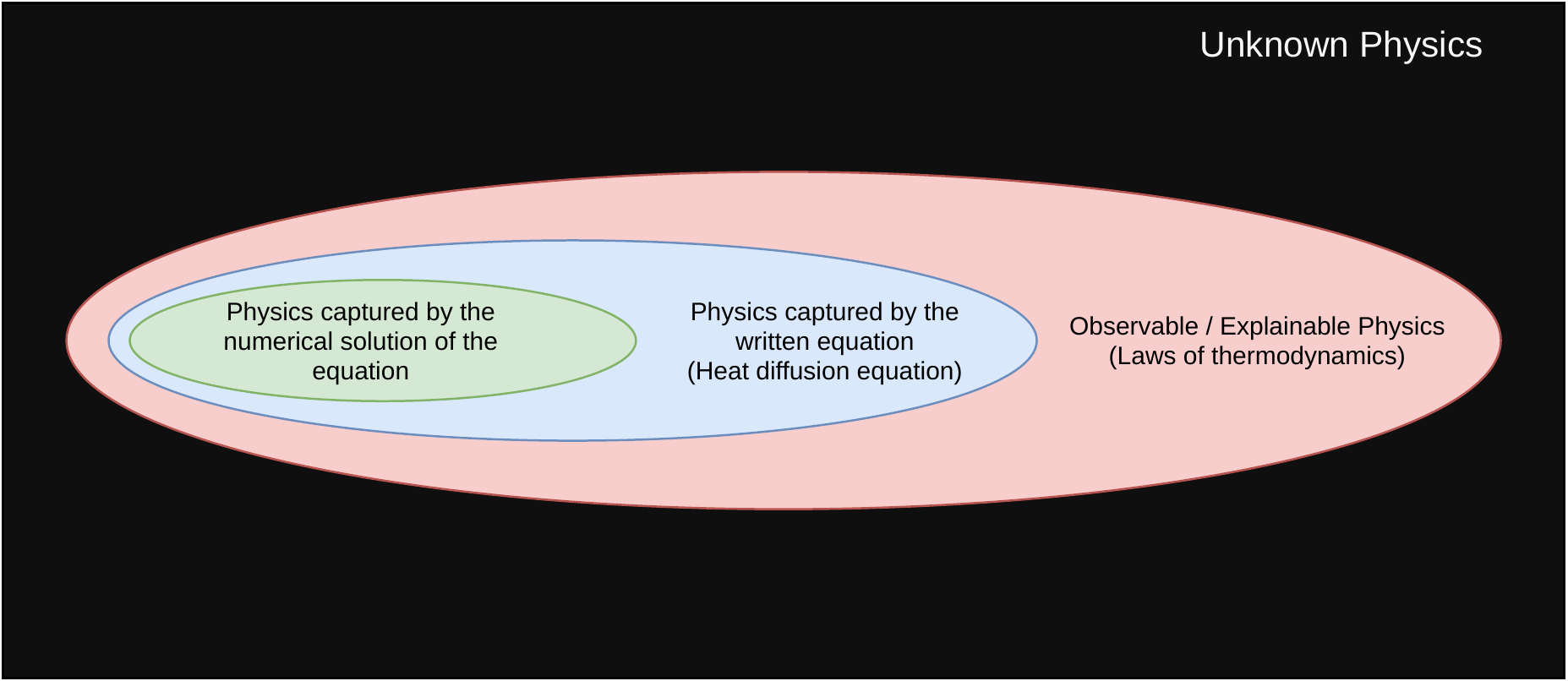}
		\caption{PBM accounts for only a fraction of the full physics.}
		\label{subfig:pbm}
	\end{subfigure}%
	\\
	\begin{subfigure}{\linewidth}
		\centering 
		\includegraphics[width=\textwidth]{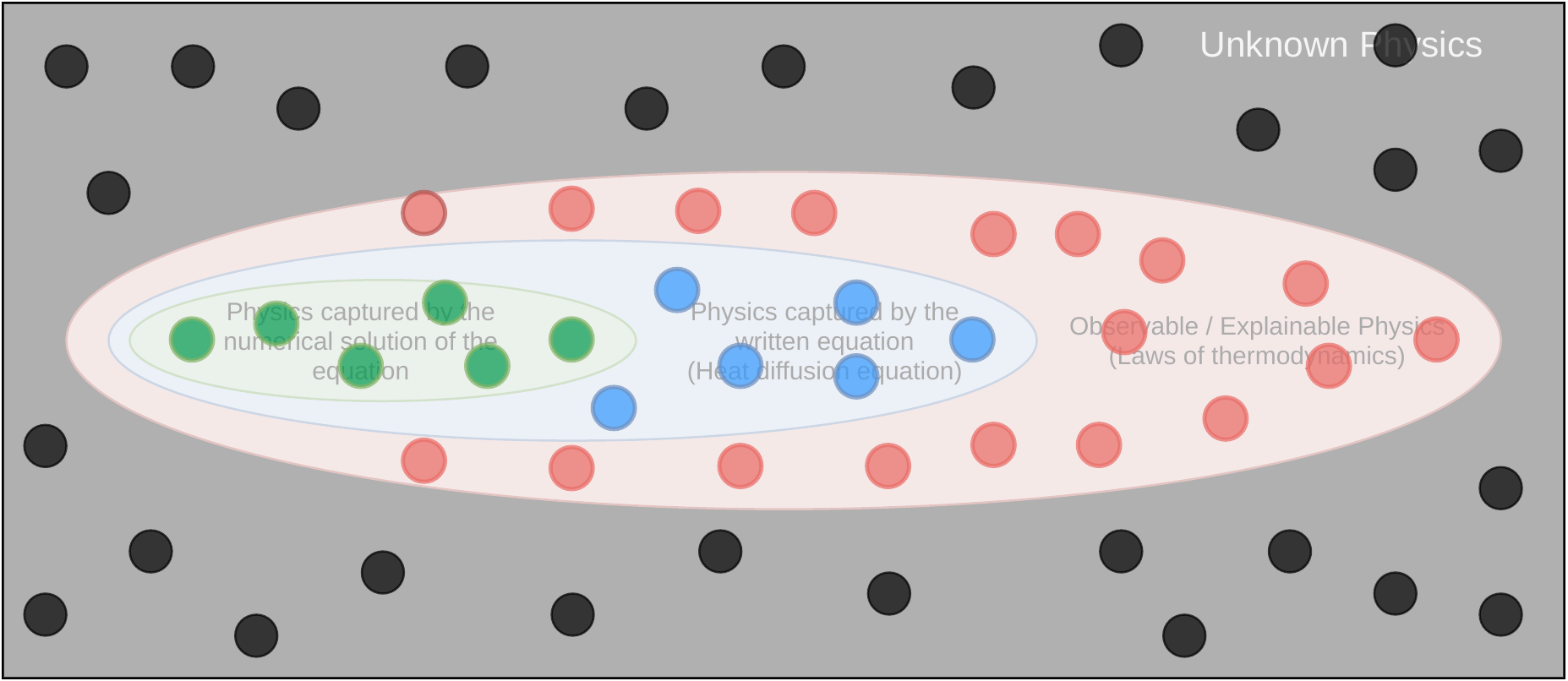}
		\caption{In DDM, data is a manifestation of all physics (known, unknown, resolved, ignored).}
		\label{subfig:ddm}
	\end{subfigure}%
	\caption{Physics based modeling vs data driven modeling.}
	\label{fig:pbm_and_ddm} 
\end{figure}

\textit{Physics-based modeling:} For any real-world system, PBM seeks to explain the system's behavior using existing knowledge of observable and explainable physics (illustrated by the pink ellipse in Figure~\ref{subfig:pbm}). As such, PBM is ignorant of any unknown physics (illustrated by the black background) which, e.g., cannot be observed directly or is inexplicable. Given the first principles of known physics, PBM requires the derivation of one or more governing equations for the system. This derivation might involve making some assumptions, such that only partial physics (the blue ellipse in Figure~\ref{subfig:pbm}) are accounted for by the governing equations. More often than not, these equations are difficult to solve analytically. Therefore, to solve them numerically (in a reasonable amount of time), we make further assumptions, resulting in further loss of physics. Thus, the resolved physics (the green ellipse in Figure~\ref{subfig:pbm}) is generally only a part of the full physics governing the system. PBM has been used extensively for engineering applications like blood flow \citep{TAYLOR1998155}, heat and mass transfer \citep{LIU2021121104} and flow around wind turbines \citep{SIDDIQUI20191058} to name a few. A good overview of PBM in the context of digital twins can be found in \cite{rasheed2020dtv} and \cite{san2021hybrid}. In most of such applications, PBMs tend to be computationally demanding. They are also typically static, meaning that they do not automatically adapt to new scenarios and hence can be inaccurate. Despite these limitations, PBM is attractive due to its sound first-principles foundation which yields great interpretability and generalizability.   

\textit{Data-driven modeling:} With the abundant supply of big data facilitated through, for example, cost-effective sensors, open-source cutting-edge and easy-to-use machine learning libraries, cheap computational infrastructure, and high-quality, readily available training resources, DDM has become very popular. Compared to the PBM approach, DDM thrives on the assumption that data is a manifestation of both known and unknown physics, and hence, when trained with an ample amount of data, DDM will learn the full physics on their own. Data-driven techniques, in particular those involving deep neural networks (DNN), have started achieving human-level performance in several tasks that were, until recently, considered impossible for computers. Notable examples include image classification \citep{inception-4}, dimensionality reduction \citep{Hinton504}, medical treatment \citep{LIU20191} and smart agriculture \citep{BU2019500}. More recent applications of DDM include tropical cyclone intensity estimation \citep{LEE2021104233}, speech recognition (a comprehensive review can be found in \cite{BAI202165}), learning of complex nonlinear dynamics from data \citep{AFEBU202149} and active noise control \citep{ZHANG20211}. Some of the advantages of DDM are their inherent online learning capability, high computational efficiency for inference, and accuracy even for very challenging problems (assuming the training, validation, and test data are prepared properly). However, acceptability of DDM in high-stake applications has been fairly limited due to their data-hungry and black-box nature, poor generalizability, inherent bias, and lack of a robust theory for model stability analysis. In fact, the numerous vulnerabilities of DNN have been highlighted in several recent works \citep{akhtar2018TAA,yuan2017AEA,xu2019adversarial}. 

Due to the challenges described above, it can be concluded that neither PBM nor DDM is ideal for usage in a digital twin context as neither satisfies all the four modeling characteristics identified in \cite{san2021hybrid}. Fortunately, a new paradigm in modeling called Hybrid Analysis and Modeling (HAM) -- which combines the generalizability, interpretability, robust foundation, and understanding of PBM with the accuracy, computational efficiency, and automatic pattern-identification capabilities of advanced DDM, in particular DNNs -- is emerging. In their recent surveys, \cite{willard2020integrating} and \cite{san2021hybrid} provide comprehensive overviews of techniques for integrating DDM with PBM. Most of the hybridization techniques lie in one of the categories of reduced order modeling, physics-guided machine learning (PGML), physics-informed neural network (PINN) or data-driven physics discovery using sparse or symbolic regression. Reduced order models (ROMs) have been proposed as a way of accelerating high-fidelity simulators by several order of magnitude. In a ROM \citep{fonn2019dcp}, complex partial differential equations are projected onto a reduced dimensional space based on an unsupervised algorithm called the proper orthogonal decomposition (more commonly known as the principal component analysis) of the offline high fidelity simulation results resulting in a set of ordinary differential equations (ODEs). One advantage of this method is that many terms in the resulting ODEs can be computed offline using the data, and hence, what remains in an online phase is a simple forward integration of the ODEs in time, which can be very fast. These models however, do not perform well when physics is either missing or get lost during the dimensionality reduction. In the PGML approach introduced by \cite{pawar2021pgml}, partially known physics or knowledge is injected at an intermediate layer in a DNN. Such injection has been shown to improve the accuracy, and reduce the uncertainty of the trained model. However, the PGML models still do not generalize well to extrapolation scenarios. Within a PINN framework \citep{raissi2019physics}, the commonly used mean squared error cost function of the DNN is regularized with the residual of the equation describing the physical laws that should be satisfied. This kind of regularization can pose a challenge during the optimization step because of the increase in the complexity of the cost function. Finally, sparse regression based on $l_1$ regularization and symbolic regression based on gene expression programming have been shown to be very effective in discovering hidden or partially known physics directly from data \citep{vaddireddy2020fes}. The data-driven discovered physics is then added to the PBM to improve their predictions. Notable work using this approach can be found in \cite{Brunton3932}. One of the limitations of this class of method is that, in the case of sparse regression, additional features are required to be handcrafted, while in the case of symbolic regression, the resulting models are often unstable and hence might not be fit for interpretation.

From a careful analysis of the work cited here, it is clear that the HAM approach has the potential to fulfil the four desirable modeling characteristics described earlier in this section. However, as discussed above, earlier approaches to HAM have some limitations. In the current article, we present the \emph{Corrective Source Term Approach} (CoSTA) to HAM. The novelty of the work from a theoretical perspective, is the development of a sound mathematical foundation of the CoSTA approach to augment the governing equation of a PBM describing partial physics with a DNN-generated corrective source term that takes into account the remaining unknown/ignored physics. By doing so, the resulting model retains the generalizability and interpretability of the PBM while exploiting DDM to make the predictions more accurate by modeling the unknown physics reflected in the data. Explainable AI is often attributed to the enhancing processes in which the results of the machine learning models and solutions can be better understood by humans. In our view, our proposed CoSTA approach can be classified as a new explainable AI approach while synthesizing a modular framework between black-box DDM and PBM. From a practical perspective, the superiority of CoSTA in terms of accuracy and generalizability is quantitatively demonstrated by comparing its results against those of pure PBM and DDM in a series of numerical experiments concerning heat diffusion.

In Section~\ref{sec:theory}, we start with the rationale behind the development of CoSTA. We then continue with a presentation of our chosen PBM (Section~\ref{subsec:PBM}) and DDM (Section~\ref{subsec:DDM}), before we explain how the PBM and a DNN can be combined using CoSTA (Section~\ref{subsec:HAM}). We also provide some background on the method of manufactured solutions (Section~\ref{subsec:manufacturedsolution}). Section~\ref{sec:setup} is devoted to explaining our experimental setup -- including the manufactured solutions considered, our DNN architecture and hyperparameter choices, and our data generation, training and testing procedures. Our experimental results are presented and discussed in Section~\ref{sec:resultsanddiscussion} before the article is concluded in Section~\ref{sec:conclusionandfuturework} with a brief summary and an outlook on future work.

\section{Theory}
\label{sec:theory}
In this section we present the rationale behind the CoSTA approach, followed by an overview of the PBM, DDM and HAM models used in our numerical experiments. The section concludes with a brief introduction to the method of manufactured solutions.  

\subsection{The rationale behind the CoSTA approach}
\label{subsec:COSTA-GI}
In this subsection, we present a mathematical foundation for the proposed CoSTA approach. Assume that we are aiming to solve a problem on a domain $\Omega$ (with boundary $\partial \Omega$ and outward pointing unit normal $n$) that can be represented by a linear partial differential equation (PDE) defined as follows:
\begin{align}
    {\cal L} u &= f \,\,\,\,\, \mbox{in} \,\, \Omega \label{equ: PDE_1}\\ 
             u &= g_{\textsc{d}} \,\,\, \mbox{on} \,\, \partial\Omega_{\textsc{d}} \label{equ: PDE_2} \\
    {\cal N} u &= g_{\textsc{n}} \,\,\, \mbox{on} \,\, \partial\Omega_{\textsc{n}} \label{equ: PDE_3}
\end{align}
Here ${\cal L}$ is a linear\footnote{We may generalize the approach to handle nonlinear differential operators, but for notational simplicity we restrict ourselves first to address linear differential operators. See~\cite{blakseth2021ica} for a consideration of non-linear operators} differential operator (e.g.\ ${\cal L}=-\nabla^2$ for Poisson problems), $u$ is the unknown (e.g.\ temperature for heat diffusion problems), $f$ is the (true) source term (e.g. a heat source/sink in heat diffusion problems), $g_{\textsc{d}}$ is the prescribed Dirichlet boundary condition along the boundary section $\partial\Omega_{\textsc{d}}$, and ${\cal N}$ is the differential operator related to the Neumann boundary condition (e.g.\ $\partial u / \partial n$ for heat diffusion problems) with prescribed value $g_{\textsc{n}}$ (e.g.\ heat flux) along the boundary section $\partial\Omega_{\textsc{n}}$. We assume that $\partial\Omega_{\textsc{d}}$ and $\partial\Omega_{\textsc{n}}$ cover the whole boundary $\partial \Omega$ without overlapping each other.

We will now address different cases of uncertainties/errors/lack of information in the abstract PDE defined above. We will, in general, let $\tilde{u}$ denote an analytical solution to a \emph{perturbed} version of the PDE problem defined in Equations~(\ref{equ: PDE_1}--\ref{equ: PDE_3}). Furthermore, we denote \emph{numerically} computed solutions of the original PDE and its perturbation as $u_{\rm{num}}$ and $\tilde{u}_{\rm{num}}$, respectively. Here the subscript ``$_{\rm{num}}$'' indicates the finite resolution of the numerical method (e.g.\ finite difference method (FDM), finite volume method (FVM) or finite element method (FEM)).

Let the error between the two analytical solutions $u$ and $\tilde{u}$
be denoted $\tilde{e}$, i.e.\ we have
\begin{equation}
   \tilde{e}=u-\tilde{u}. 
\end{equation}
We define the corresponding residual $\tilde{r}$ as follows:
\begin{align}
   \tilde{r} &= f - {\cal L}\tilde{u} \\
       &= {\cal L}u - {\cal L}\tilde{u} \\
       &=  {\cal L} \tilde{e} \label{equ:Residual_Error}
\end{align}
Notice that there is a unique relationship between the error in the analytical solution of the perturbed PDE $\tilde{u}$ and the residual obtained by inserting this solution into the (true) original PDE. The CoSTA approach utilizes this relationship, as illustrated for different cases 1--4 below. Cases 1 and 2 concern possible sources of error in the governing PDE itself, while Case 3 is the case when the governing PDE is known without error but cannot be solved analytically. Combinations of Cases 1--3 are treated as Case 4.

{\em Case 1: Inaccurate source term or boundary conditions:}
In many real-world problems, the source term $f$ (e.g.\ describing internal heat generation) may not be known exactly.

Let the inaccurate source term be denoted $\tilde{f}$ and assume that we are able to compute exactly the corresponding PDE such that:
\begin{equation}
   {\cal L} \tilde{u} = \tilde{f} \,\,\,\,\, \mbox{in} \,\, \Omega \\. 
\end{equation}

Assuming that we know $u$ (e.g.\ by measurements or analytical solution), we can add a corrective source term $\tilde{r}$ to compute an improved (analytical) solution denoted $\tilde{u}_{\textsc{costa}}$:
\begin{align}
   {\cal L}u_{\textsc{costa}} &= {\cal L}\tilde{u} + ({\cal L}u - {\cal L} \tilde{u}) \\
               &= \tilde{f} + {\cal L}\tilde{e} \\
               &= \tilde{f} + \tilde{r}
\end{align}
If we are able to evaluate ${\cal L}u$ for a given solution $u$, we get the following relationship from the equations above\footnote{If $u$ is only known in discrete points (e.g.\ it is measured) we may interpolate it or project it onto a polynomial basis of order $p$ to achieve $u_p$ which then can be differentiated and used instead of $u$ in Equation~(\ref{eq:Lu_costa=Lu})}:
\begin{equation}
   {\cal L}u_{\textsc{costa}} = {\cal L}u
   \label{eq:Lu_costa=Lu}
\end{equation}
Thus, by the herein developed CoSTA, we may retain the true analytical solution without any modeling error caused by the inaccurate source term $\tilde{f}$. The main point to be observed here is that inaccuracies in the source term can be corrected for by computing the residual from measured (or manufactured) solutions.

Suppose now that, instead of having an inaccurate source term $\tilde{f}$, we have an inaccurate Dirichlet condition $\tilde{g}_{\textsc{d}}$ (e.g.\ inaccurate surface temperature) or an inaccurate Neumann condition $\tilde{g}_{\textsc{n}}$ (e.g.\ unknown heat flux). We may correct for any error in $\tilde{u}$ caused by $\tilde{g}_{\textsc{d}}$ by replacing it with $u$ along the Dirichlet boundary $\partial\Omega_{\textsc{d}}$ and similarly any error caused by $\tilde{g}_{\textsc{n}}$ is taken care of by replacing it with ${\cal N}u$ along $\partial\Omega_{\textsc{n}}$.

{\em Case 2: Inaccurate physical parameters and differential operators:} Let the true differential operator ${\cal L}$ be dependent on some physical parameter $k$ (e.g.\ heat conductivity) that may be a spatial and/or temporal function. We indicate this dependency by writing ${\cal L}(k)$. Assume now that we do not know the exact value of $k$, but instead only an approximation $\tilde{k}$, such that our perturbed PDE is defined using the operator $\tilde{{\cal L}}={\cal L}(\tilde{k})$.
Alternatively, assume that we do not know (or simply neglect) some terms in the true operator ${\cal L}$ and denote the resulting inaccurate operator $\tilde{{\cal L}}$.

For these two situations we will typically solve the following problem:
\begin{equation}
   \tilde{{\cal L}} \tilde{u} = f \,\,\,\,\, \mbox{in} \,\, \Omega \\. 
\end{equation}

Assuming again that we know $u$ (e.g.\ by measurements or analytical solution) the residual, due to inaccurate differential operator where $\tilde{{\cal L}} \neq {\cal L}$, is given by  Equation~(\ref{equ:Residual_Error}), i.e., $\tilde{r}= {\cal L} \tilde{e}$.
However, if ${\cal L}$ is unknown we cannot compute $\tilde{r}$ from the relations above. Therefore, we introduce an alternative residual $\hat{\tilde{r}}$ corresponding to using the perturbed differential operator as follows:
\begin{equation}
    \hat{\tilde{r}} := \tilde{{\cal L}}\tilde{e}
\end{equation}
We then add a $\hat{\tilde{r}}$ as a corrective source term to find the CoSTA-improved (analytical) solution $u_{\textsc{costa}}$:
\begin{align}
   \tilde{{\cal L}}u_{\textsc{costa}} &= f + \hat{\tilde{r}}  \label{equ:COSTA-Pertubation} \\
    &= \tilde{{\cal L}}\tilde{u} + \tilde{{\cal L}}\tilde{e} \\
                          &= \tilde{{\cal L}}u  \label{equ:ProjectedTrueSolution}
\end{align}
Thus, CoSTA can be looked upon as either solving a ``manufactured solution" defined by the true solution in Equation~(\ref{equ:ProjectedTrueSolution}), or as solving the problem using a perturbed (corrected) source term $f + \hat{\tilde{r}}$ as given in Equation~(\ref{equ:COSTA-Pertubation}) -- in both cases using the (inaccurate) perturbed differential operator $\tilde{{\cal L}}$. 
Notice that, in the above, we get an analytical solution $u_{\textsc{costa}}$ that corresponds to a source term $\tilde{{\cal L}u}$ defined by the true solution $u$ on a perturbed PDE defined by $\tilde{{\cal L}}$. If the perturbed PDE admits a unique analytical solution, then the use of CoSTA will imply that $u_{\textsc{costa}}=u$.

{\em Case 3: Inaccurate differential operator due to discretization errors:} Above we have described inaccuracy in the continuous PDE due to modeling errors. However, when we solve a PDE with FDM/FVM/FEM we introduce discretization errors as we are solving the problem with a discrete approximation ${\cal L}_{\rm{num}}$ of the true differential operator ${\cal L}$ (e.g.\ FDM using central differences). Following the approach for {\em Case 2\/} above, we get the same relationships as given in Equation~(\ref{equ:COSTA-Pertubation}--\ref{equ:ProjectedTrueSolution}) by substituting  $\tilde{\cal L}$ with ${\cal L}_{\rm{num}}$ for problems with only discretization errors and no modeling errors.

{\em Case 4: Combined modeling and discretization errors:} In our study herein, we will address problems where we have both modeling and discretization errors. Denote the corresponding differential operator $\tilde{\cal L}_{\rm{num}}$ and the inaccurate source term $\tilde{f}$. Our approach for retaining the true solution $u$ of the true problem defined by ${\cal L}$ and $f$, is outlined below.

We first solve the following problem to find a predictor $\tilde{u}_{\rm{num}}$:
\begin{equation}
   \tilde{\cal L}_{\rm{num}} \tilde{u}_{\rm{num}} = \tilde{f} \,\,\,\,\, \mbox{in} \,\, \Omega.
   \label{equ:Predictor_Final}
\end{equation}
Then we compute the residual, i.e.\ the corrective source term, corresponding to the error $\tilde{e}_{\rm{num}}= u - \tilde{u}_{\rm{num}}$ in the predictor:\footnote{See Section~\ref{subsec:HAM} for an example of how to compute the corrective source term in practice.}
\begin{equation}
    \hat{\tilde{r}}_{\rm{num}} := \tilde{\cal L}_{\rm{num}}\tilde{e}_{\rm{num}}
    \label{equ:Residual_Final}
\end{equation}
Finally, we do the following corrector step to compute the CoSTA-improved numerical solution:
\begin{align}
   \tilde{\cal L}_{\rm{num}}u_{\textsc{costa}} &= \tilde{f} + \hat{\tilde{r}}_{\rm{num}}  \label{equ:COSTA-Pertubation_Final} \\
    &= \tilde{\cal L}_{\rm{num}}\tilde{u}_{\rm{num}} + \tilde{\cal L}_{\rm{num}}\tilde{e}_{\rm{num}} \\
                          &= \tilde{\cal L}_{\rm{num}}u  \label{equ:ProjectedTrueSolution_Final}
\end{align}
Notice that if we knew the true solution $u(x,t;\bm{\mu})$ at any node, at every time step for any choice of the parameter vector $\bm{\mu}$ {\em a priori\/}, we would not need to do the predictor step or compute the related corrective source term, because we could have solved Equation~(\ref{equ:ProjectedTrueSolution_Final}) directly. However, in practice we do not know $u$ for all choices of $\bm{\mu}$, but we may train a neural network to return a quite accurate corrective source term (formally defined by Equation~(\ref{equ:Residual_Final})) given a predictor $\tilde{u}_{\rm{num}}$ computed by Equation~(\ref{equ:Predictor_Final}). Thus, the corrector step in CoSTA corresponds then to solving Equation~(\ref{equ:COSTA-Pertubation_Final}).

{\em Applications:}
To test and demonstrate the value of the proposed CoSTA approach, we choose the problem of one-dimensional heat conduction described by a PDE derived from the first principles using the first law of thermodynamics. Accurate modeling of heat conduction is vital for a wide array of problems ranging from the modeling of heat transfer from the earth to its atmosphere, modeling heat transfer characteristics of the built environment, and modeling the accumulation of thermal stresses in heat storage infrastructures. However, the accuracy in most of these applications is compromised due to geometric simplifications, uncertainty associated with the values of thermophysical properties used in the calculation, neglection of unknown (and even known) phenomena, and numerical approximations. The CoSTA approach, if successful, has the potential to solve these kinds of issues --- not only for heat conduction modeling, but also for the modeling of any other steady-state or dynamical system that can described by a (system of) PDE(s).

\begin{figure*}[tbh]
	\begin{subfigure}{\linewidth}
		\centering 
		\includegraphics[width=\textwidth]{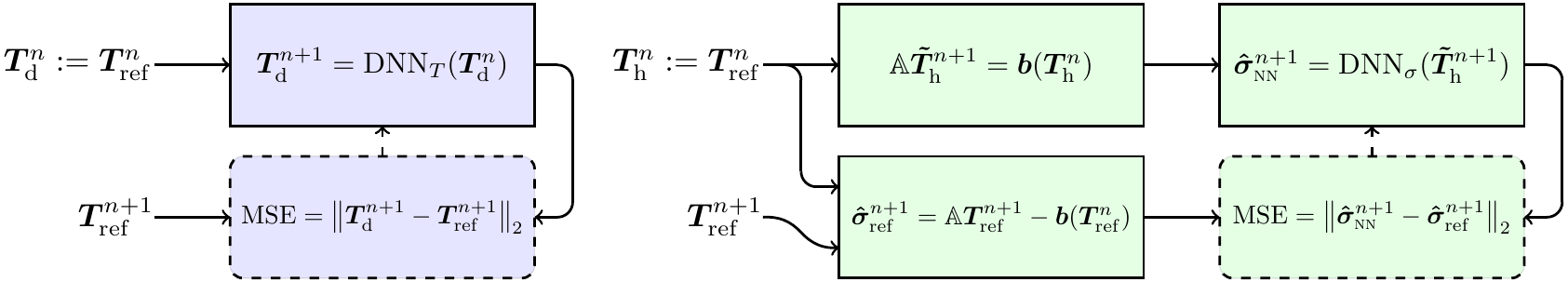}
		\caption{Training procedures for the DNNs used in DDM (left) and HAM (right).}
		\label{subfig:training}
	\end{subfigure}%
	\\
	\begin{subfigure}{\linewidth}
	\vspace{0.5em}
		\centering 
		\includegraphics[width=0.9\textwidth]{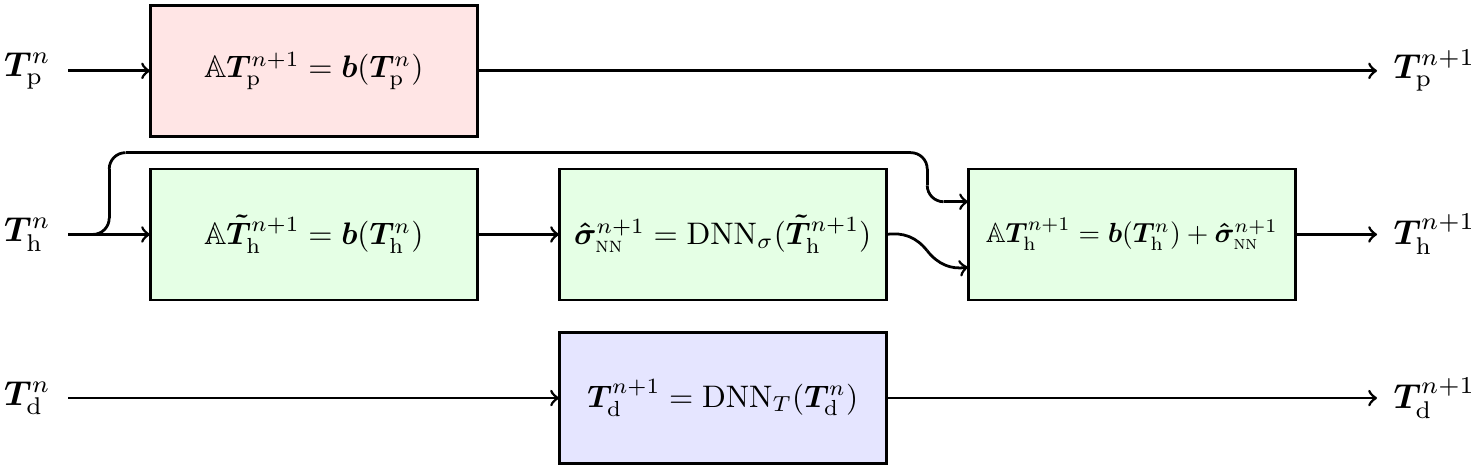}
		\caption{Testing procedures for PBM (top), HAM (middle) and DDM (bottom).}
		\label{subfig:testing}
	\end{subfigure}%
	\caption{Training and testing procedures for the three modeling approaches PBM, DDM and HAM. Note that the PBM does not require any training.}
	\label{fig:traintest} 
\end{figure*}
\subsection{Physics-based modeling}
\label{subsec:PBM}

In PBM, PDEs are widely used as governing equations, describing various physical phenomena by relating partial derivatives of relevant physical quantities. In this paper, we consider the one-dimensional (1D) unsteady heat diffusion equation, which describes 1D transient heat transfer in a system of volume $V$ and cross-sectional area $A$. The equation, which can be derived by applying the principle of energy conservation to the 1st law of thermodynamics, reads
\begin{equation}
\int\limits_V \rho c_V \diffp{T}{t} \mathrm{d}V
= \left( kA \diffp{T}{x} \right)_e - \left( kA \diffp{T}{x} \right)_w + \int\limits_V \hat{q}\ \mathrm{d}V,
\label{eq:integral_form}
\end{equation}
where $T$, $\rho$, $c_V$, and $k$ denote temperature, density, heat capacity, and thermal diffusivity, respectively. The term on the equation's left-hand side represents the momentary change in the system's internal energy. Furthermore, the first two terms on the right-hand side represent the heat flux across the system's right (eastern, denoted by subscript $e$) boundary and left (western, denoted by subscript $w$) boundary, respectively, while the last term on the right-hand side ($\hat{q})$ is a source term which accounts for heat generated within the system. Under certain smoothness requirements, the 1D unsteady heat equation can also be written on the so-called differential form:
\begin{equation}
\rho c_V \diffp{T}{t}
= \diffp{}{x} \left( k \diffp{T}{x} \right) +  \hat{q}.
\label{eq:diff_form}
\end{equation}  
Comparing to Equation~\eqref{equ: PDE_1}, the differential operator of the heat equation is given by
\begin{equation}
    {\cal L} T =  \rho c_V \diffp{T}{t} - \diffp{}{x} \left( k \diffp{T}{x} \right)
\end{equation}
while the source term is $f= \hat{q}$.

In the cases where the solution of Equation~\eqref{eq:integral_form} (or Equation~\eqref{eq:diff_form}) cannot be expressed analytically, approximate solutions can be obtained using numerical methods such as FDMs, FVMs, and FEMs. For Dirichlet boundary conditions (BCs), Equation~\eqref{eq:integral_form} can be written in the form
\begin{equation}
\mathbb{A}\vect{T}^{n+1} = \vect{b}\left(\vect{T}^n; T_a, T_b, \hat{q}\right),
\label{eq:FVM}
\end{equation}
when discretized using the Implicit Euler FVM.
Here, $\vect{T} = [T_1, \dots, T_{N_j}]^{\mathrm{T}}$ denotes the temperature at all $N_j$ interior grid nodes $x_1, \dots, x_{N_j}$. Furthermore, the superscripts $^n$ and $^{n+1}$ denote two subsequent time levels, $\mathbb{A}$ is a tri-diagonal matrix and $\vect{b}$ is a vector depending on $\vect{T}^n$, the BCs ($T_a$ and $T_b$) and $\hat{q}$. Notice that $\mathbb{A}$ is the algebraic matrix representation of the discrete differential operator ${\cal L}_{\rm{num}}$, and $\vect{b}$ is the vector representation of the source term $f$ which also includes the effects of the boundary conditions.

Since Equation~\eqref{eq:FVM} is an approximation of Equation~\eqref{eq:integral_form}, a solution of one of the equations is generally not a solution of the other. Note also that, in cases where the governing equation \eqref{eq:integral_form} is not fully known, Equation~\eqref{eq:FVM} has to be based on an approximation of Equation~\eqref{eq:integral_form}, which causes further discrepancies between the solutions of Equation~\eqref{eq:integral_form} and Equation~\eqref{eq:FVM}, as discussed in Section~\ref{sec:introduction}. To distinguish the two classes of solutions, we use the notation $T_{\mathrm{ref}}$ to denote a solution of the true governing equation~\eqref{eq:integral_form} (i.e., similar to $u$ given by the Equations~(\ref{equ: PDE_1}--\ref{equ: PDE_3}) in the general case) and $\vect{T}_{\mathrm{p}}$ to denote a solution of the discrete system~\eqref{eq:FVM}. In the context of a prediction problem, $T_{\mathrm{ref}}$ is then the ideal prediction, while $\vect{T}_{\mathrm{p}}$ is the prediction made by the PBM (i.e., similar to $\tilde{u}_{\rm{num}}$ given by Equation~(\ref{equ:Predictor_Final}) in the general case). Thus the equation that we solve to generate the PBM solution is given by
\begin{equation}
    \mathbb{A}\vect{T}_{\rm{p}}^{n+1} = \vect{b}\left(\vect{T}_{\rm{p}}^n\right)
    \label{eq:PBM}
\end{equation}
with the prescribed boundary conditions implicitly included in $\vect{b}$. It should be stressed that there is no learning involved in PBM and hence, in Figure~\ref{fig:traintest} -- where we illustrate the training and testing processes for PBM, DDM and HAM -- there is no mention of PBM in the part concerning training (Figure~\ref{subfig:training}).

\subsection{Data-driven modeling}
\label{subsec:DDM}
In DDM, physics is learned directly from the observation data. For transient systems, one common DDM approach is to define a mapping from the observed state at one time level to the observed state at the subsequent time. A DNN is then trained to approximate this mapping. In the context of 1D heat diffusion problems with known Dirichlet BCs, the mapping to be learned by the DNN is
\begin{align}
\label{eq:ideal_hybrid}
\mathrm{DNN}_T : \mathbb{R}^{N_j+2} &\rightarrow \mathbb{R}^{N_j} \quad \mathrm{such\ that} \quad \vect{T}_{\mathrm{d}}^{n+1} = \vect{T}_{\mathrm{ref}}^{n+1},\\
\vect{T}_{\mathrm{d}}^{n} &\mapsto \vect{T}_{\mathrm{d}}^{n+1} \nonumber
\end{align}
where $\vect{T}_{\mathrm{d}}^n$ refers to the temperature profile predicted by the DDM at time level $n$, and $\vect{T}_{\mathrm{ref}}^n$ is the solution $T_{\mathrm{ref}}$ of the true governing equation (Equation~\eqref{eq:integral_form}) sampled at the grid nodes $x_1, \dots, x_{N_j}$ and time level $n$. Note that the dimensionality discrepancy between the DNN's input and output is due to the input containing the boundary temperatures, which the output does not include; since the boundary temperatures are assumed known, there is no reason to have the DNN predict them. However, we do want to include the boundary temperature in the DNN input, since they represent relevant physical information. To avoid notational complexity, we use $\vect{T}_{\mathrm{d}}^n$ to denote both vectors containing and not containing the boundary temperatures. 

Our reason for choosing DNN-based DDM over other applicable DDMs is that DNNs have the ability to approximate any nonlinear function, as guaranteed by the universal approximation theorem. DNNs \citep{Goodfellow-et-al-2016}, which are inspired by the biological neural networks found in e.g. human brains, typically consist of multiple layers with one layer's output being passed through a non-linear activation function before being used as the input of the next layer. Each layer typically consists of a number of processing units with their own tunable parameters commonly called weights and biases. The nature of, and relations between, the processing units vary across different layer types. We refer to the specific composition of different layers used to define a DNN as that DNN's `architecture'. 

We say that a neural network is `trained' when the individual parameters are tuned in an effort to make the network a better approximator of the desired function. In this paper, we train the DNNs using the framework of supervised learning, which requires the preparation of sample DNN inputs and corresponding target outputs. During training, for any sample input, the DNN's output is compared to the corresponding target output using a chosen cost function. Then, the backpropagation algorithm \citep{Goodfellow-et-al-2016} is used to calculate the gradients of the computed cost with respect to the individual network parameters. Finally, the network parameters are updated, typically using a gradient descent algorithm, such as to minimize the cost function. The cost function used in this work is the commonly used mean squared error. An overview of the training and testing approach for DDM is illustrated in Figure~\ref{fig:traintest} with the color blue.

\subsection{Hybrid analysis and modeling with CoSTA}
\label{subsec:HAM}
Given a PBM, the principal goal of the corrective source term approach (CoSTA) is to modify the governing equation solved by the PBM using a corrective source term, such as to recover the true solution of the problem at hand. In this section, we demonstrate how CoSTA can be used in practice to correct the Implicit Euler FVM for unsteady heat transfer (Equation~\eqref{eq:FVM}).

The first step of applying CoSTA to the Implicit Euler FVM is to add the corrective source term $\vect{\hat{\sigma}}^{n+1}$ to the right hand side of Equation~\eqref{eq:FVM}, such as to obtain the modified system
\begin{equation}
    \mathbb{A}\vect{T}_\mathrm{h}^{n+1} = \vect{b}\left(\vect{T}_\mathrm{h}^n; T_a, T_b, \hat{q} \right) + \vect{\hat{\sigma}}^{n+1},
    \label{eq:mod}
\end{equation}
whose solutions we denote $\vect{T}_{\mathrm{h}}$ (the subscript $_{\mathrm{h}}$ corresponds to HAM). Our goal is now to obtain an explicit expression for $\vect{\hat{\sigma}}^{n+1}$ using the framework from Section~\ref{subsec:COSTA-GI}. To this end, notice that Equation~\eqref{eq:mod} is analogous to Equation~\eqref{equ:COSTA-Pertubation_Final} with the following relations:

\begin{equation}
    \tilde{\mathcal{L}}_{\textrm{num}} = \mathbb{A}, \ \
    u_{\textsc{costa}} = \vect{T}_{\mathrm{h}}^{n+1}, \ \ \tilde{f} = \vect{b}, \ \
    \hat{\tilde{r}} = \vect{\hat{\sigma}}^{n+1}.
    \label{eq:relations}
\end{equation}
From the definition of $\hat{\tilde{r}}$ (cf.\ Equation~\eqref{equ:Residual_Final}), we thus have
\begin{align}
    \vect{\hat{\sigma}}^{n+1} = \hat{\tilde{r}} := &\tilde{\mathcal{L}}_{\mathrm{num}}\tilde{e}_{\mathrm{num}}, \\
    = &\tilde{\mathcal{L}}_{\mathrm{num}}u - \tilde{\mathcal{L}}_{\mathrm{num}}\tilde{u}_{\mathrm{num}}
    \label{eq:corr_src_Lu}
\end{align}
where we utilized the definition of $\tilde{e}_{\mathrm{num}}$ to transition from the first to the second line.
As in Section~\ref{subsec:DDM}, we let $\vect{T}_{\mathrm{ref}}^{n+1}$
denote the true solution which we aim to find, i.e., we have $u=\vect{T}_{\mathrm{ref}}^{n+1}$. Moreover, as the analogue of the predictor $\tilde{u}_{\mathrm{num}}$, we choose $\vect{\tilde{T}}_{\mathrm{h}}^{n+1}$ given by
\begin{equation}
    \mathbb{A}\vect{\tilde{T}}_{\mathrm{h}}^{n+1} = \vect{b} \left( \vect{T}_{\mathrm{h}}^{n} \right),
    \label{eq:FVM_predictor}
\end{equation}
where $\mathbb{A}$ and $\vect{b}$ are as defined in Equation~\eqref{eq:mod}. By inserting for $\tilde{\mathcal{L}}_{\mathrm{num}}$, $u$ and $\tilde{u}_ {\mathrm{num}}$ in Equation~\eqref{eq:corr_src_Lu}, we thus obtain
\begin{align}
    \vect{\hat{\sigma}}^{n+1} &= \mathbb{A}\vect{T}_{\mathrm{ref}}^{n+1} - \mathbb{A}\vect{\tilde{T}}_{\mathrm{h}}^{n+1} \\
    &= \mathbb{A}\vect{T}_{\mathrm{ref}}^{n+1} - \vect{b} \left( \vect{T}_{\mathrm{h}}^{n} \right).
\end{align}
If we now insert $\vect{T}_{\mathrm{h}}^n = \vect{T}_{\mathrm{ref}}^{n}$ into the equation above,\footnote{Our primary motivation for doing this is that it produces the desirable result $\vect{T}_{\mathrm{h}}^n = \vect{T}_{\mathrm{ref}}^n$ $\implies$ $\vect{T}_{\mathrm{h}}^{n+1} = \vect{T}_{\mathrm{ref}}^{n+1}\ \forall n \geq 0$.} we get

\begin{equation}
    \vect{\hat{\sigma}}^{n+1} = \mathbb{A}\vect{T}_{\mathrm{ref}}^{n+1} - \vect{b} \left( \vect{T}_{\mathrm{ref}}^{n} \right),
    \label{eq:FVM_corr_src}
\end{equation}
which is the definition of the corrective source term that we will use to generate data for our numerical experiments (cf.\ Section~\ref{subsec:datageneration}). Note that $\vect{\hat{\sigma}}^{n+1}$ corrects the error of the Implicit Euler FVM over \emph{a single} time step. Starting from some know initial temperature profile $\vect{T}^0 = \vect{T}_{\mathrm{h}}^0 = \vect{T}_{\mathrm{ref}}^0$, the combined use of Equations~\eqref{eq:mod} and~\eqref{eq:FVM_corr_src}, guarantees $\vect{T}_{\mathrm{h}}^n = \vect{T}_{\mathrm{ref}}^n$ also for all $n > 0$.\footnote{This can be proven by induction: $\vect{T}_{\mathrm{h}}^n = \vect{T}_{\mathrm{ref}}^n$ $\implies$ $\mathbb{A}\vect{T}_{\mathrm{h}}^{n+1} = \vect{b}(\vect{T}_{\mathrm{h}}^n) + \vect{\hat{\sigma}}^{n+1} = \vect{b}(\vect{T}_{\mathrm{ref}}^n) + \mathbb{A}\vect{T}_{\mathrm{ref}}^{n+1} - \vect{b}(\vect{T}_{\mathrm{ref}}^n) = \mathbb{A}\vect{T}_{\mathrm{ref}}^{n+1}$, which implies $\vect{T}_{\mathrm{h}}^{n+1} = \vect{T}_{\mathrm{ref}}^{n+1}$ since $\mathbb{A}$ has full rank.}

Since $\vect{T}_{\mathrm{ref}}^{n+1}$ is not known \textit{a priori}, Equation~\eqref{eq:FVM_corr_src} -- and hence also Equation~\eqref{eq:mod} -- cannot be evaluated using pure PBM. On the other hand, pure DDM \emph{can} (implicitly) take $\vect{\hat{\sigma}}^{n+1}$ into account, but also completely discards what is already known about heat diffusion problems. Instead of going to either of these extremes, we choose a middle ground by training a deep neural network $\mathrm{DNN}_{\sigma}$ to approximate Equation~\eqref{eq:FVM_corr_src} given the predictor $\vect{\tilde{T}}_{\mathrm{h}}^{n+1}$ defined in Equation~\eqref{eq:FVM_predictor}. The DNN approximation is then inserted into the modified PBM (Equation~\eqref{eq:mod}). That is, we insert $\vect{\hat{\sigma}}_{\textsc{nn}}^{n+1} = \mathrm{DNN}_{\sigma} (\vect{\tilde{T}}_{\mathrm{h}}^{n+1})$ in the place of $\vect{\hat{\sigma}}^{n+1}$ in Equation~\eqref{eq:mod} to obtain the HAM prediction $\vect{T}_{\mathrm{h}}^{n+1}$.

Training data for the DNN can be generated from a known time series describing the system's past, or from time series describing the temporal development of similar systems, using Equation~\eqref{eq:FVM_corr_src}. The whole process of training and testing the complete CoSTA-based HAM model is illustrated in Figure~\ref{fig:traintest} using the color green.

\subsection{Method of manufactured solutions}
\label{subsec:manufacturedsolution}
A central part of the present study is the method of manufactured solutions (MMS), which has long been a popular tool for verifying the numerical PDE solvers used in PBMs (see e.g.~\cite{Roache2002cvb} for an extensive introduction). The key concept of MMS is to choose some explicitly expressible function, and then calculate the source term required for this function to be a solution of the PDE in question. For the 1D unsteady heat diffusion equation, this amounts to deciding upon some temperature function $T(x,t)$ and calculating the source term $\hat{q}(x,t)$ required for the differential form of the equation (Equation~\eqref{eq:diff_form}) to be satisfied. Thus, the use of MMS allows us to obtain exact reference solutions $T_{\mathrm{ref}}$ of the 1D unsteady heat equation without running expensive high-fidelity simulations. We can then use these reference solutions to evaluate the accuracy of the temperature profiles predicted using PBM, DDM and HAM, as well as for generating DNN training and validation data.

\section{Experimental setup and procedures}
\label{sec:setup}

To evaluate the performance of the PBM, DDM and HAM models described in the previous section, we have designed a series of numerical experiments (the results of which are presented and discussed in Section~\ref{sec:resultsanddiscussion}), where each experiment is based on a manufactured solution of the 1D unsteady heat equation (Equation~\eqref{eq:integral_form}). The experimental setup and procedures used to conduct these experiments are described in the following section.

\subsection{Choice of manufactured solutions}
\label{subsec:manufacturedsolutionevaluated}
 All manufactured solutions $T(x,t;\alpha)$ used to conduct the present study are listed in Table~\ref{tab:manufactured_solutions}. They include both polynomials and trigonometric functions, so as to cover a wide variety of different functional behaviours. Each solution is parametrized by a quantity $\alpha$, which allows us to investigate the generalizability of the PBM, DDM and HAM methods across different $\alpha$-values (corresponding to generalizability across different operating conditions in an application context). Along with each manufactured solution, we have included the corresponding source term $\hat{q}$ required for the listed functions $T(x,t;\alpha)$ to satisfy the 1D heat equation (Equation~\eqref{eq:integral_form} or, equivalently, Equation~\eqref{eq:diff_form}). From a real world application perspective, the problem corresponds to a one dimensional solid body initialized (at $t=0$) with a temperature profile using the functions given in the Table \ref{tab:manufactured_solutions}. Then the evaluation of the temperature across the solid body is predicted under the influence of differential heating across the length of the solid body given by the source term $\hat{q}$. This heat transfer phenomenon is encountered in a wide variety of real life applications like heat loss / gain through the building walls, transfer of heat from the center of the earth to the atmosphere, and cooling / heating of electric chip on electrical equipment.

\begin{table*}[tb]
	\centering
	\caption{Manufactured solutions: Functions $T(x,t;\alpha)$ used for performance evaluation of the studied PBM, DDM and HAM methods (top five rows) and for DNN hyperparameter tuning (bottom two rows). All $T(x,t;\alpha)$ were defined on the spatial domain $x\in[\SI{0}{\meter}, \SI{1}{\meter}]$ and the temporal domain $t\in[\SI{0}{\second}, \SI{5}{\second}]$.}
	\begin{tabular}{lll}
		\toprule
		Solutions & $T(x,t;\alpha)$ & $\hat{q}(x,t;\alpha)$ \\
		\midrule
		0  & $\alpha\left(t + \frac{1}{2}x^2\right)$ & 0 \\
		1  & $t + \frac{1}{2}\alpha x^2$ & $1-\alpha$ \\
		2  & $\sqrt{t + \alpha + 1} + 10x^2(x-1)(x+2)$ & $\frac{1}{2\sqrt{t+\alpha+1}} - 120x^2 -60x + 40$ \\
		3  & $2 + \alpha(x-1)\tanh{(\frac{x}{t+0.1})}$ & $\frac{\alpha}{(t+0.1)^2}\left( x(1-x) + 2 \left( (x-1)\tanh{(\frac{x}{t+0.1})} - t - 0.1 \right) \right) \sech^2{(\frac{x}{t+0.1})}$  \\
		4  & $1 + \sin{(2\pi t + \alpha)}\cos{(2\pi x)}$ & $2\pi\left( \cos{(2\pi t + \alpha)} - 2\pi\sin{(2\pi t + \alpha)} \right) \cos{(2\pi x)}$ \\
		\midrule
		A  & $\sqrt{t + \alpha + 1} + 7x^2(x-1)(x+2)$ & $\frac{1}{2\sqrt{t+\alpha+1}} - 84x^2 -42x + 28$ \\
		B  & $-\frac{x^3(x-\alpha)}{t + 0.1}$    & $\frac{x^4 - \alpha x^3}{(t+0.1)^2} + \frac{12x^2-6\alpha x}{t+0.1}$ \\
		\bottomrule
	\end{tabular}
	\label{tab:manufactured_solutions}
\end{table*}



\subsection{Parametrization} 
For each manufactured solution in Table~\ref{tab:manufactured_solutions}, the 22 different $\alpha$-values listed in Table~\ref{tab:alphas} were used to generate the training, validation and testing data. Of the 22 $\alpha$-values, 16 were used to generate training data, 2 were used to generate validation data and 4 were used to generate testing data, as indicated in the table. Of the $\alpha$-values used for testing, two lie within the interval $[0.1, 2.0]$ covered by $\mathcal{A}_{\mathrm{train}}$ (defined in Table~\ref{tab:alphas}) while the other two do not. This allows us to evaluate the generalizability of PBM, DDM and HAM in both interpolation and extrapolation scenarios.

\begin{table}[tb]
	\centering
	\caption{Parametrization: Selection of $\alpha$-values used for generating training, validation and testing time series.}
	\begin{tabular}{lll}
		\toprule
		Purpose & Set of $\alpha$-values & Symbol     \\
		\midrule
		Training & $\{0.1, 0.2, \dots, 2.0\}\backslash\{0.7, 0.8, 1.1, 1.5\}$ & $\mathcal{A}_{\mathrm{train}}$ \\
		Validation & \{0.8, 1.1\} & $\mathcal{A}_{\mathrm{val}}$ \\
		Testing  & $\{-0.5, 0.7, 1.5, 2.5\}$ & $\mathcal{A}_{\mathrm{test}}$ \\
		\bottomrule
	\end{tabular}
	\label{tab:alphas}
\end{table}

\subsection{Training and validation data generation}\label{subsec:datageneration}
Given an $\alpha$ and a manufactured solution $T(x,t;\alpha)$ of the 1D unsteady heat equation, we sample $T(x,t;\alpha)$ at the center of $N_j=20$ equally sized grid cells\footnote{With the exception of one experiment in Section~\ref{subsec:exp_no_mod_error}, where we used $N_j=200$ equally sized grid cells.} on the spatial domain $[\SI{0}{\meter}, \SI{1}{\meter}]$ and at $N_t = 5001$ equally spaced time levels on the temporal domain $[\SI{0}{\second}, \SI{5}{\second}]$ to generate a reference time series $\{\vect{T}_{\mathrm{ref}}^n\}_{n=0}^{N_t-1}$. Training and validation datasets are then generated using the reference time series as described in Algorithm~\ref{alg:data}. This results in a total of $5000\cdot 16 = 80000$ training examples and $5000\cdot2 = 10000$ validation examples (per model) for each experiment, since we use 16 $\alpha$-values to generate training data and 2 $\alpha$-values to generate validation data (cf.\ Table~\ref{tab:alphas}). Moreover, each data example contains two vectors, an $N_j+2$-dimensional DNN input vector and an $N_j$-dimensional DNN target output vector.\footnote{As explained for DDM in Section~\ref{subsec:DDM}, the dimensionality reduction occurs because we take the BCs to be known without error, which means the DNN need not make predictions for the boundaries. This also applies to HAM.} The precise defintions of these vectors are given in Algorithm~\ref{alg:data}. Note that, for the experiments described in Section~\ref{subsec:exp_with_mod_error}, we set $\hat{q}=0$ in Equation~\eqref{eq:FVM} and Equation~\eqref{eq:mod} to synthesize modeling error. Note also that, for each time level $n$, all data in the datasets are generated based on $\vect{T}_{\mathrm{ref}}^{n-1}$. This means that no error can accumulate across multiple time steps, so DNNs trained using these datasets will necessarily be trained to make corrections across individual time steps. That is, they correct only local, not global, time stepping errors.
This choice was made based on three observations: 1) A successful, generalizable reduction in local error will necessarily also reduce global error. 2) Corrections of global errors are hard to interpret due to error accumulation. 3) Data is more efficiently obtained when one data example corresponds to a single time step rather than a complete time series comprising multiple time steps.

\subsection{DNN training} 
Both DDM and CoSTA-based HAM require training of DNNs using the datasets described above. For each training iteration, one batch of data examples $(\vect{T}_{\mathrm{ref}}^{n-1}, \vect{\tilde{T}}_{\mathrm{h}}^n, \vect{T}_{\mathrm{ref}}^n, \vect{\hat{\sigma}}^n)$ is used to update the network's parameters. However, the components of each data example are used differently for the two methods. For HAM, $\vect{\tilde{T}}_{\mathrm{h}}^n$ is given as input to the DNN, and the corresponding DNN output is then compared to $\vect{\hat{\sigma}}^n$ using a cost function. On the other hand, for DDM, $\vect{T}_{\mathrm{ref}}^{n-1}$ is used as DNN input, and the corresponding output is compared to $\vect{T}_{\mathrm{ref}}^{n}$, again using the cost function. We use the mean squared error (MSE) cost function, as implemented in the PyTorch ML framework \cite{paszke_pytorch_2019}. Given the computed cost, the DNN parameters are updated using backpropagation and the chosen optimization algorithm. We use the Adam optimizer introduced in \cite{kingma2014aam}. The full DNN training process is illustrated in Figure~\ref{subfig:training}, where we have also illustrated how $\vect{\hat{\sigma}}^n$ and $\vect{\tilde{T}}_{\mathrm{h}}^n$ were generated from the reference profiles $\vect{T}_{\mathrm{ref}}^{n-1}$ and $\vect{T}_{\mathrm{ref}}^n$, as described in the previous section. 

At regular intervals (validation periods), the total MSE cost for all data examples in the \emph{validation} set is computed. The validation cost is computed analogously to the training cost, as described above.
We utilize the early stopping regularization technique by stopping the DNN training if a new lowest validation cost has not been recorded for a certain number of consecutive validation periods (this number is denoted ``overfit limit'' in Table~\ref{tab:hyperparameters}).

\subsection{Model evaluation} 
When deploying a predictive model for use on a practical application, it is imperative that the model remains accurate across multiple time levels without relying on the reference data. Thus, when evaluating the performance of PBM, DDM and HAM, we need to look at the \emph{global} error of their predictions, even though, for DDM and HAM, the DNNs were trained to make local corrections only. For DDM, this means that $\vect{T}_{\mathrm{d}}^{n-1}$ -- not $\vect{T}_{\mathrm{ref}}^{n-1}$ -- must be used as DNN input to generate $\vect{T}_{\mathrm{d}}^{n}$. Similarly, for HAM, $\vect{T}_{\mathrm{h}}^{n-1}$ -- not $\vect{T}_{\mathrm{ref}}^{n-1}$ -- must be inserted into Equation~\eqref{eq:FVM_predictor} (with $\hat{q}=0$ for the experiments of Section~\ref{subsec:exp_with_mod_error}) to generate the DNN input $\vect{\tilde{T}}_{\mathrm{h}}^{n}$.
For the sake of completeness, we also state that for PBM, $\vect{T}_{\mathrm{p}}^{n}$ is calculated by inserting $\vect{T}_{\mathrm{p}}^{n-1}$ -- not $\vect{T}_{\mathrm{ref}}^{n-1}$ -- into Equation~\eqref{eq:PBM} (again with $\hat{q}=0$ for the experiments described in Section~\ref{subsec:exp_with_mod_error}).
The testing procedure is described in its entirety in Algorithm~\ref{alg:test} and is illustrated in Figure~\ref{subfig:testing} for a single time step.
Since we perform testing using 4 distinct $\alpha$-values (cf.\ Table~\ref{tab:alphas}), this procedure yields $5000\cdot4 = 20000$ test predictions (per model) in any given experiment. We use the $\ell_2$-norm to quantify the accuracy of these predictions, as described in the beginning of Section~\ref{sec:resultsanddiscussion}.
For a general $N$-dimensional vector $\vect{v} \in \mathbb{R}^{N}$, the definition of the $\ell_2$-norm reads
\begin{equation}
    \norm{\vect{v}}_2 = \frac{1}{N} \left( \sum\limits_{i=1}^N v_i^2 \right)^{1/2},
\end{equation}
where $v_i$ are the components of $\vect{v}$.

\begin{algorithm}[tbh]
	\SetAlgoNoLine
	Define grid nodes $\vect{x}$ and time levels $t^0, t^1, \dots, t^{N_t-1}$. \\
	\For{\textnormal{$T(x,t; \alpha)$ in Table~\ref{tab:manufactured_solutions}}}{
		\For{\textnormal{$\alpha$ in $\mathcal{A}_{\mathrm{train}}$ or $\mathcal{A}_{\mathrm{val}}$}}{
			Initialize: $\vect{T}_{\mathrm{ref}}^0  = T(\vect{x}, t^0; \alpha)$ \\
			\For{$n = 1, 2, \dots, N_t-1$}{
			    Get DDM input by sampling:
			    $\vect{T}_{\mathrm{ref}}^{n-1} \gets T(\vect{x}, t^{n-1}; \alpha)$. \\
    			Get HAM input $\vect{\tilde{T}}_{\mathrm{h}}^n$ by inserting $\vect{T}_{\mathrm{ref}}^{n-1}$ into Equation~\eqref{eq:FVM_predictor}. \\
				Get DDM target output by sampling: $\vect{T}_{\mathrm{ref}}^n \gets T(\vect{x}, t^n; \alpha)$. \\
				Get HAM target output $\vect{\hat{\sigma}}^n$ using $\vect{T}_{\mathrm{ref}}^{n}$ and $\vect{T}_{\mathrm{ref}}^{n-1}$ in Equation~\eqref{eq:FVM_corr_src}. \\
				Store $(\vect{T}_{\mathrm{ref}}^{n-1}, \vect{\tilde{T}}_{\mathrm{h}}^n, \vect{T}_{\mathrm{ref}}^n, \vect{\hat{\sigma}}^n)$ as an individual data example in the appropriate dataset.
			}
		}
		Normalize all data examples in both datasets.
	}
	\caption{Data generation procedure for the training and validation datasets. All training data is normalized to have a component-wise mean of 0 and standard deviation of 1. The validation data is normalized using the same normalization coefficients as the training set. For the experiments where modeling error is synthesized, we set $\hat{q}=0$ in Equations~\eqref{eq:FVM_predictor} and~\eqref{eq:mod}.}
	\label{alg:data}
\end{algorithm}

\begin{algorithm}[tbh]
	\SetAlgoNoLine
	Define grid nodes $\vect{x}$ and time levels $t^0, t^1, \dots, t^{N_t - 1}$. \\
	\For{\textnormal{$T(x,t;\alpha)$ in top part of Table~\ref{tab:manufactured_solutions}}}{
		\For{$\alpha$ $\mathrm{in}$ $\mathcal{A}_{\mathrm{test}}$}{
			Initialize: $\vect{T}_{\mathrm{ref}}^0 = \vect{T}_{\mathrm{p}}^0 = \vect{T}_{\mathrm{d}}^0 = \vect{T}_{\mathrm{h}}^0 = T(\vect{x}, t^0; \alpha)$ \\
			\For{$n = 1, 2, \dots, N_t-1$}{
				Sample manufactured solution: $\vect{T}_{\mathrm{ref}}^n = T(\vect{x}, t^n; \alpha)$. \\
				Get PBM prediction $\vect{T}_{\mathrm{p}}^n$ by inserting $\vect{T}_{\mathrm{p}}^{n-1}$ into Equation~\eqref{eq:PBM}.\\
				Get HAM DNN input $\vect{\tilde{T}}_{\mathrm{h}}^n$ by inserting
				$\vect{T}_{\mathrm{h}}^{n-1}$ into Equation~\eqref{eq:FVM_predictor}.\\
				Get DDM prediction: $\vect{T}_d^n \gets \mathrm{DNN}_{T}(\vect{T}_d^{n-1})$ \\
				$\vect{\hat{\sigma}}_{\textsc{nn}}^n \gets \mathrm{DNN}_{\sigma}(\vect{\tilde{T}}_{\mathrm{h}}^n)$ \\
				Get HAM prediction $\vect{T}_{\mathrm{h}}^n$ by inserting $\vect{T}_{\mathrm{h}}^{n-1}$ and $\vect{\hat{\sigma}}_{\textsc{nn}}^n$ into Equation~\eqref{eq:mod}.\\
				Relative $l_2$ error of PBM prediction $\gets$ $\norm{\vect{T}_{\mathrm{p}}^n - \vect{T}_{\mathrm{ref}}^n}_2 / \norm{\vect{T}_{\mathrm{ref}}^n}_2$ \\
				Relative $l_2$ error of DDM prediction $\gets$ $\norm{\vect{T}_d^n - \vect{T}_{\mathrm{ref}}^n}_2 / \norm{\vect{T}_{\mathrm{ref}}^n}_2$ \\
				Relative $l_2$ error of HAM solution $\gets$ $\norm{\vect{T}_{\mathrm{h}}^n - \vect{T}_{\mathrm{ref}}^n}_2 / \norm{\vect{T}_{\mathrm{ref}}^n}_2$
			}
		}
	}
	\caption{Testing algorithm used in our numerical experiments. Note that, to match the operating conditions seen during training, all DNN inputs/outputs are normalized/unnormalized using the same normalization coefficients as were used to normalize the training and validation data (cf. Algorithm~\ref{alg:data}). For the experiments where modeling error is synthesized, we set $\hat{q}=0$ in Equations~\eqref{eq:PBM},~\eqref{eq:FVM_predictor} and~\eqref{eq:mod}.}
	\label{alg:test}
\end{algorithm}

\subsection{Neural network architectures and hyperparameters}
\label{subsec:architectures_and_hyperparameters}

\begin{figure}
    \centering
	\includegraphics[trim=150 550 140 120,clip, width=0.9\linewidth]{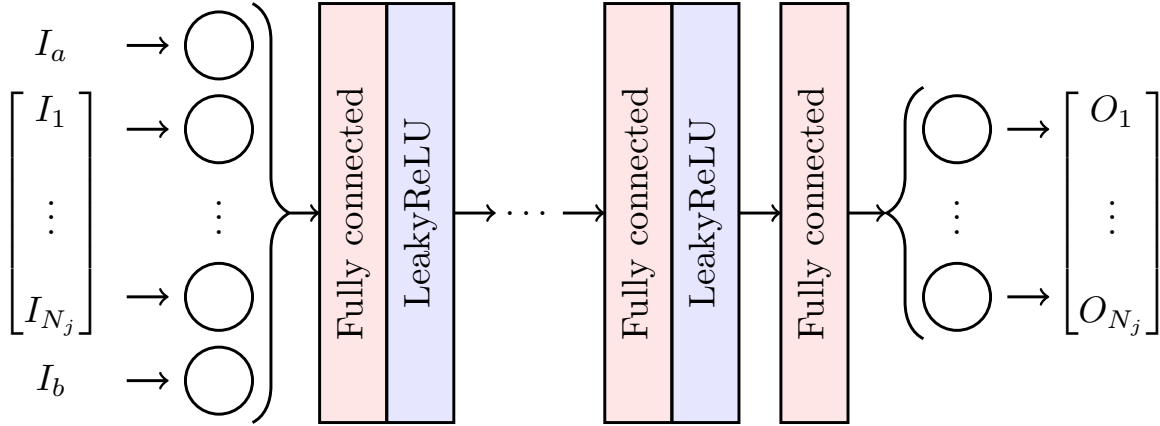}
	\caption{The fully connected neural network architecture used in the present work.
	The definitions of the input vector $\vect{I}$ and the output vector $\vect{O}$ depend on whether the network is used for HAM or DDM. However, note that $\vect{I}$ always has two more components than $\vect{O}$ due to the Dirichlet boundary conditions considered in this work.
	}
	\label{fig:arch} 
\end{figure}

For both DDM and CoSTA-based HAM, we use DNNs with the fully connected neural network (FCNN) architecture illustrated in Figure~\ref{fig:arch}, and the hyperparameters listed in Table~\ref{tab:hyperparameters}. The optimal set of hyperparameters were found using a grid search for the DDM and then applied to HAM as well. The number of fully connected (FC) layers, the width of each FC layer and the learning rate were chosen such as to minimize the total validation loss obtained for the two manufactured solutions A and B at the bottom of Table~\ref{tab:hyperparameters} in a parameter grid search. The other hyperparameters values were chosen based on prior experience with DNN tuning.
We make no claim that our architecture choices are optimal. To the contrary, we expect that using DNN architectures specifically constructed for time series problems (e.g.\ LSTM networks) would improve DDM and HAM performance. However, in the present work, we have chosen to use a simple DNN such that its technical details would not obscure our main contribution, which is the formal introduction and justification of CoSTA. Therefore, comparisons of different architectures and hyperparameters are deferred to future research, as touched upon in Section~\ref{sec:conclusionandfuturework}.

\begin{table}
	\centering
	\caption{Hyperparameters of the DNN used in the present work.}
	\begin{tabular}{ll}
		\toprule
		Parameter               & Value     \\
		\midrule
		Loss function           & MSE       \\
		Learning rate           & 1e-5      \\
		Optimizer               & Adam      \\
		Batch size              & 32        \\
		Number of hidden FC layers     & 4         \\
		Hidden FC layer width   & 80        \\
		LeakyReLU slope         & 0.01      \\
		Validation period       & 1e2       \\
		Overfit limit           & 20        \\
		\bottomrule
	\end{tabular}
	\label{tab:hyperparameters}
\end{table}

\section{Results and discussion}
\label{sec:resultsanddiscussion}

In each numerical experiment, we consider one of the manufactured solutions listed in the top section of Table~\ref{tab:manufactured_solutions}. In any given experiment, we aim to reproduce the time series $\{\vect{T}_{\mathrm{ref}}^{n}\}_{n=0}^{N_t-1}$ for each $\alpha \in \mathcal{A}_{\mathrm{test}}$ using our PBM, DDM and HAM models. To quantify the models' performance, we present the temporal development of the relative $l_2$-errors, $\norm{\vect{T}_{\mathrm{p}}^n - \vect{T}_{\mathrm{ref}}^n}_2 / \norm{\vect{T}_{\mathrm{ref}}^n}_2$, $\norm{\vect{T}_{\mathrm{d}}^n - \vect{T}_{\mathrm{ref}}^n}_2 / \norm{\vect{T}_{\mathrm{ref}}^n}_2$ and $\norm{\vect{T}_{\mathrm{h}}^n - \vect{T}_{\mathrm{ref}}^n}_2 / \norm{\vect{T}_{\mathrm{ref}}^n}_2$, of the PBM-, DDM- and HAM predictions with respect to the sampled manufactured solution $\vect{T}_{\mathrm{ref}}^n$. We also present the temperature profiles predicted by PBM, DDM and HAM at the final time level $n = N_t -1$ (corresponding to $t=\SI{5.0}{\second}$) alongside the manufactured solution $T(x, t^{N_t-1}; \alpha)$.


The boundary conditions and initial conditions corresponding to the manufactured solutions are assumed known and utilized as applicable for all three model types in all experiments.  However, in order to synthesize scenarios where some relevant physics are unknown, for the experiments discussed in Section~\ref{subsec:exp_with_mod_error}, it is assumed that we do not know the source terms $\hat{q}$ required for the manufactured solutions to satisfy Equation~\eqref{eq:integral_form}. No such limitation is imposed on the models discussed in Section~\ref{subsec:exp_no_mod_error}.

\subsection{Experiments without modeling error}
\label{subsec:exp_no_mod_error}
In this section, we compare the performance of the three approaches PBM, DDM and HAM in situations where we have full knowledge of the physics. This implies that the structure of the governing equation, including the source term and the exact values of the PBM parameters, are fully known. The following sections present the results for interpolation and extrapolation scenarios for Solutions~0 and~3 (cf. Table~\ref{tab:manufactured_solutions}). For the current experiment using Solution~3, we used a finer spatial discretization than elsewhere in this work ($N_j=200$, rather than $N_j=20$) in order to demonstrate the full power PBM when all physics is known.

\subsubsection{Interpolation scenarios}
\begin{figure}
	\begin{subfigure}[b]{0.5\linewidth}
		\centering 
		\includegraphics[width=\textwidth]{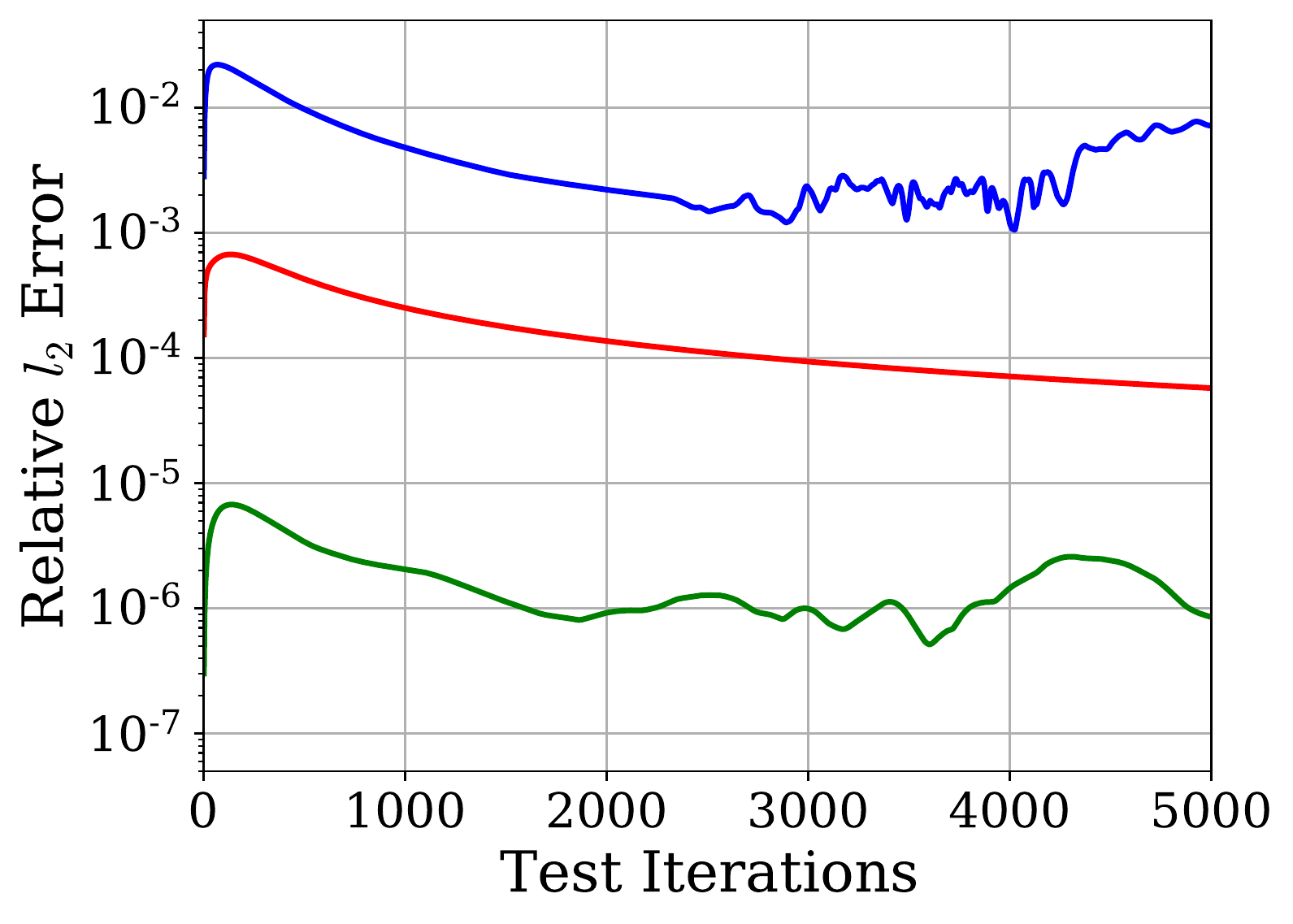}
		\caption{$\alpha = 0.7$, relative errors.}
		\label{subfig:s0_error_a0.7}
	\end{subfigure}%
	\begin{subfigure}[b]{0.5\linewidth}
		\centering 
		\includegraphics[width=\textwidth]{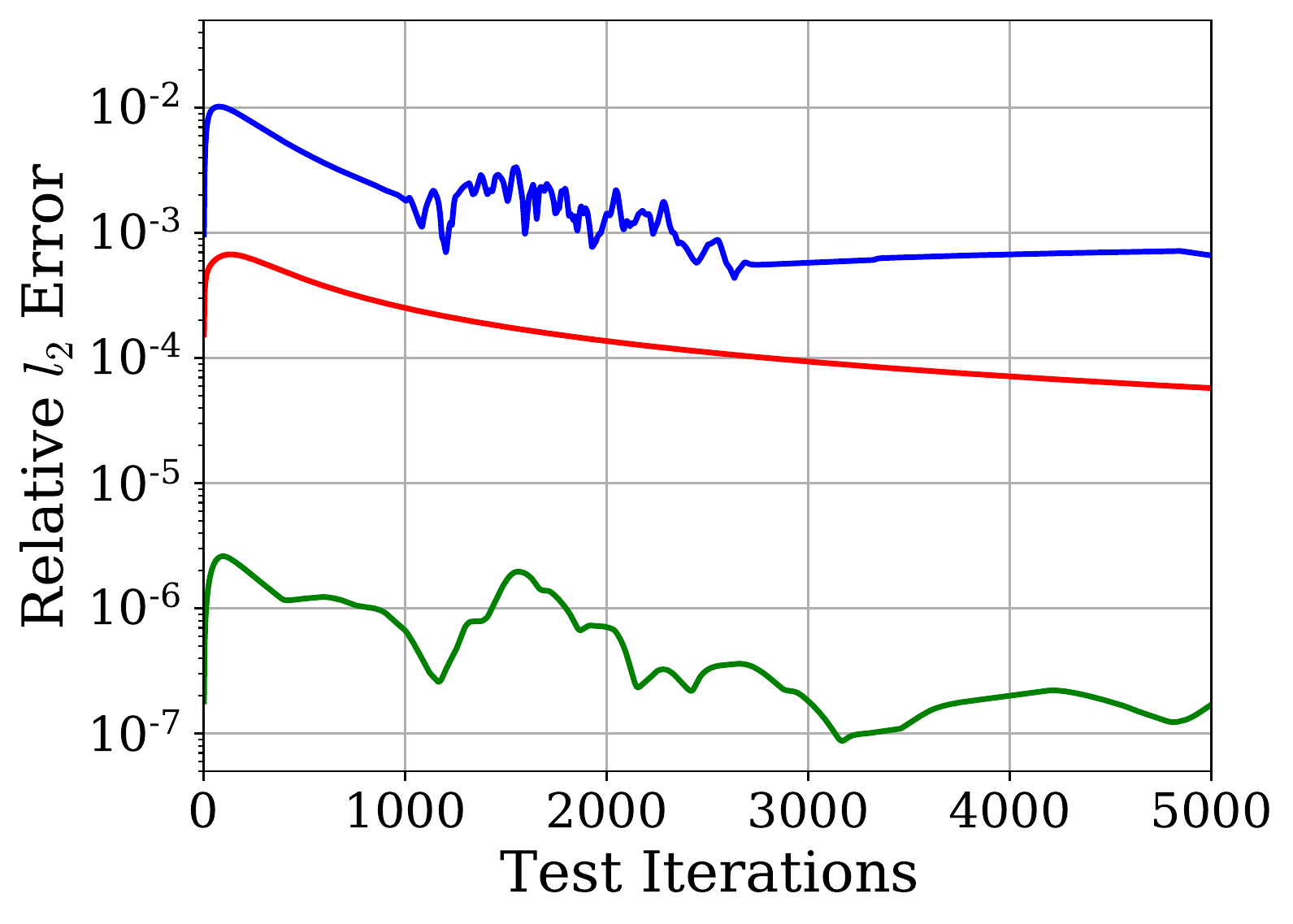}
		\caption{$\alpha = 1.5$, relative errors.}
		\label{subfig:s0_error_a1.5}
	\end{subfigure}%
	\\
	\begin{subfigure}[b]{0.5\linewidth}
		\centering 
		\includegraphics[width=\textwidth]{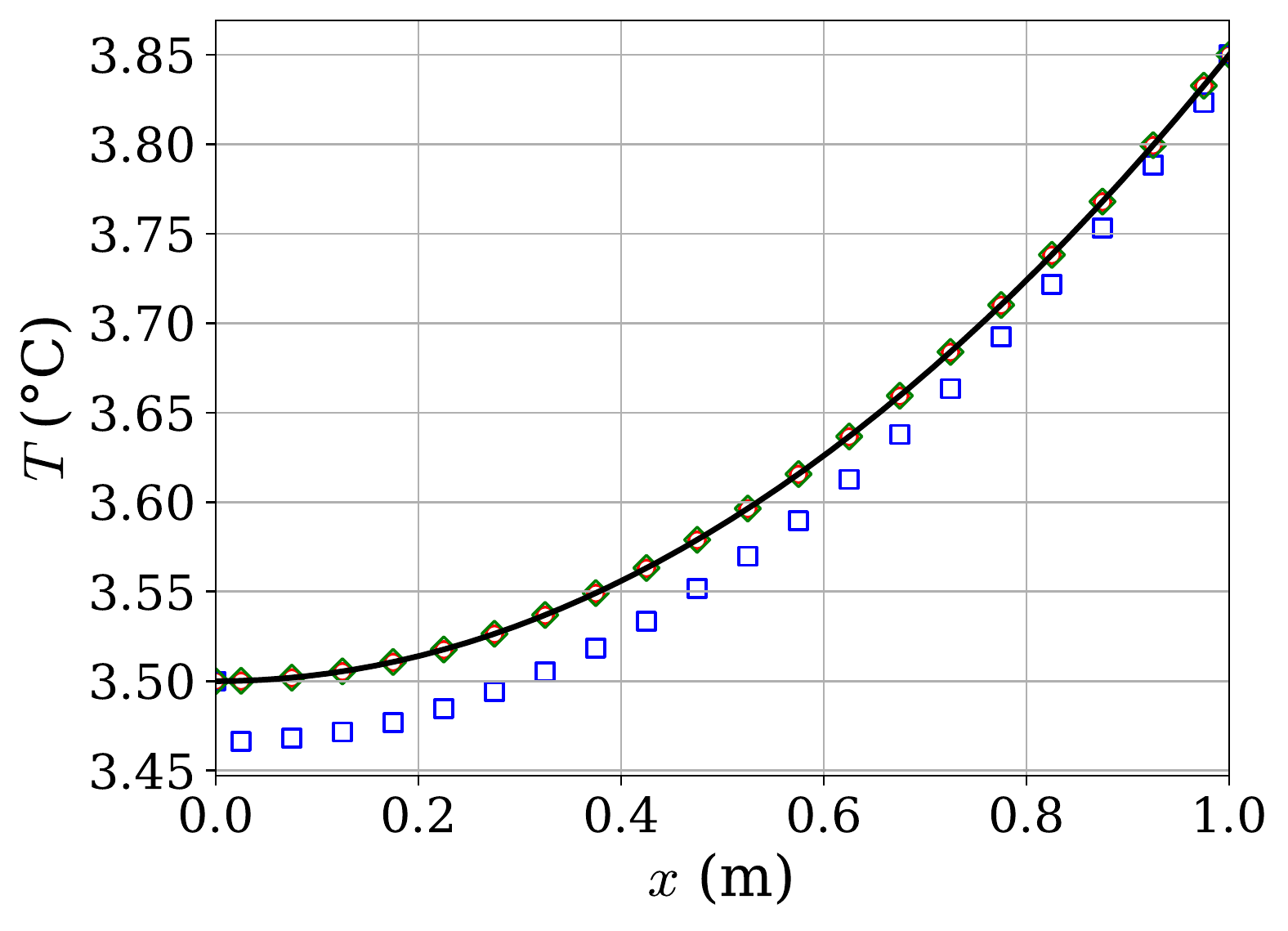}
		\caption{$\alpha = 0.7$, time level $n=5000$.}
		\label{subfig:s0_profile_a0.7_t2}
	\end{subfigure}%
	\begin{subfigure}[b]{0.5\linewidth}
		\centering 
		\includegraphics[width=\textwidth]{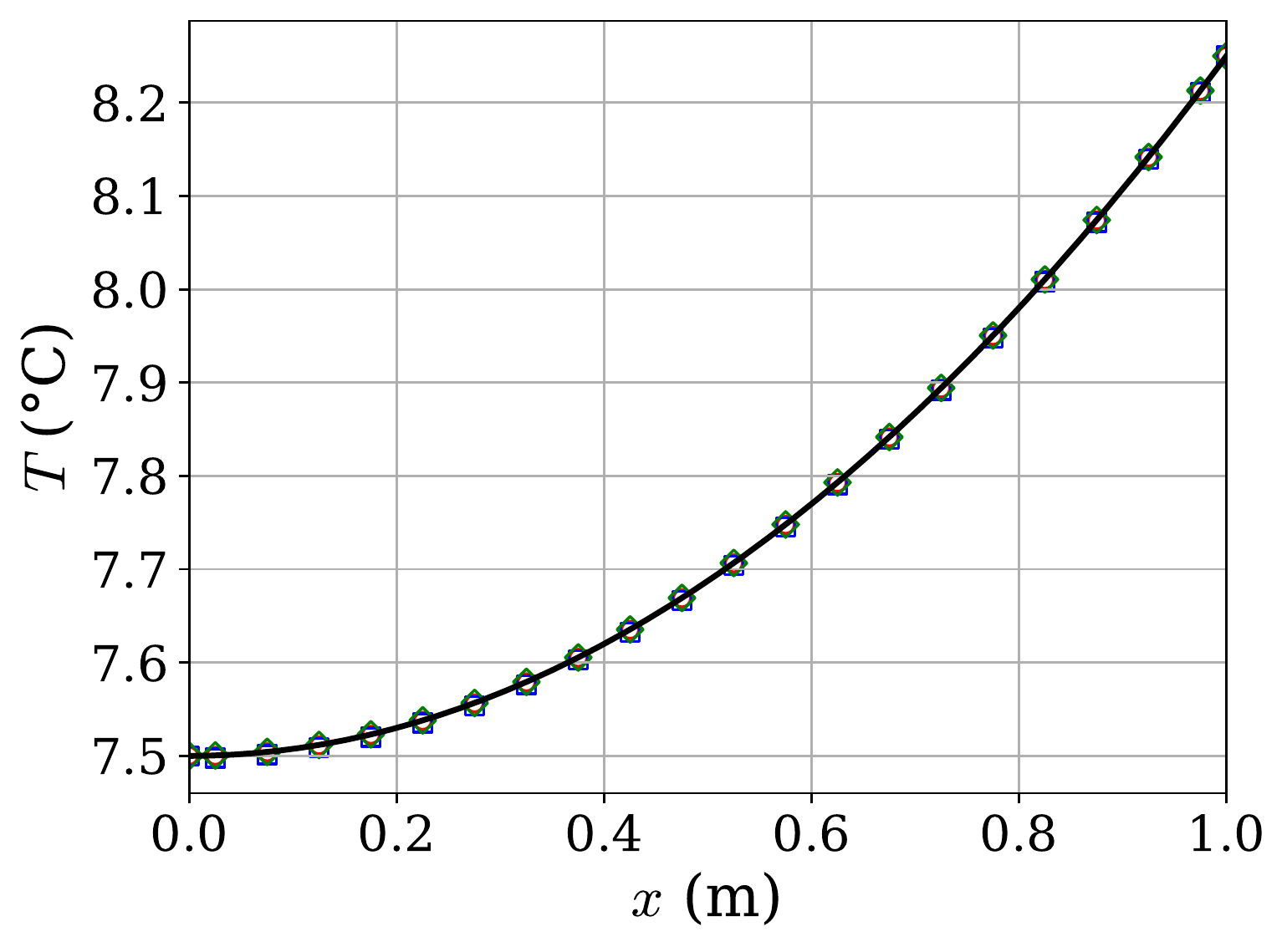}
		\caption{$\alpha = 1.5$, time level $n=5000$.}
		\label{subfig:s0_profile_a1.5_t2}
	\end{subfigure}
	\caption{Solution 0, interpolation: Comparison of relative errors and final temperature profiles for $\alpha=0.7,1.5$ (\blackline\ Exact, \raisebox{0.5pt}{\textcolor{red}{$\circ$}} PBM, \textcolor{blue}{$\square$} DDM, \raisebox{0.5pt}{\textcolor{green}{$\diamond$}} HAM). HAM's predictions are by far the most accurate, followed by PBM, while DDM is least accurate.}
	\label{fig:s0_interp} 
\end{figure}

\begin{figure}
	\begin{subfigure}[b]{0.5\linewidth}
		\centering 
		\includegraphics[width=\textwidth]{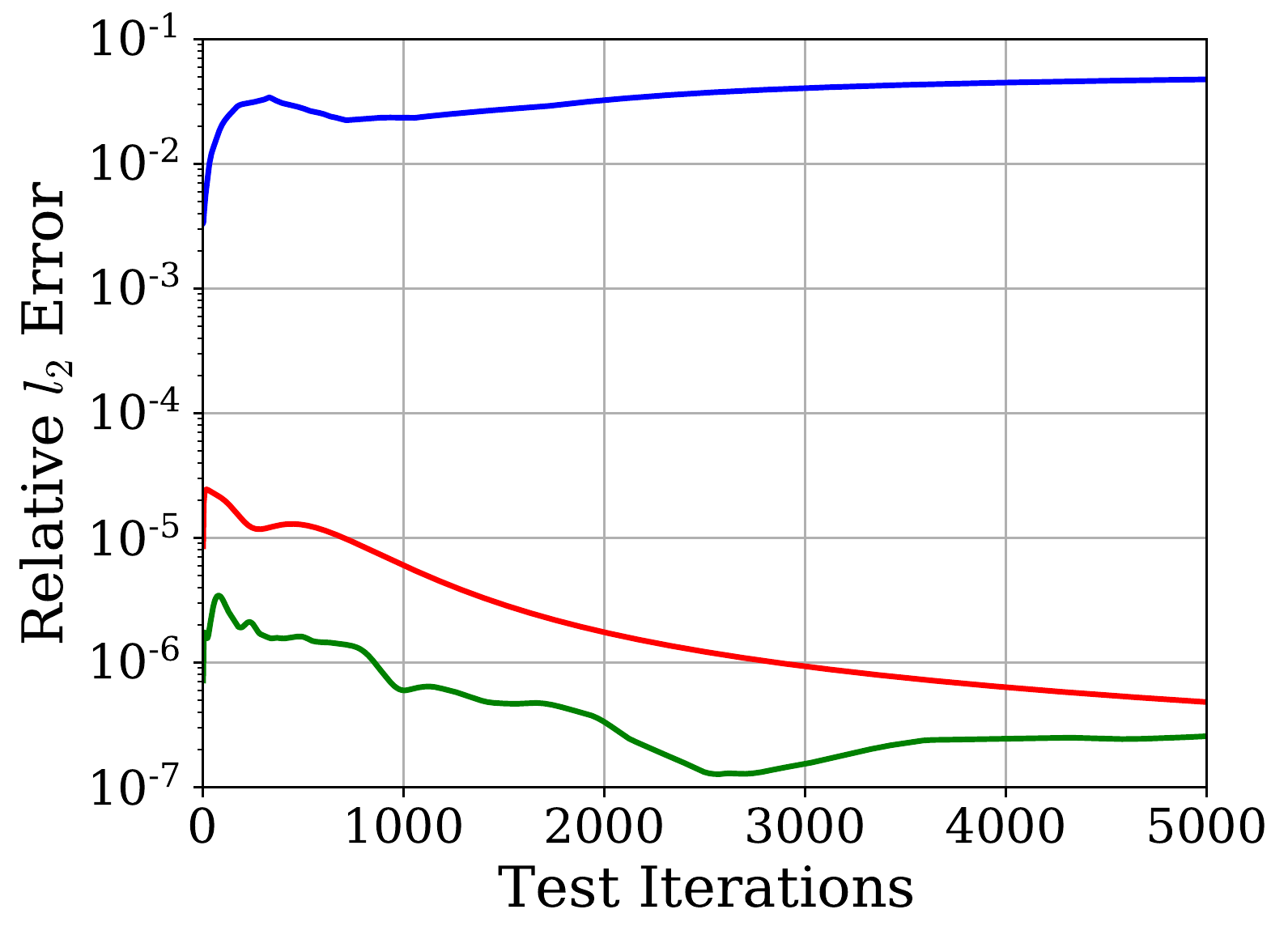}
		\caption{$\alpha = 0.7$, relative errors.}
		\label{subfig:s6_finer_error_a0.7}
	\end{subfigure}%
	\begin{subfigure}[b]{0.5\linewidth}
		\centering 
		\includegraphics[width=\textwidth]{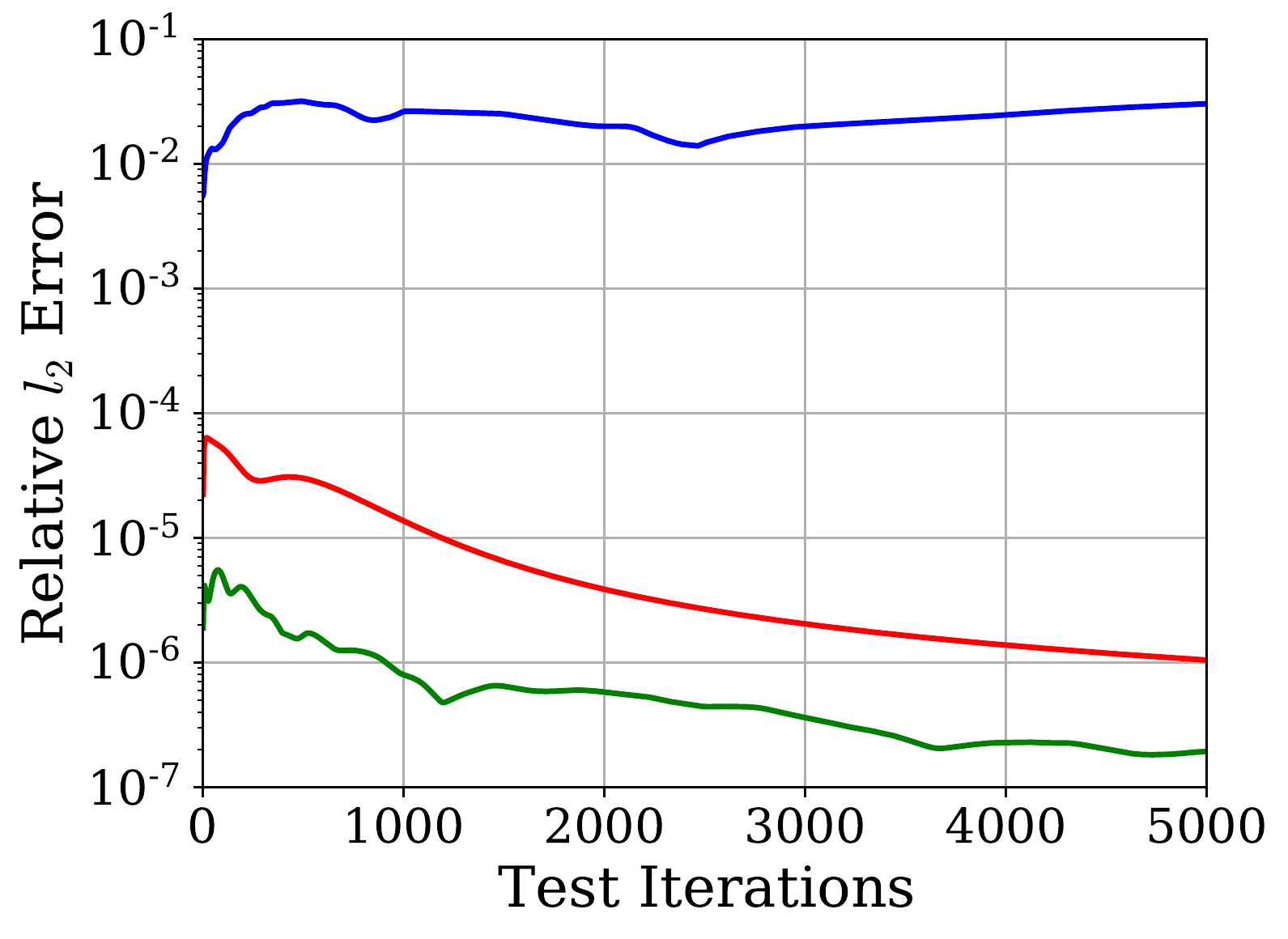}
		\caption{$\alpha = 1.5$, relative errors.}
		\label{subfig:s6_finer_error_a1.5}
	\end{subfigure}%
	\\
	\begin{subfigure}[b]{0.5\linewidth}
		\centering 
		\includegraphics[width=\textwidth]{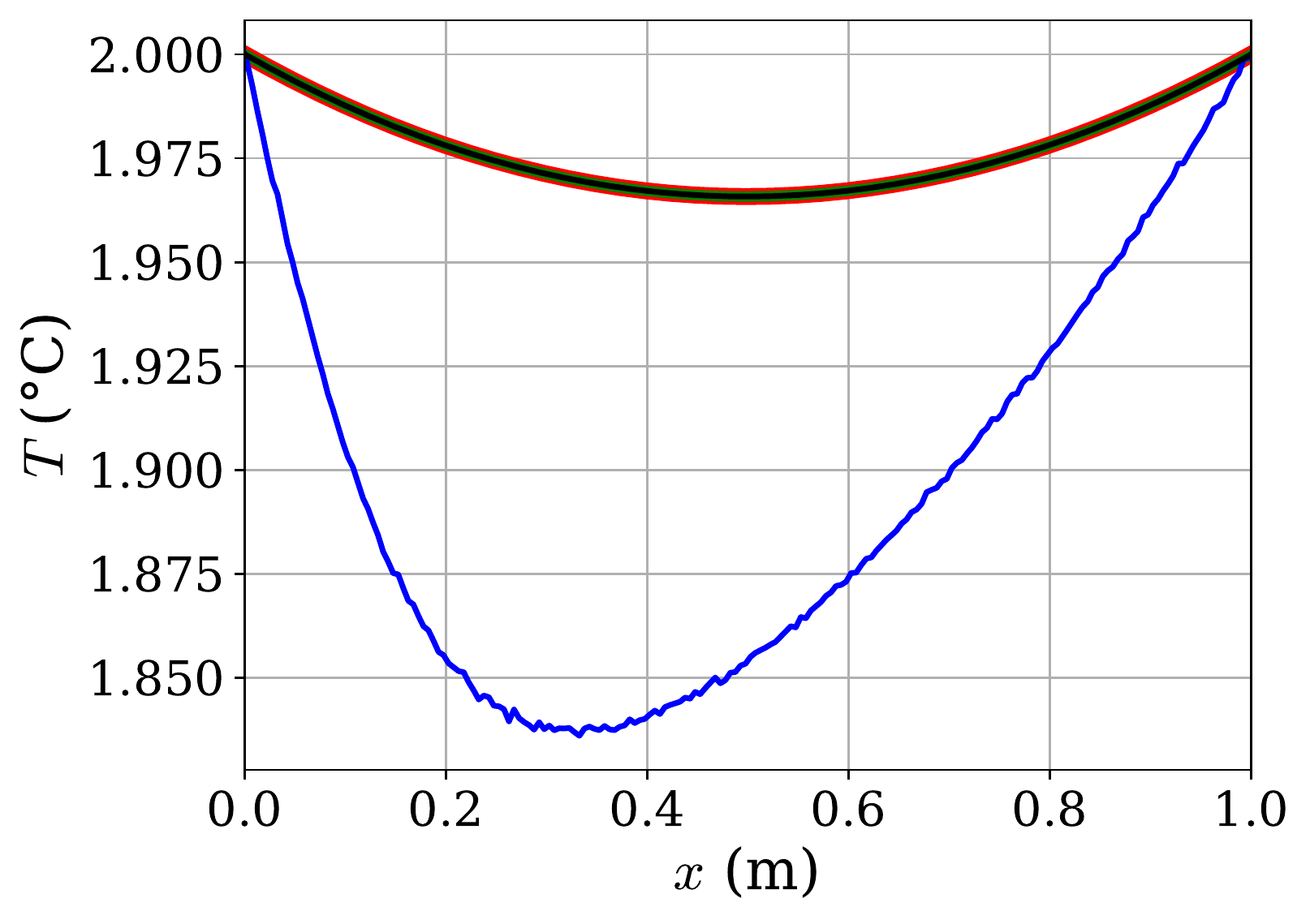}
		\caption{$\alpha = 0.7$, time level $n=5000$.}
		\label{subfig:s6_finer_profile_a0.7_t2}
	\end{subfigure}%
	\begin{subfigure}[b]{0.5\linewidth}
		\centering 
		\includegraphics[width=\textwidth]{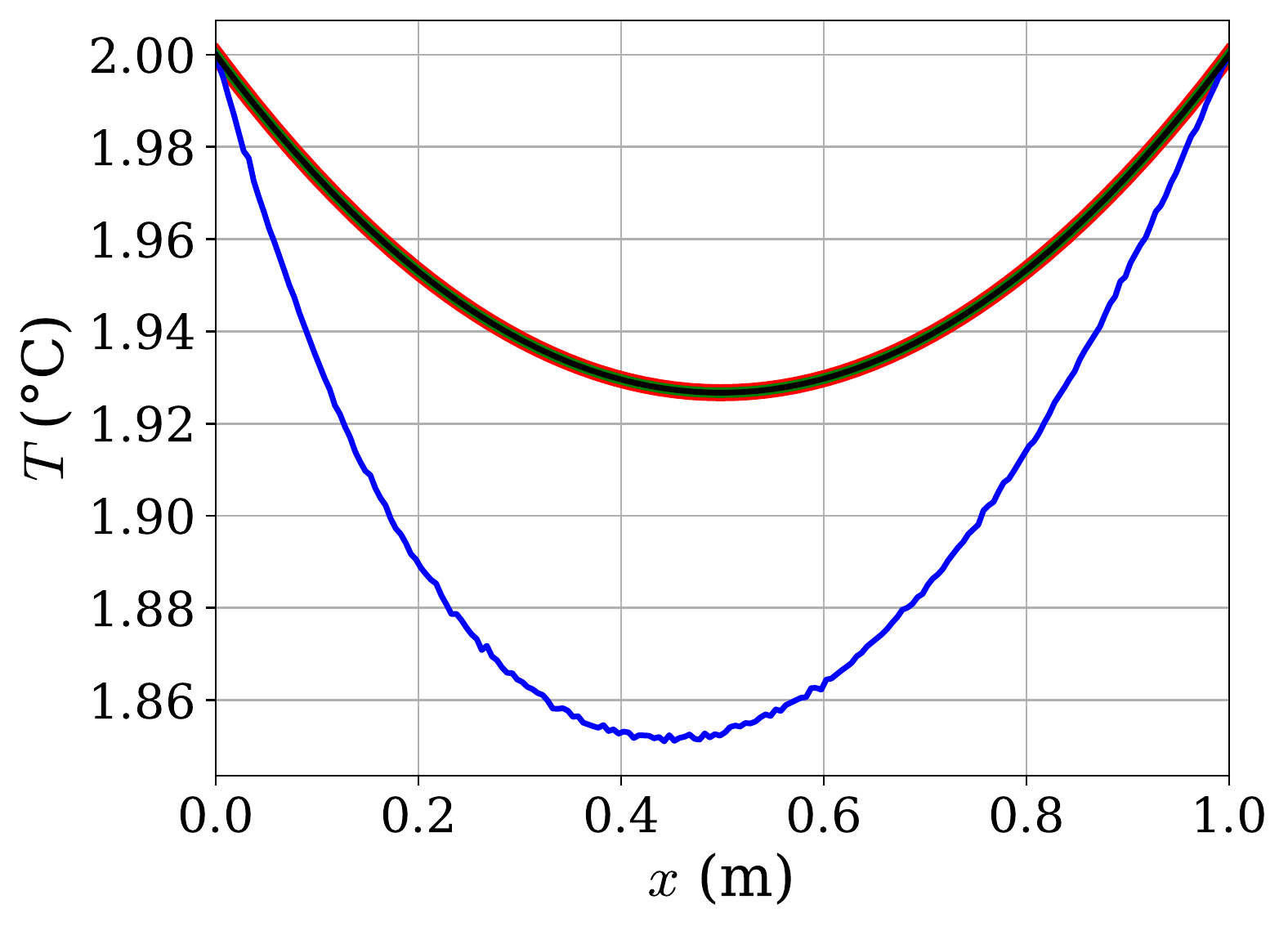}
		\caption{$\alpha = 1.5$, time level $n=5000$.}
		\label{subfig:s6_finer_profile_a1.5_t2}
	\end{subfigure}
	\caption{Solution 3 with fine grid, interpolation: Comparison of relative errors and final temperature profiles for $\alpha=0.7,1.5$ (\blackline\ Exact, \redline\ PBM, \blueline\ DDM, \greenline\ HAM). HAM improves on an already highly accurate PBM, while DDM is inaccurate.}
	\label{fig:s6_finer_interp} 
\end{figure}

Figures~\ref{fig:s0_interp} and~\ref{fig:s6_finer_interp} present the results for the two interpolation cases corresponding to Solutions~0 and~3. From the figures, it is clear that the predictions of the PBM are better than those of the DDM in both the cases. The errors associated with the PBM can be fully attributed to the discretization error, which is especially small for Solution~3 where a finer spatial grid was used. For both solutions, HAM improves on the accuracy of PBM by compensating for the discretization error, even when the discretization error is small.    

\subsubsection{Extrapolation scenarios}
\begin{figure}
	\begin{subfigure}[b]{0.5\linewidth}
		\centering 
		\includegraphics[width=\textwidth]{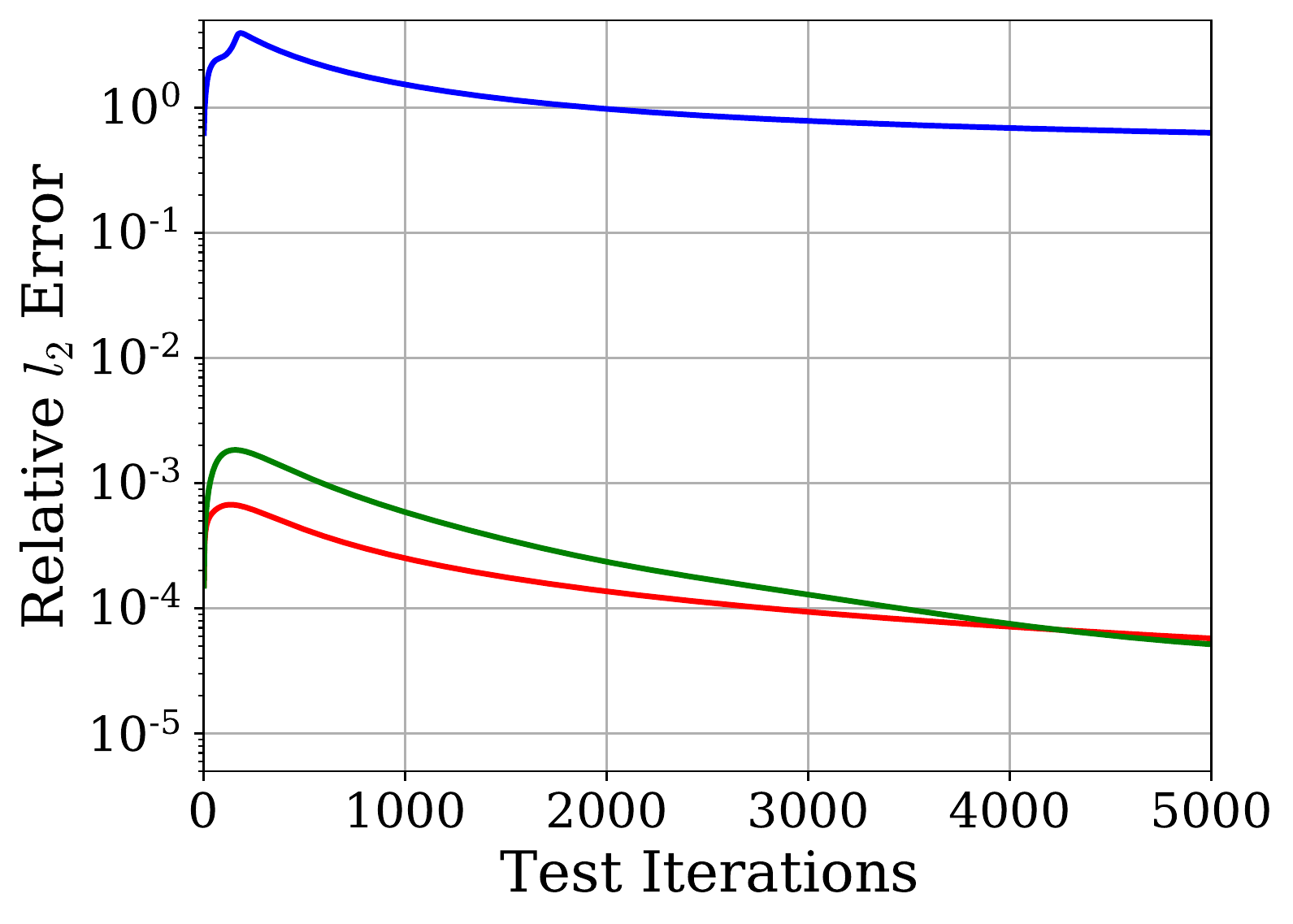}
		\caption{$\alpha = -0.5$, relative errors.}
		\label{subfig:s0_error_a-0.5}
	\end{subfigure}%
	\begin{subfigure}[b]{0.5\linewidth}
		\centering 
		\includegraphics[width=\textwidth]{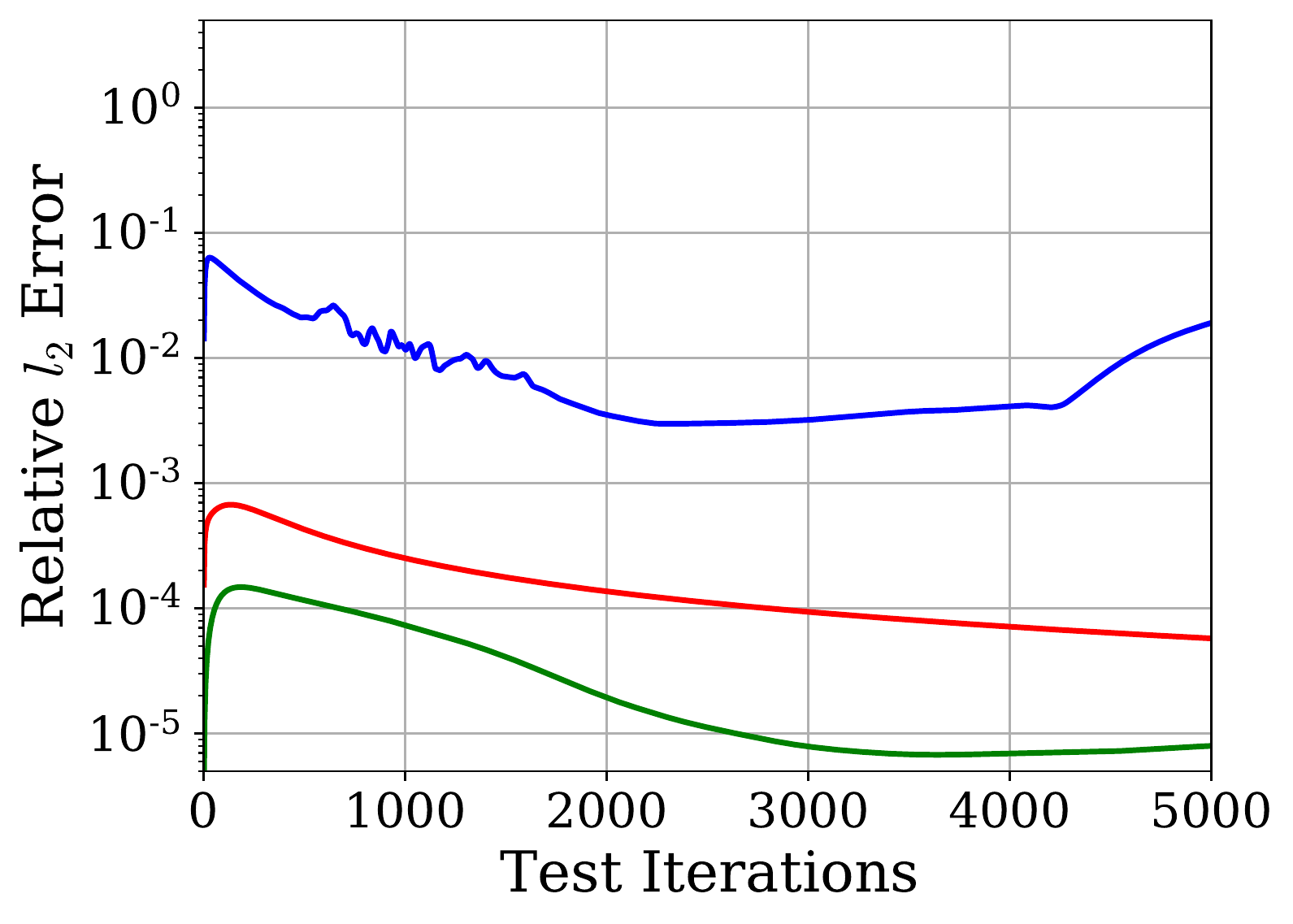}
		\caption{$\alpha = 2.5$, relative errors.}
		\label{subfig:s0_error_a2.5}
	\end{subfigure}%
	\\
	\begin{subfigure}[b]{0.5\linewidth}
		\centering 
		\includegraphics[width=\textwidth]{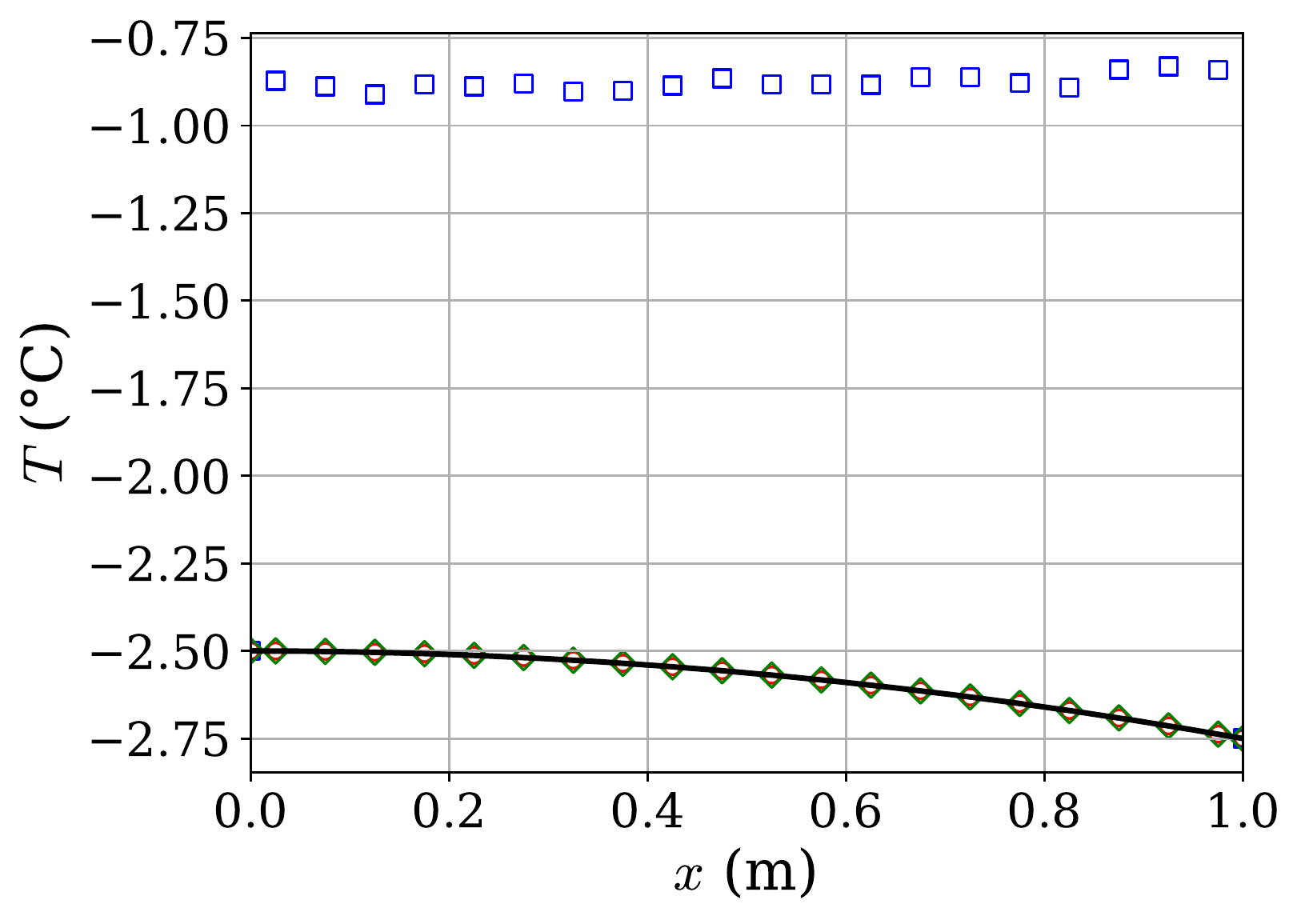}
		\caption{$\alpha = -0.5$, time level $n=5000$.}
		\label{subfig:s0_profile_a-0.5_t2}
	\end{subfigure}%
	\begin{subfigure}[b]{0.5\linewidth}
		\centering 
		\includegraphics[width=\textwidth]{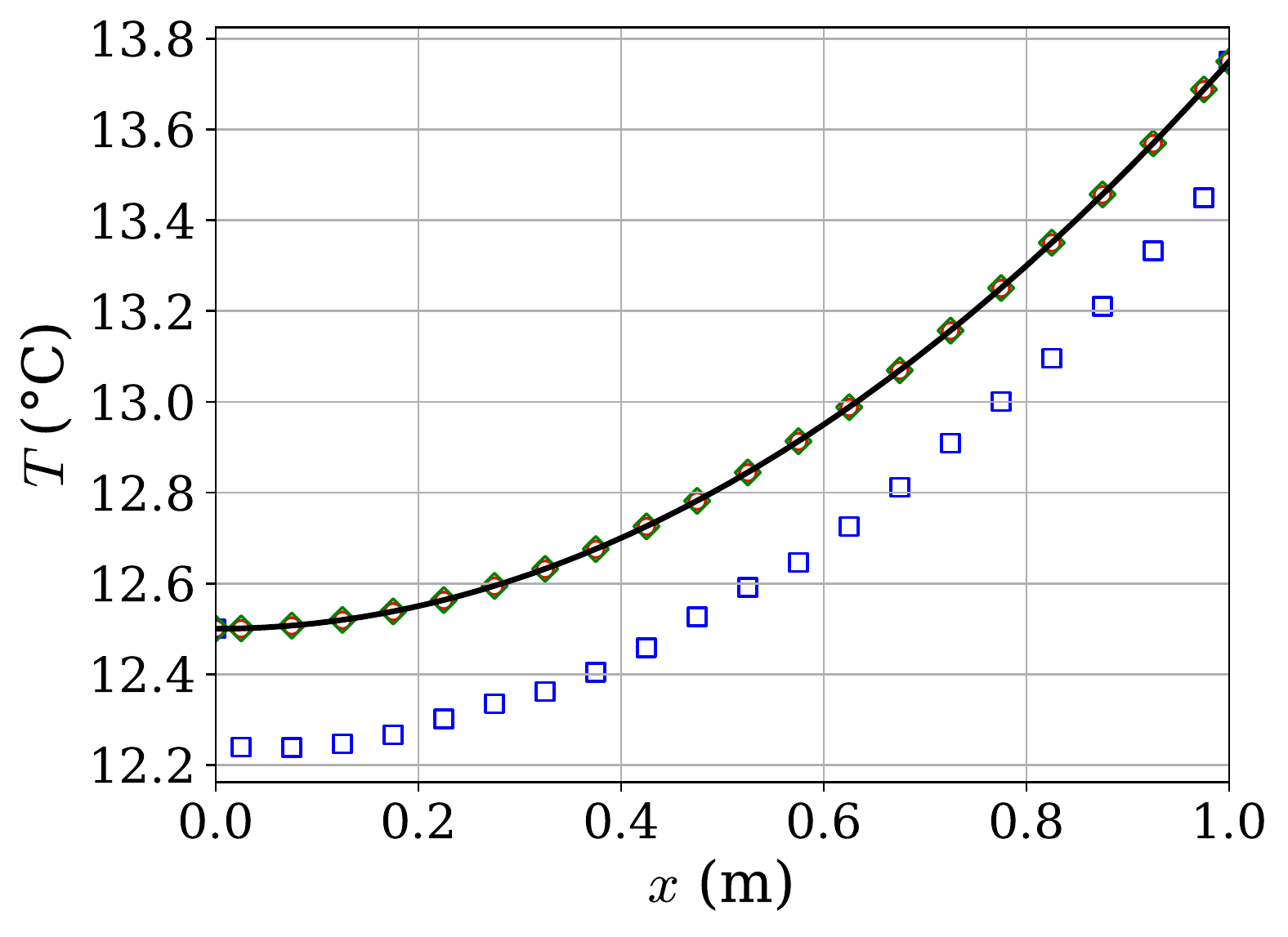}
		\caption{$\alpha = 2.5$, time level $n=5000$.}
		\label{subfig:s0_profile_a2.5_t2}
	\end{subfigure}
	\caption{Solution 0, extrapolation: Comparison of relative errors and final temperature profiles for $\alpha=-0.5,2.5$ (\blackline\ Exact, \raisebox{0.5pt}{\textcolor{red}{$\circ$}} PBM, \textcolor{blue}{$\square$} DDM, \raisebox{0.5pt}{\textcolor{green}{$\diamond$}} HAM). Both HAM and PBM outperform DDM. HAM improves on PBM for $\alpha=2.5$, which is the scenario most similar to the training scenarios, but diminishes accuracy for $\alpha=-0.5$.}
	\label{fig:s0_extrap} 
\end{figure}

\begin{figure}
	\begin{subfigure}[b]{0.5\linewidth}
		\centering 
		\includegraphics[width=\textwidth]{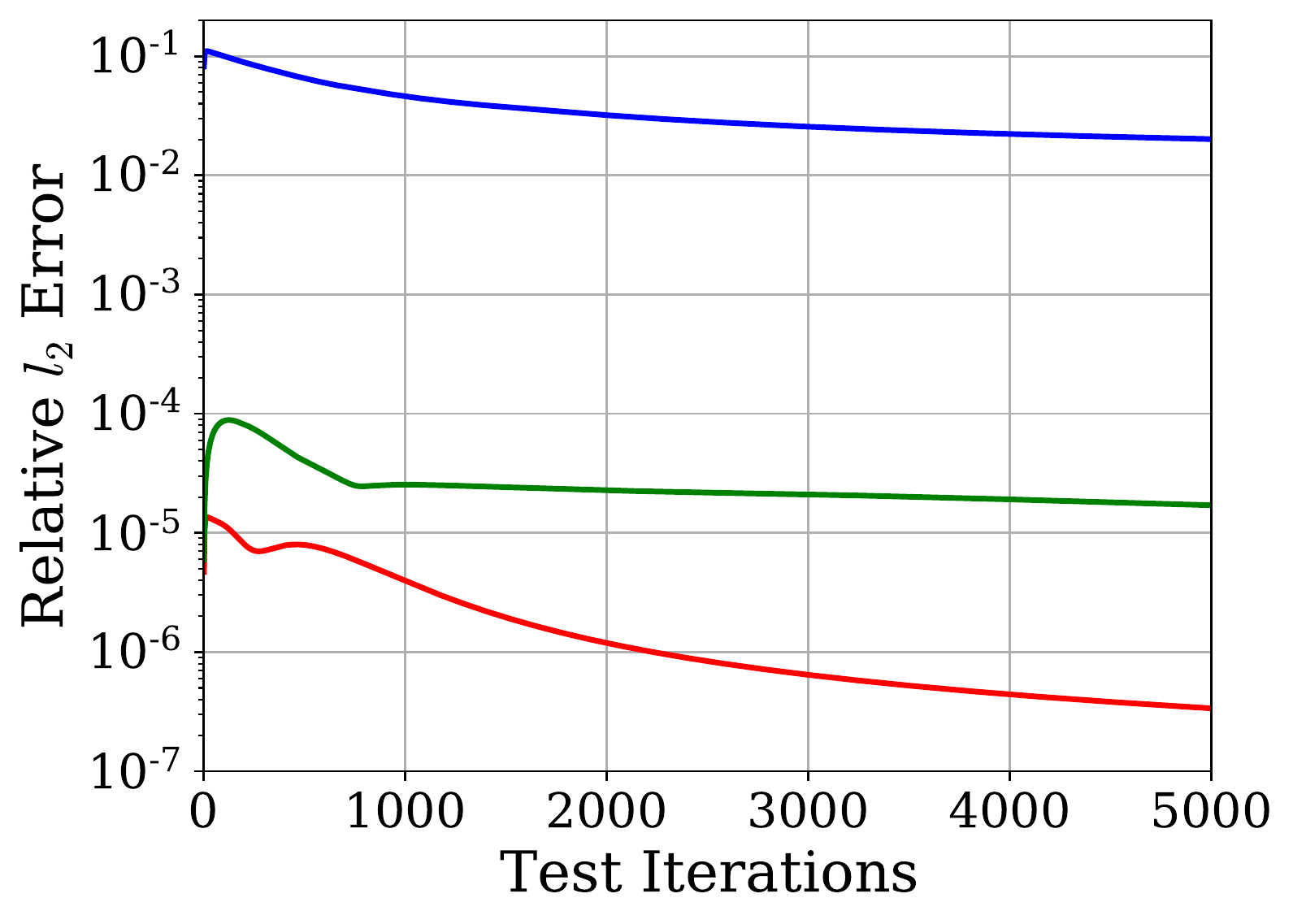}
		\caption{$\alpha = -0.5$, relative errors.}
		\label{subfig:s6_finer_error_a-0.5}
	\end{subfigure}%
	\begin{subfigure}[b]{0.5\linewidth}
		\centering 
		\includegraphics[width=\textwidth]{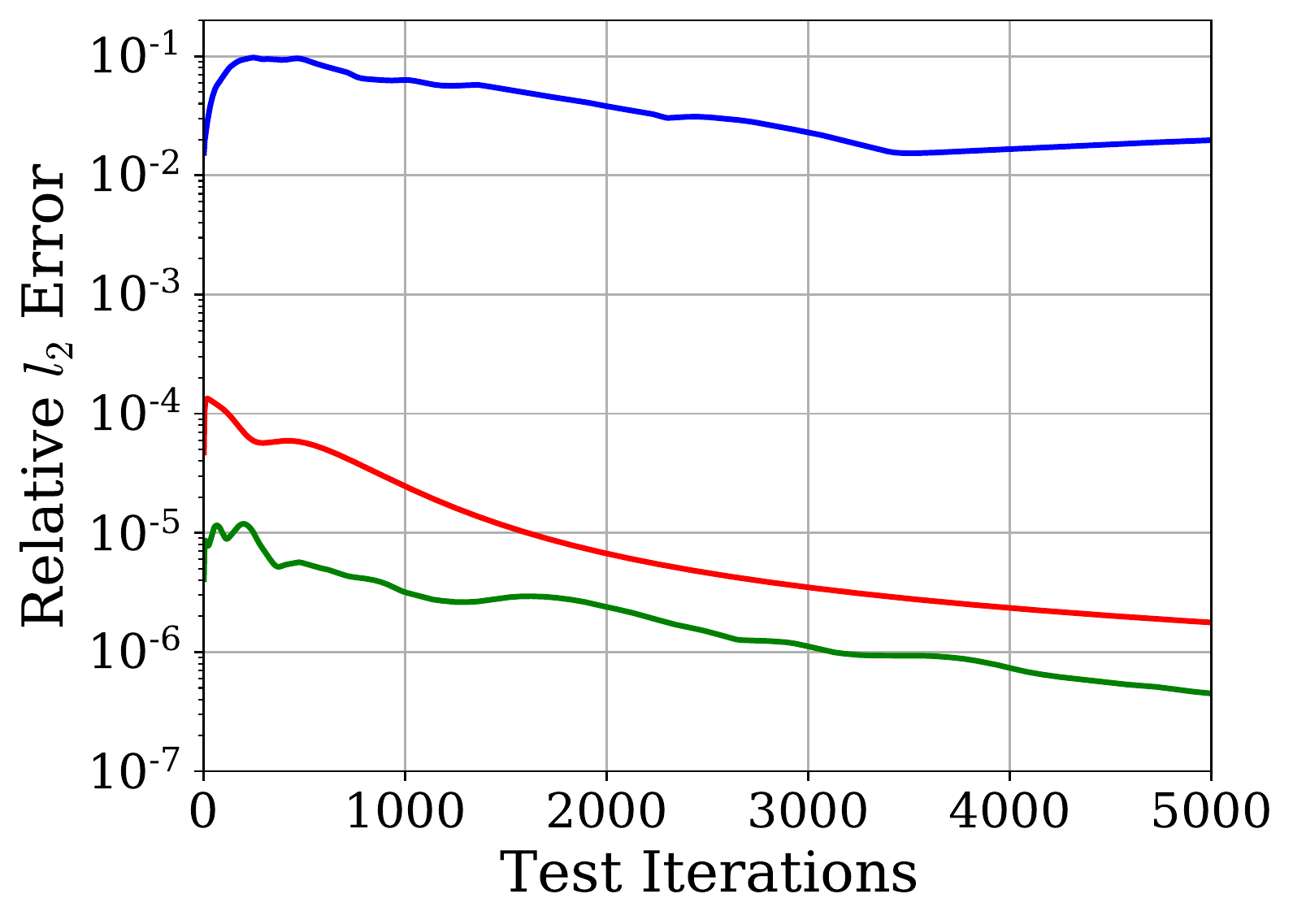}
		\caption{$\alpha = 2.5$, relative errors.}
		\label{subfig:s6_finer_error_a2.5}
	\end{subfigure}%
	\\
	\begin{subfigure}[b]{0.5\linewidth}
		\centering 
		\includegraphics[width=\textwidth]{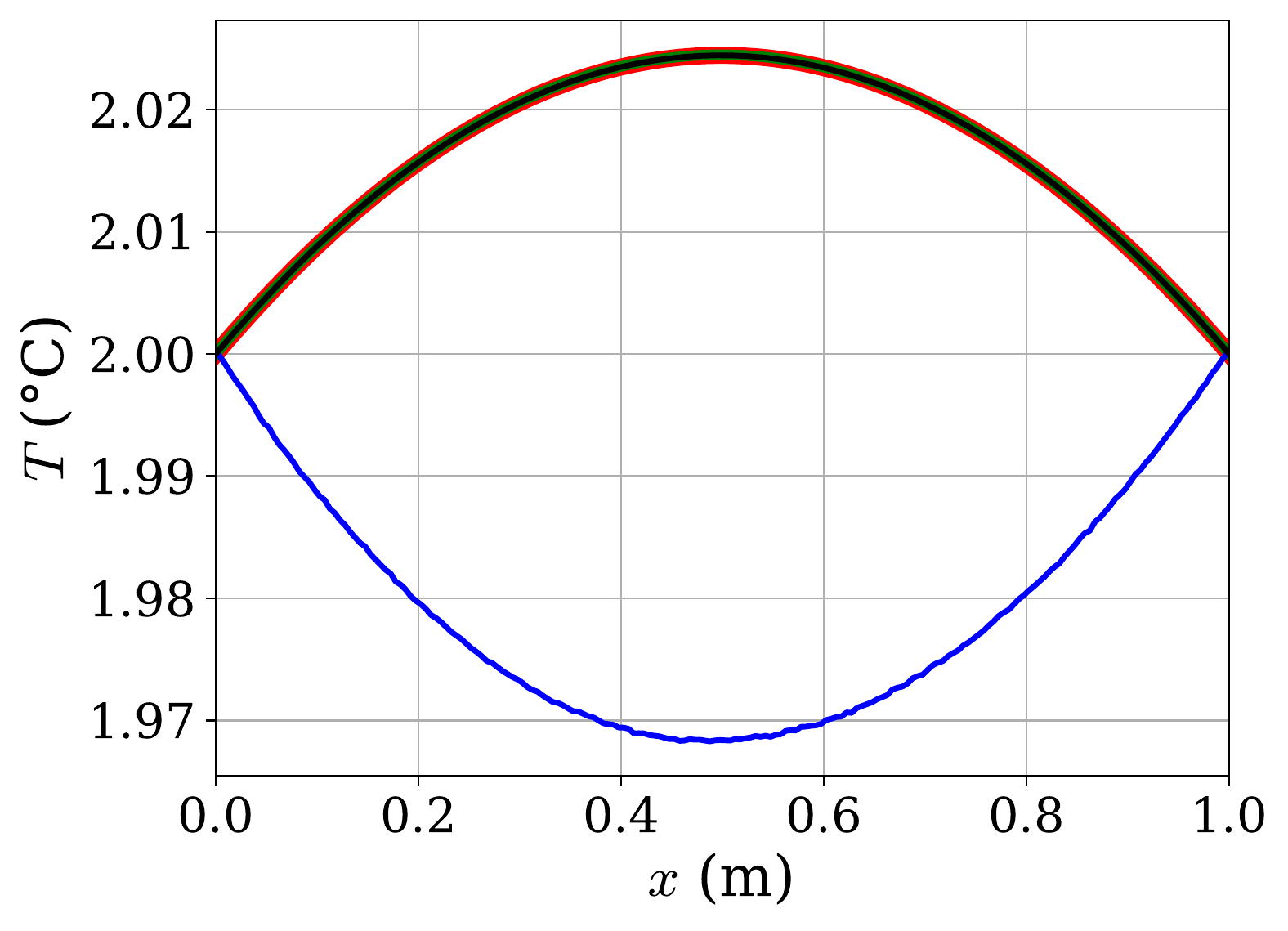}
		\caption{$\alpha = -0.5$, time level $n=5000$.}
		\label{subfig:s6_finer_profile_a-0.5_t2}
	\end{subfigure}%
	\begin{subfigure}[b]{0.5\linewidth}
		\centering 
		\includegraphics[width=\textwidth]{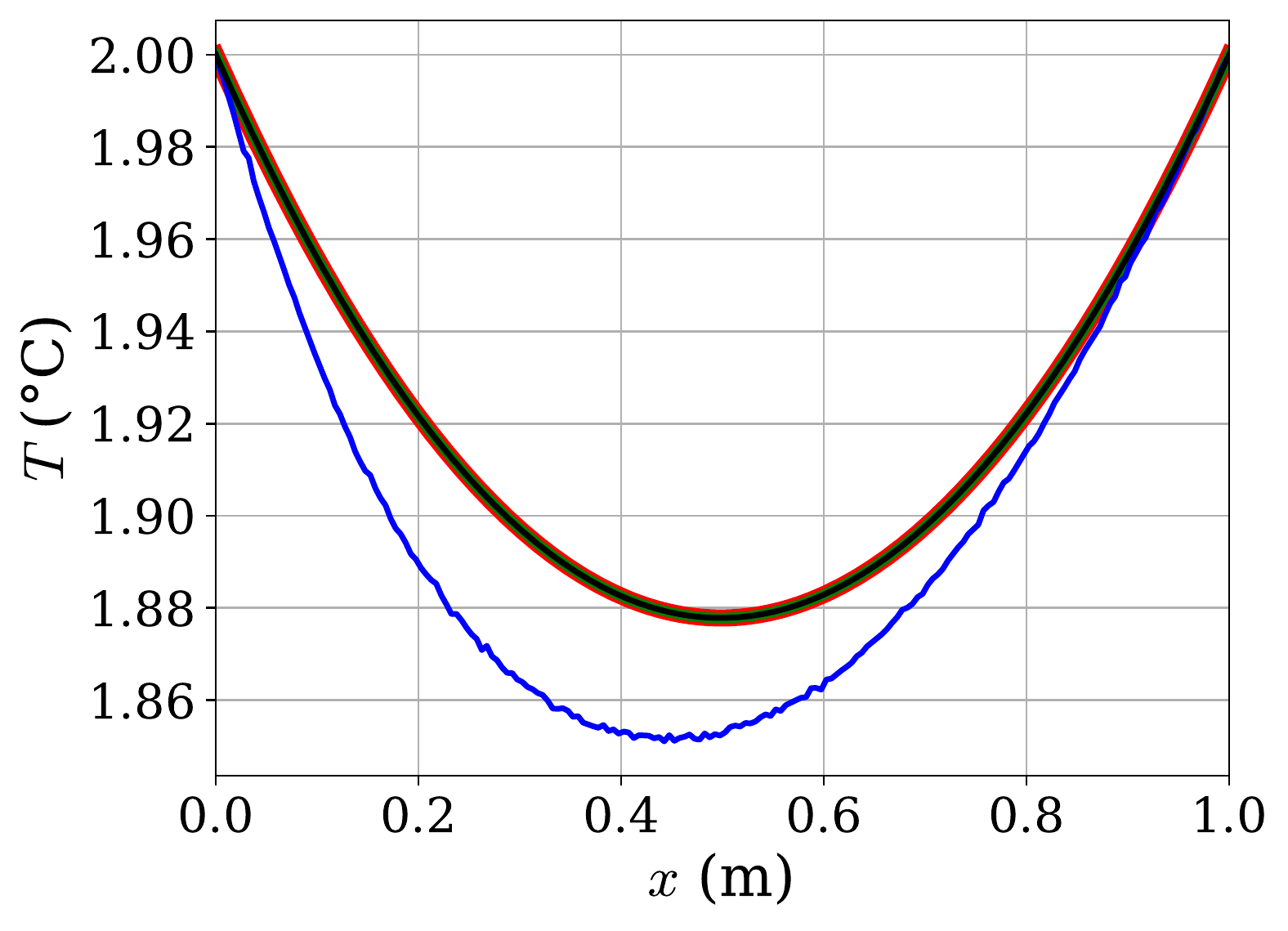}
		\caption{$\alpha = 2.5$, time level $n=5000$.}
		\label{subfig:s6_finer_profile_a2.5_t2}
	\end{subfigure}
	\caption{Solution 3 with fine grid, extrapolation: Comparison of relative errors and final temperature profiles for $\alpha=-0.5,2.5$ (\blackline\ Exact, \redline\ PBM, \blueline\ DDM, \greenline\ HAM). For both $\alpha$, PBM exhibits high accuracy while DDM exhibits comparatively low accuracy. HAM is able to improve on the accuracy of the PBM for the $\alpha=2.5$, but yields reduced accuracy for $\alpha=-0.5$.}
	\label{fig:s6_finer_extrap} 
\end{figure}

Figures \ref{fig:s0_extrap} and \ref{fig:s6_finer_extrap} present the results for the two extrapolation cases corresponding to Solutions~0 and~3. As expected, DDM fails to generalize to both the extrapolation cases, while PBM exhibits similar level of accuracy for both interpolation and extrapolation. HAM generalizes better than DDM, and yields high accuracy in both extrapolation cases. However, HAM does not significantly outperform PBM here.

In conclusion, PBM generalizes better than DDM, while HAM generalizes approximately as well as PBM in the presence of numerical error. This ability of the HAM to correct for the numerical error can be exploited to relax the mesh requirement needed for solving equations at a fixed error tolerance. That is, equations can be solved on a coarse mesh in a computationally efficient manner, and CoSTA-based HAM can be used to correct for the numerical errors resulting from the coarse discretization.   

\subsection{Experiments with modeling error}
\label{subsec:exp_with_mod_error}
In this section we present the results corresponding to manufactured solutions~1--4, given in Table~\ref{tab:manufactured_solutions}. The results corresponding to interpolation are presented first, followed by the results corresponding to extrapolation.  

\subsubsection{Interpolation scenarios}
\label{subsec:interpolationscenarios}
\begin{figure}
	\begin{subfigure}[b]{0.5\linewidth}
		\centering 
		\includegraphics[width=\textwidth]{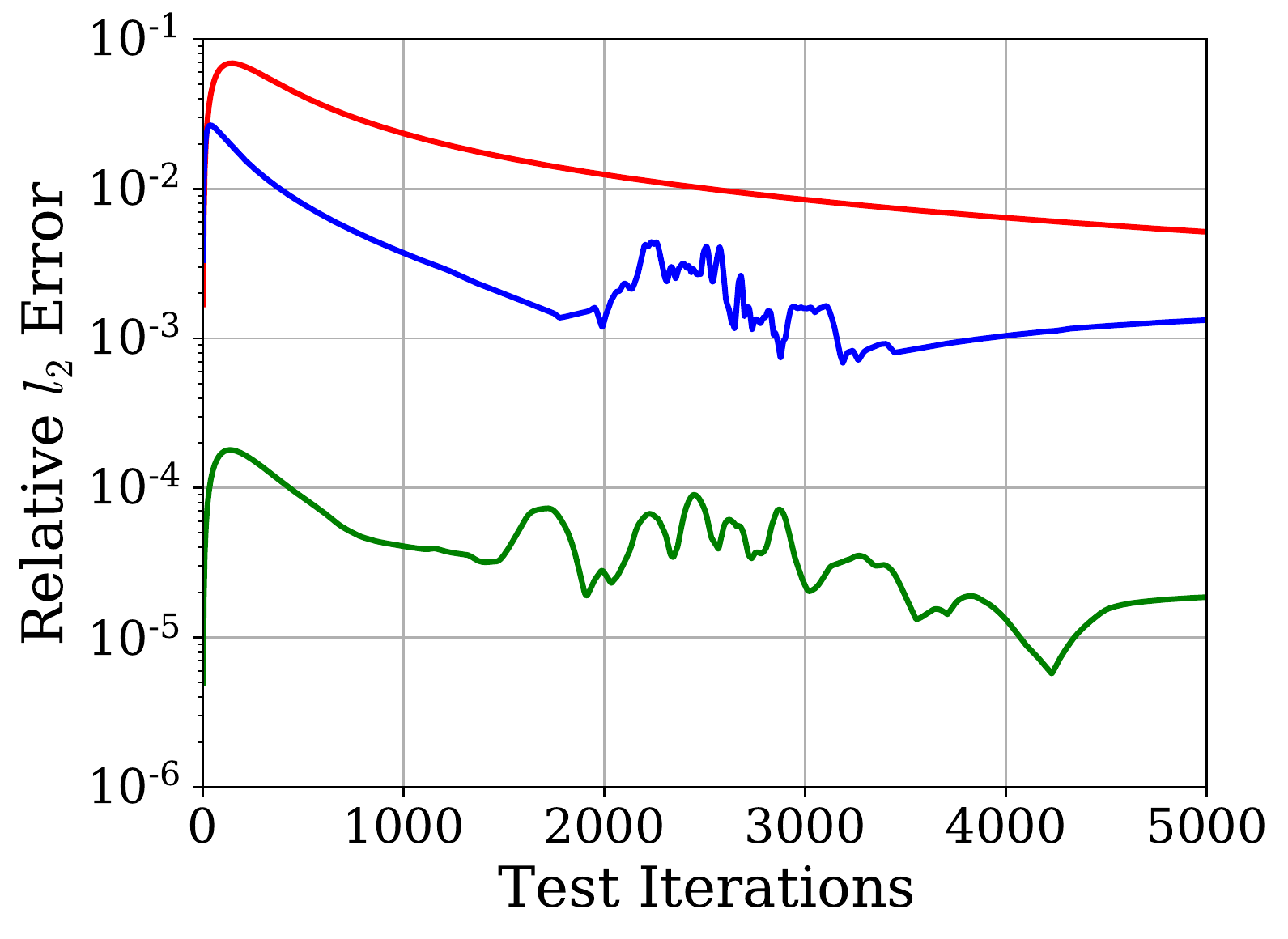}
		\caption{$\alpha = 0.7$, relative errors.}
		\label{subfig:s1_error_a0.7}
	\end{subfigure}%
	\begin{subfigure}[b]{0.5\linewidth}
		\centering 
		\includegraphics[width=\textwidth]{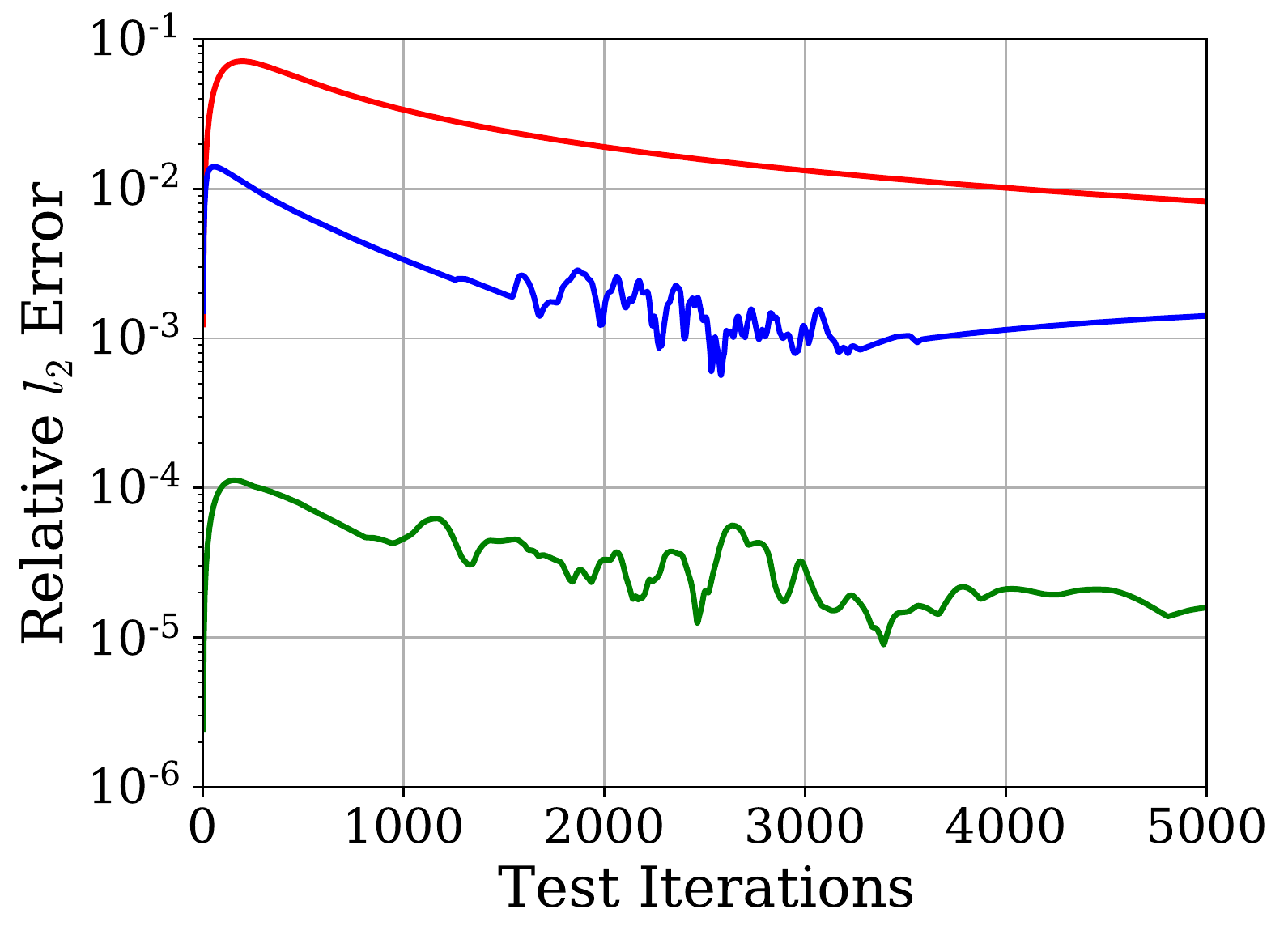}
		\caption{$\alpha = 1.5$, relative errors.}
		\label{subfig:s1_error_a1.5}
	\end{subfigure}%
	\\
	\begin{subfigure}[b]{0.5\linewidth}
		\centering 
		\includegraphics[width=\textwidth]{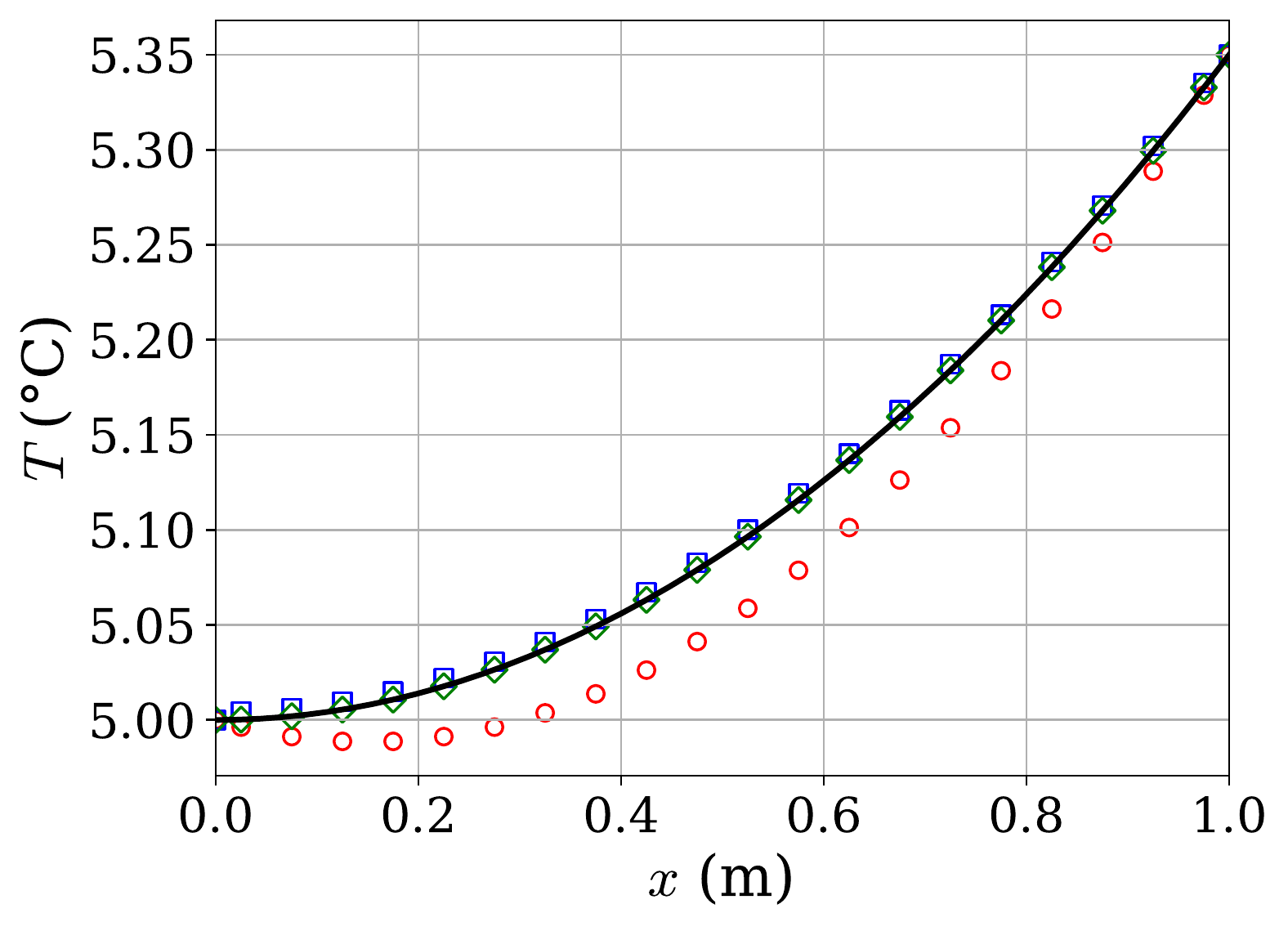}
		\caption{$\alpha = 0.7$, time level $n=5000$.}
		\label{subfig:s1_profile_a0.7_t2}
	\end{subfigure}%
	\begin{subfigure}[b]{0.5\linewidth}
		\centering 
		\includegraphics[width=\textwidth]{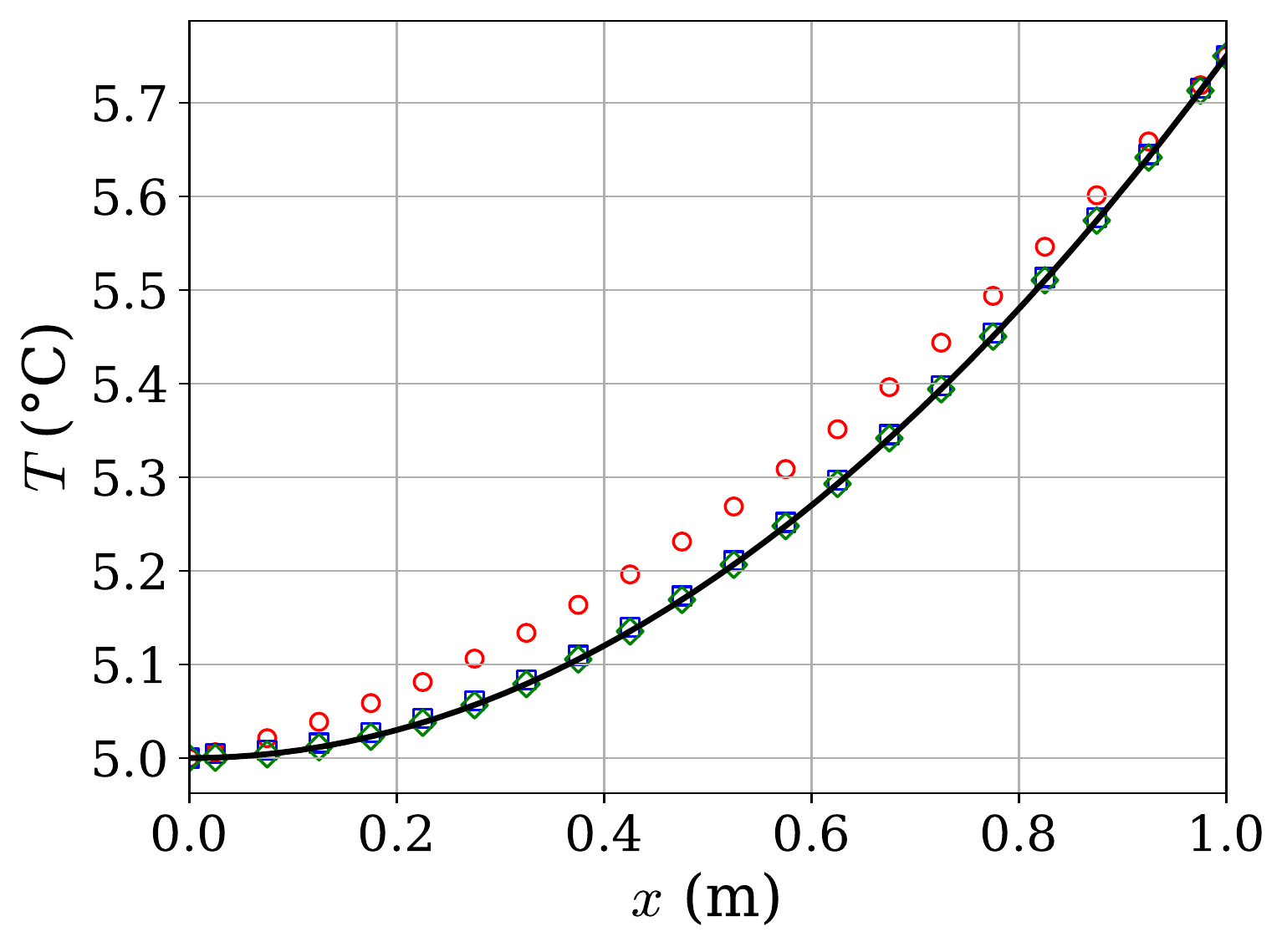}
		\caption{$\alpha = 1.5$, time level $n=5000$.}
		\label{subfig:s1_profile_a1.5_t2}
	\end{subfigure}
	\caption{Solution 1, interpolation: Comparison of relative errors and final temperature profiles for $\alpha=0.7,1.5$ (\blackline\ Exact, \raisebox{0.5pt}{\textcolor{red}{$\circ$}} PBM, \textcolor{blue}{$\square$} DDM, \raisebox{0.5pt}{\textcolor{green}{$\diamond$}} HAM). All predictions are qualitatively correct, but HAM yields superior accuracy.}
	\label{fig:s1_interp} 
\end{figure}

\begin{figure}
	\begin{subfigure}[b]{0.5\linewidth}
		\centering 
		\includegraphics[width=\textwidth]{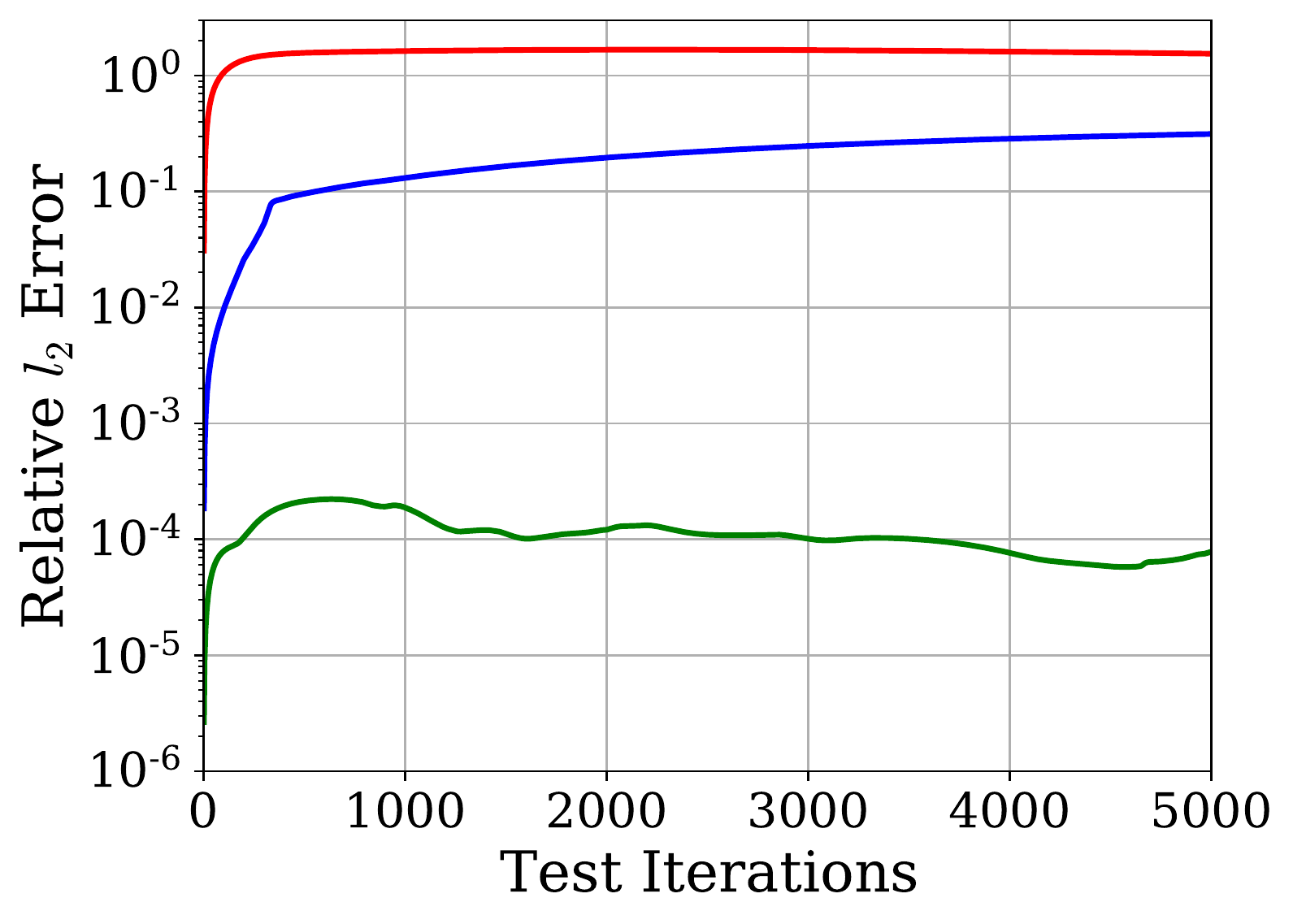}
		\caption{$\alpha = 0.7$, relative errors.}
		\label{subfig:s2_error_a0.7}
	\end{subfigure}%
	\begin{subfigure}[b]{0.5\linewidth}
		\centering 
		\includegraphics[width=\textwidth]{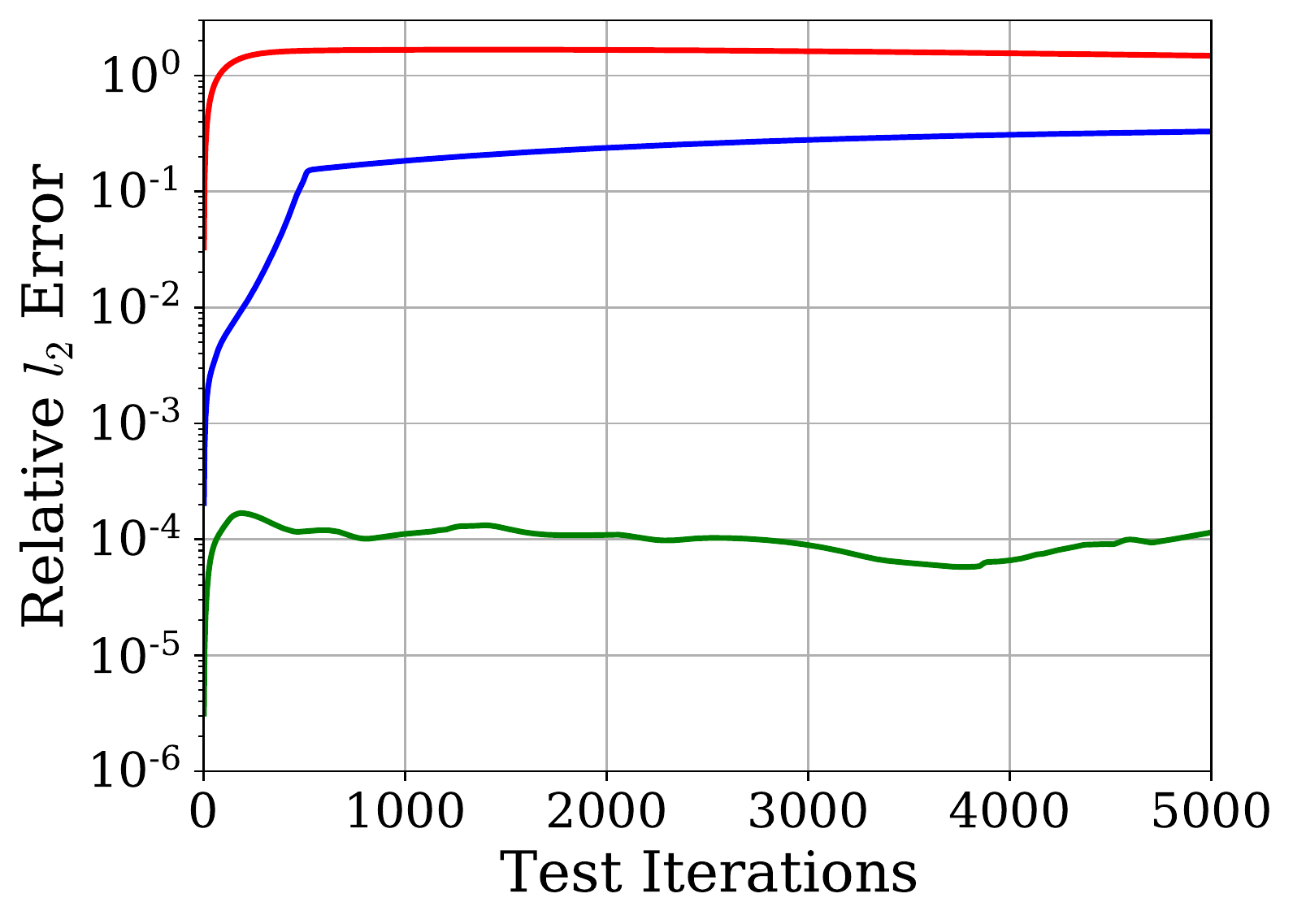}
		\caption{$\alpha = 1.5$, relative errors.}
		\label{subfig:s2_error_a1.5}
	\end{subfigure}%
	\\
	\begin{subfigure}[b]{0.5\linewidth}
		\centering 
		\includegraphics[width=\textwidth]{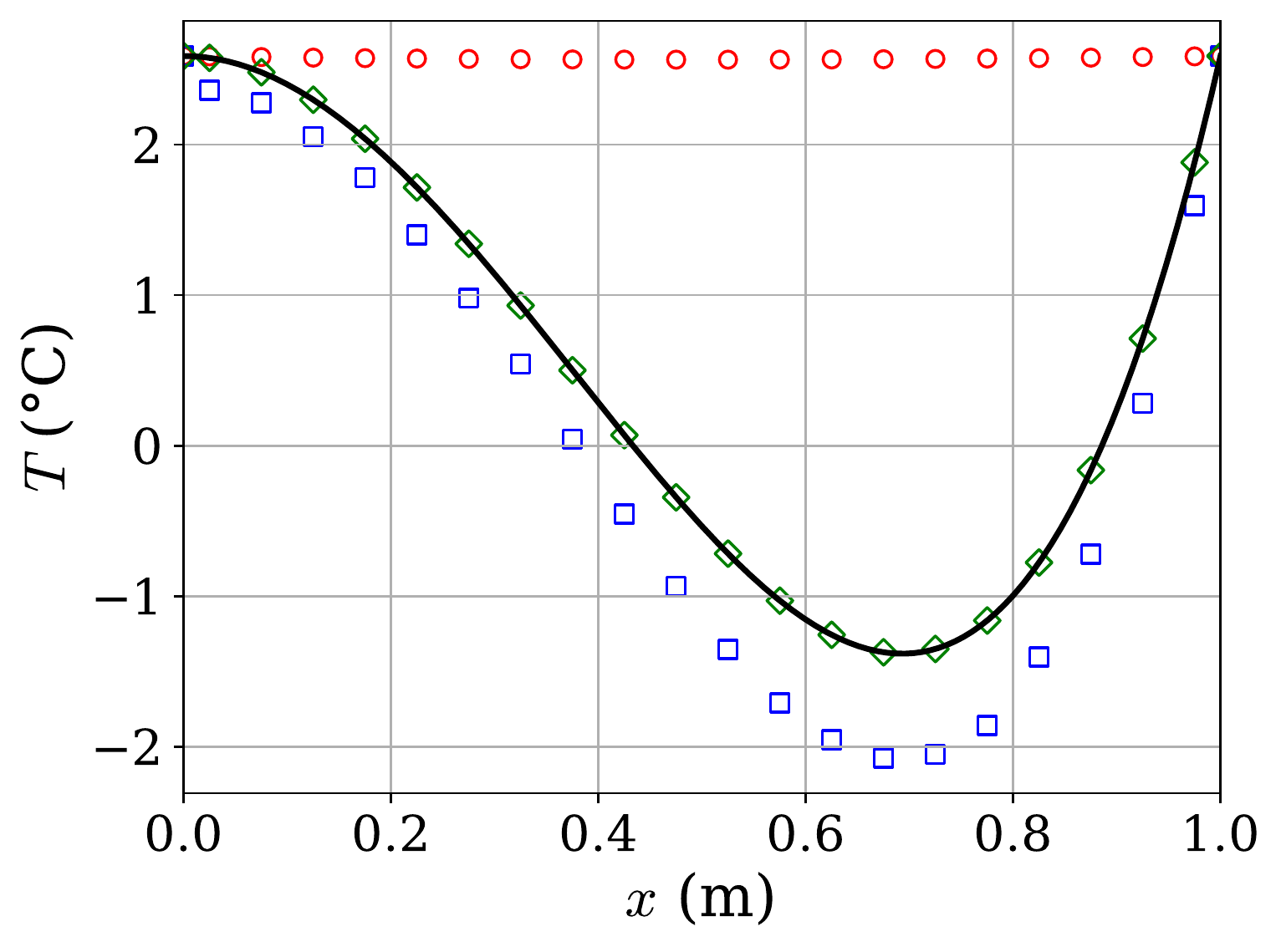}
		\caption{$\alpha = 0.7$, time level $n=5000$.}
		\label{subfig:s2_profile_a0.7_t2}
	\end{subfigure}%
	\begin{subfigure}[b]{0.5\linewidth}
		\centering 
		\includegraphics[width=\textwidth]{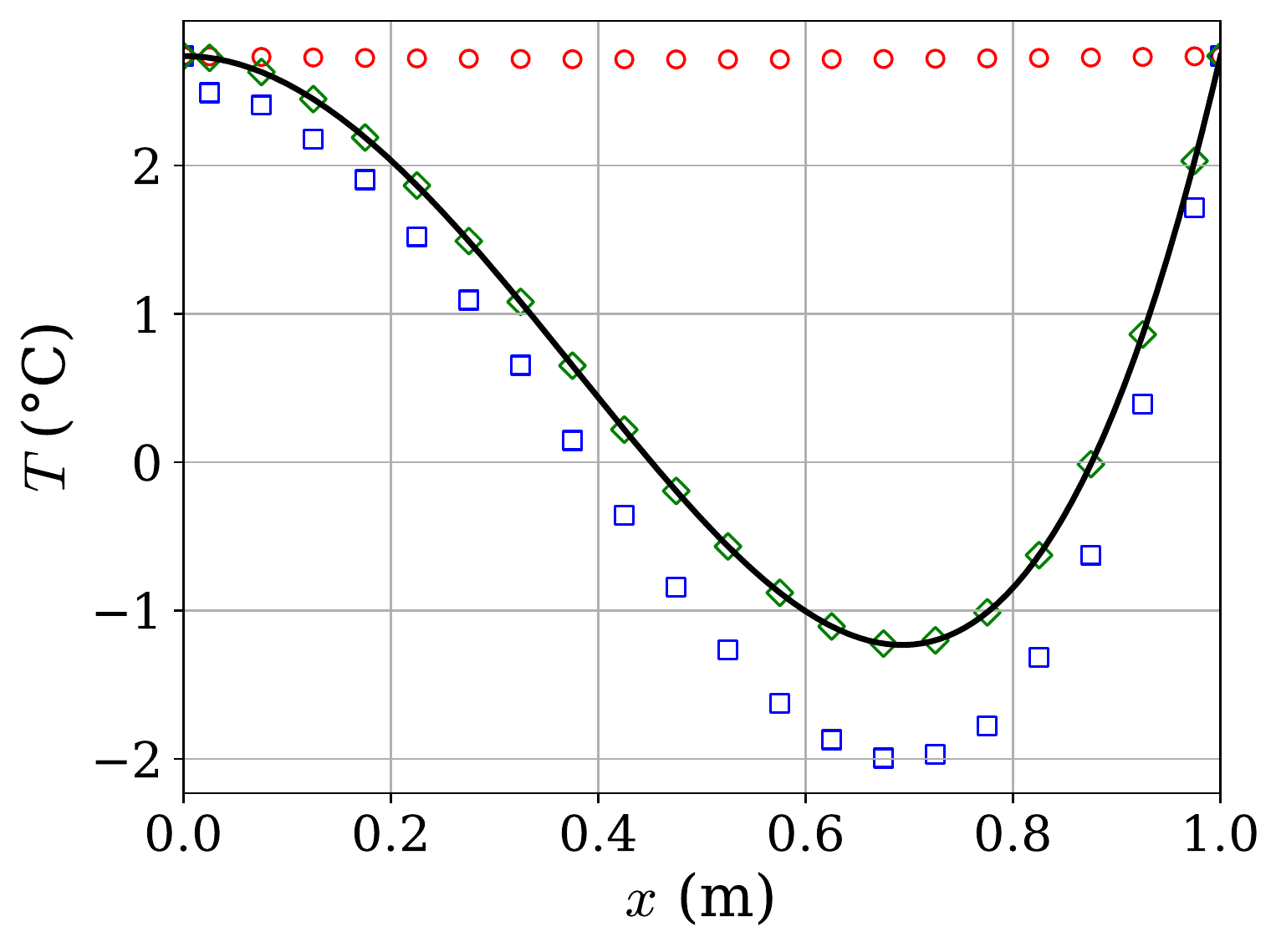}
		\caption{$\alpha = 1.5$, time level $n=5000$.}
		\label{subfig:s2_profile_a1.5_t2}
	\end{subfigure}
	\caption{Solution 2, interpolation: Comparison of relative errors and final temperature profiles for $\alpha=0.7,1.5$ (\blackline\ Exact, \raisebox{0.5pt}{\textcolor{red}{$\circ$}} PBM, \textcolor{blue}{$\square$} DDM, \raisebox{0.5pt}{\textcolor{green}{$\diamond$}} HAM). Only HAM yields predictions that are qualitatively correct and of reasonable accuracy.}
	\label{fig:s2_interp} 
\end{figure}

\begin{figure}
	\begin{subfigure}[b]{0.5\linewidth}
		\centering 
		\includegraphics[width=\textwidth]{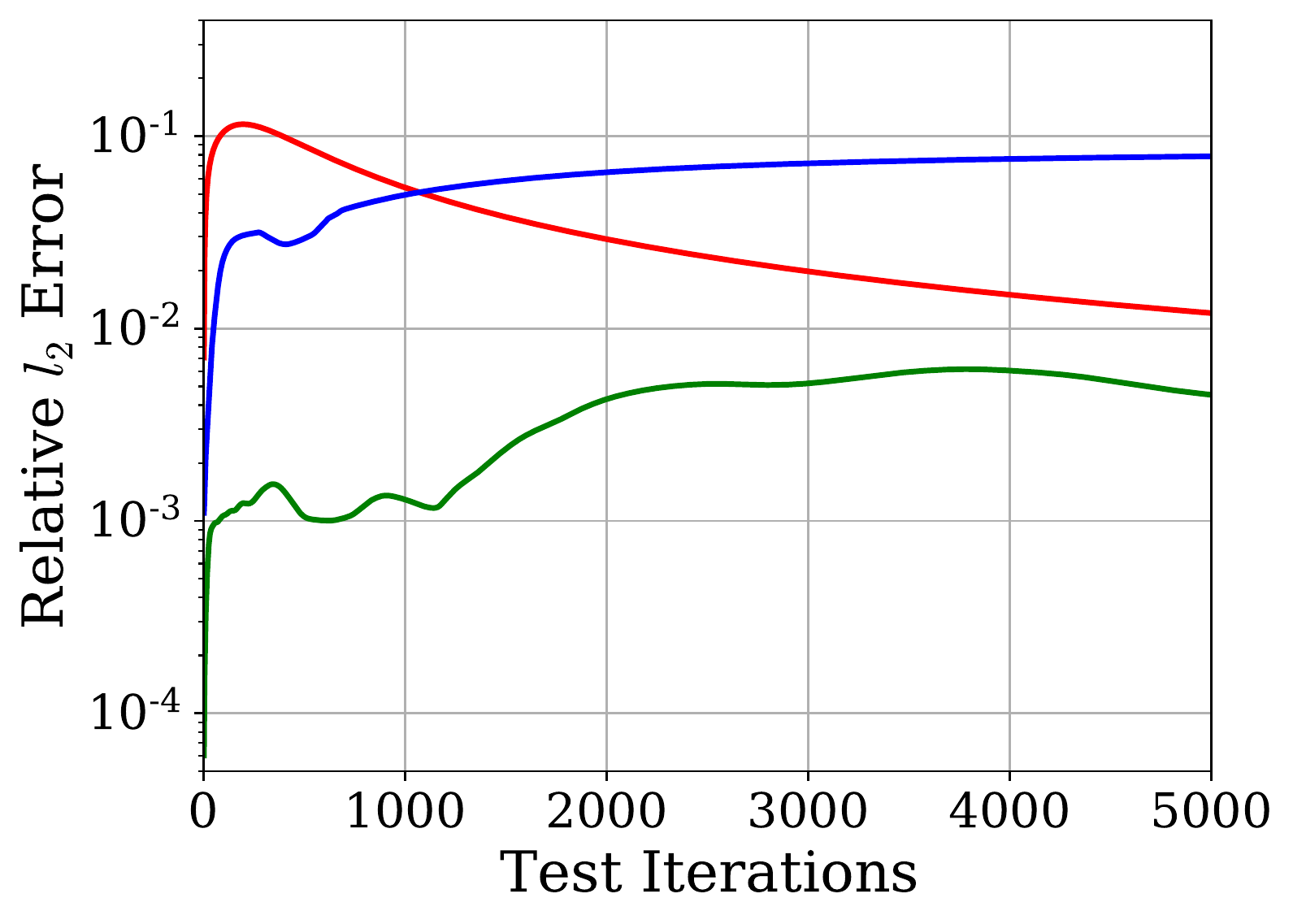}
		\caption{$\alpha = 0.7$, relative errors.}
		\label{subfig:s6_error_a0.7}
	\end{subfigure}%
	\begin{subfigure}[b]{0.5\linewidth}
		\centering 
		\includegraphics[width=\textwidth]{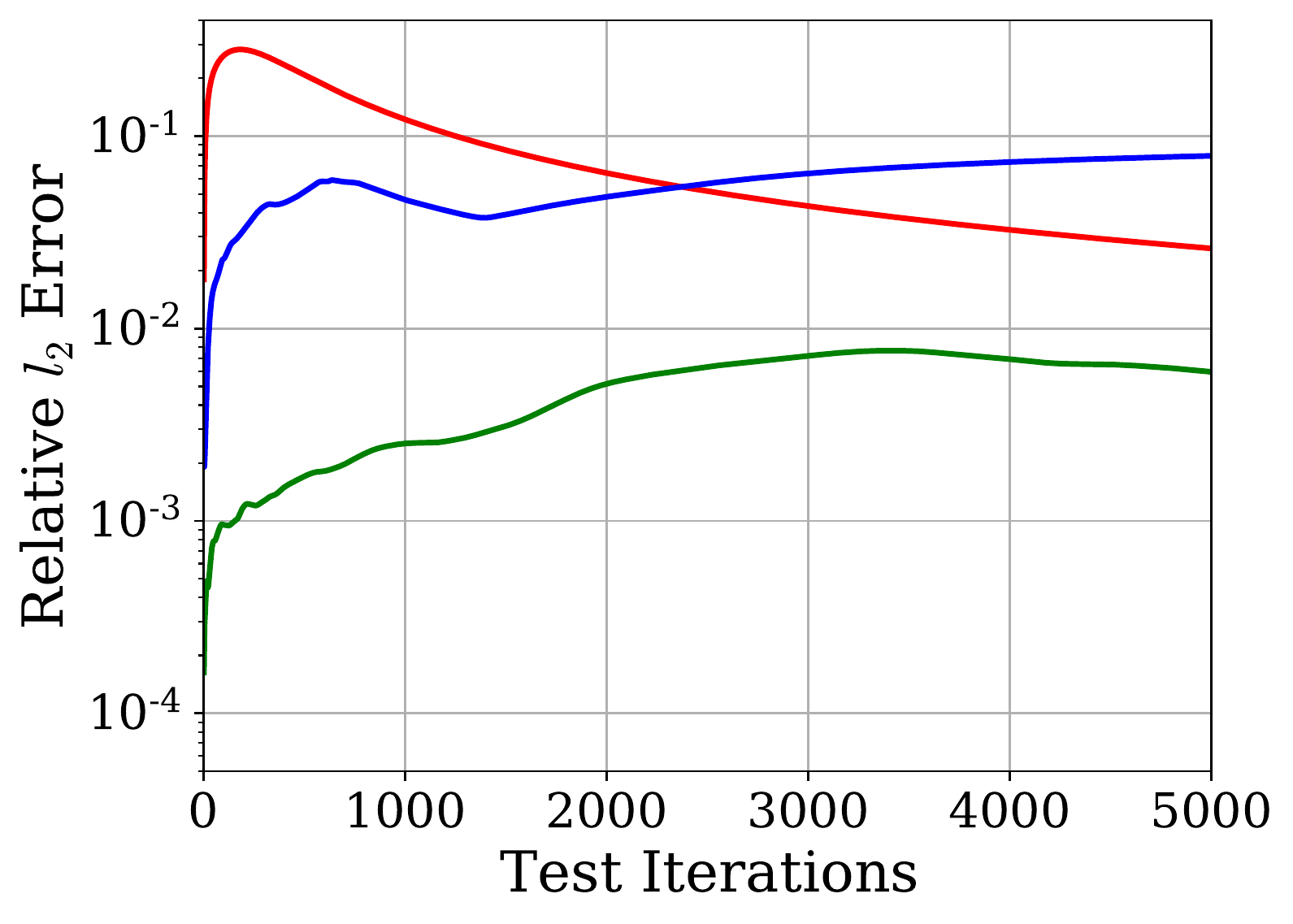}
		\caption{$\alpha = 1.5$, relative errors.}
		\label{subfig:s6_error_a1.5}
	\end{subfigure}%
	\\
	\begin{subfigure}[b]{0.5\linewidth}
		\centering 
		\includegraphics[width=\textwidth]{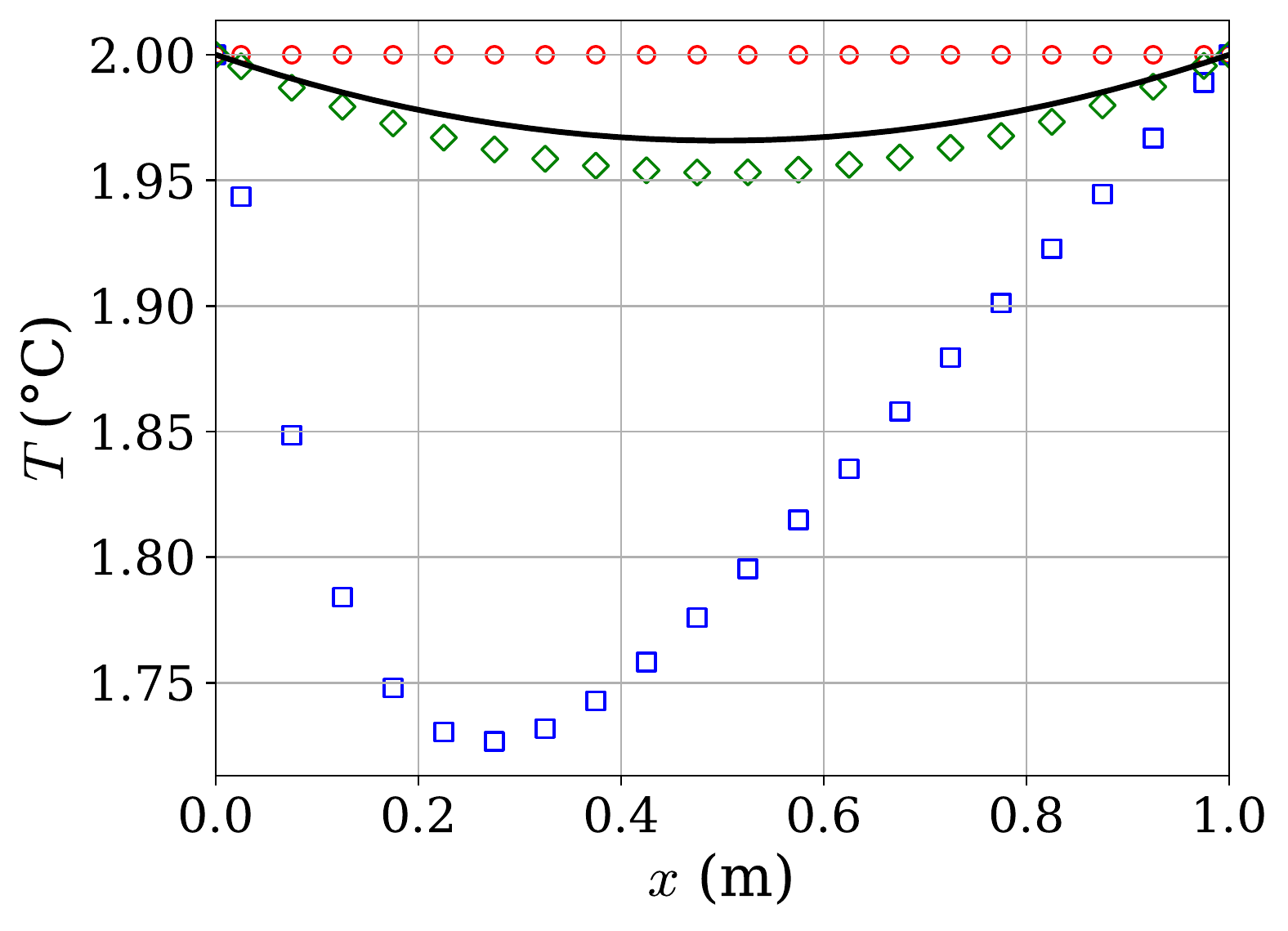}
		\caption{$\alpha = 0.7$, time level $n=5000$.}
		\label{subfig:s6_profile_a0.7_t2}
	\end{subfigure}%
	\begin{subfigure}[b]{0.5\linewidth}
		\centering 
		\includegraphics[width=\textwidth]{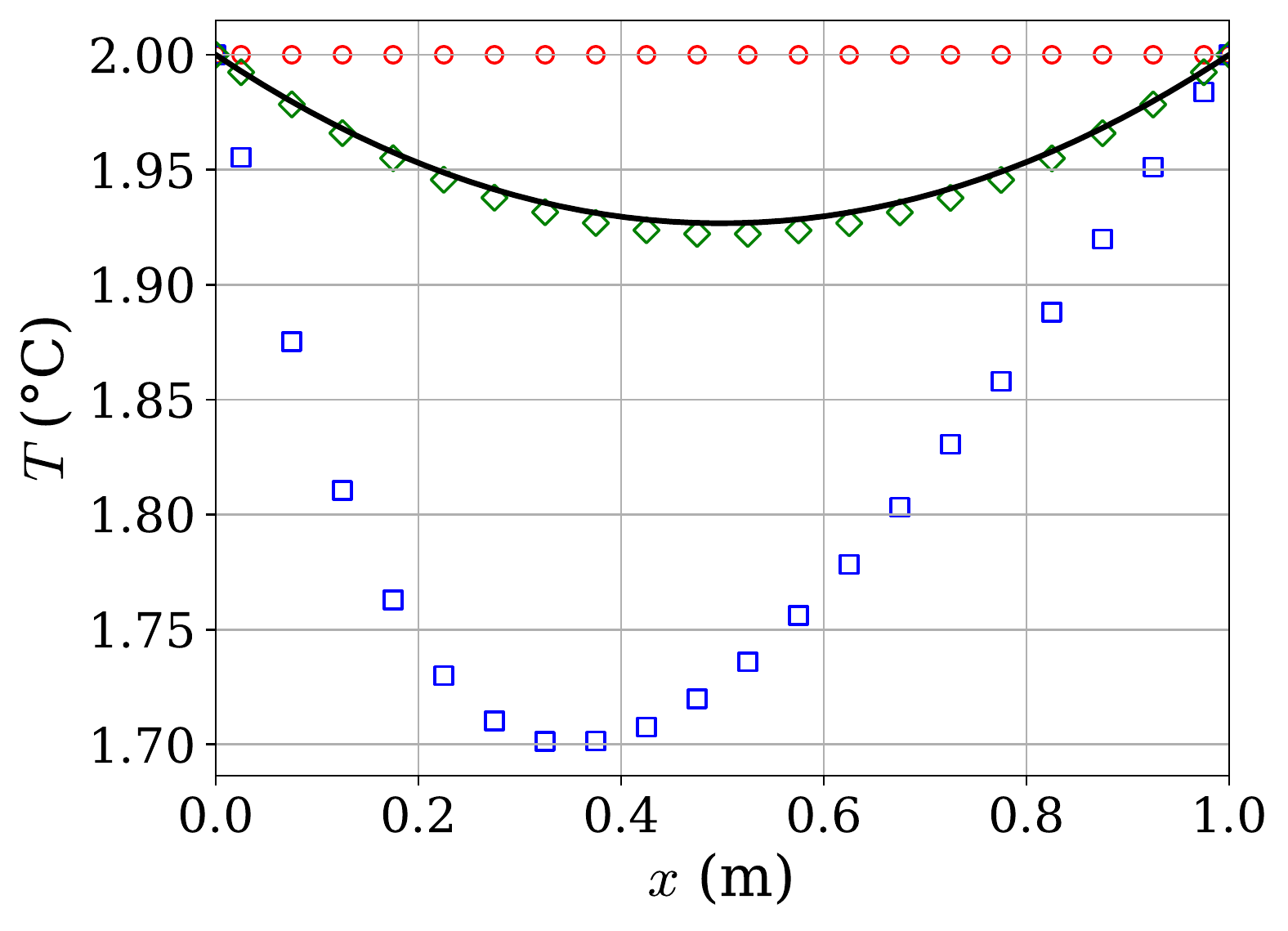}
		\caption{$\alpha = 1.5$, time level $n=5000$.}
		\label{subfig:s6_profile_a1.5_t2}
	\end{subfigure}
	\caption{Solution 3, interpolation: Comparison of relative errors and final temperature profiles for $\alpha=0.7,1.5$ (\blackline\ Exact, \raisebox{0.5pt}{\textcolor{red}{$\circ$}} PBM, \textcolor{blue}{$\square$} DDM, \raisebox{0.5pt}{\textcolor{green}{$\diamond$}} HAM). Only HAM is able to provide qualitatively correct predictions at the final time level.}
	\label{fig:s6_interp} 
\end{figure}

\begin{figure}
	\begin{subfigure}[b]{0.5\linewidth}
		\centering 
		\includegraphics[width=\textwidth]{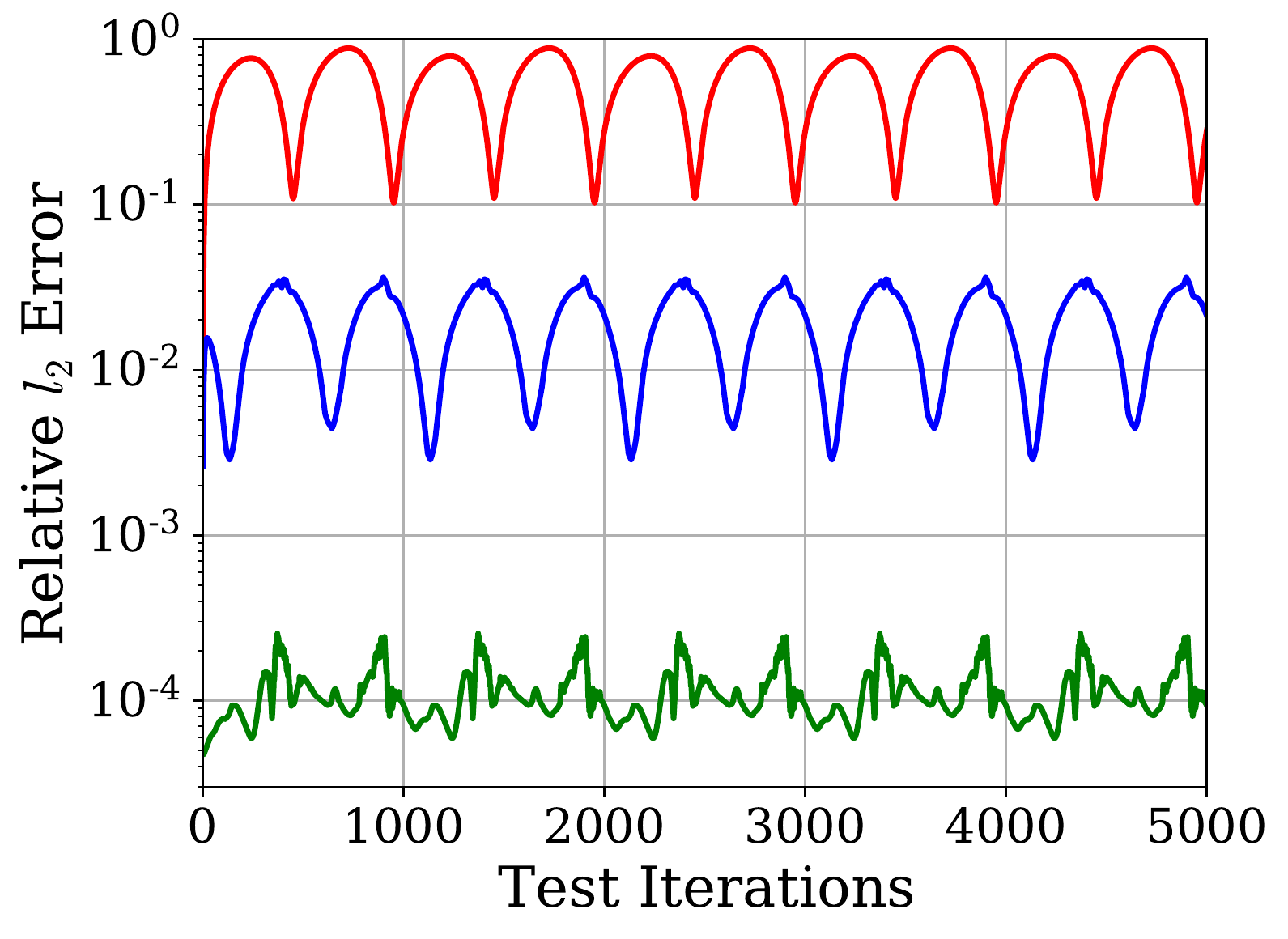}
		\caption{$\alpha = 0.7$, relative errors.}
		\label{subfig:s8_error_a0.7}
	\end{subfigure}%
	\begin{subfigure}[b]{0.5\linewidth}
		\centering 
		\includegraphics[width=\textwidth]{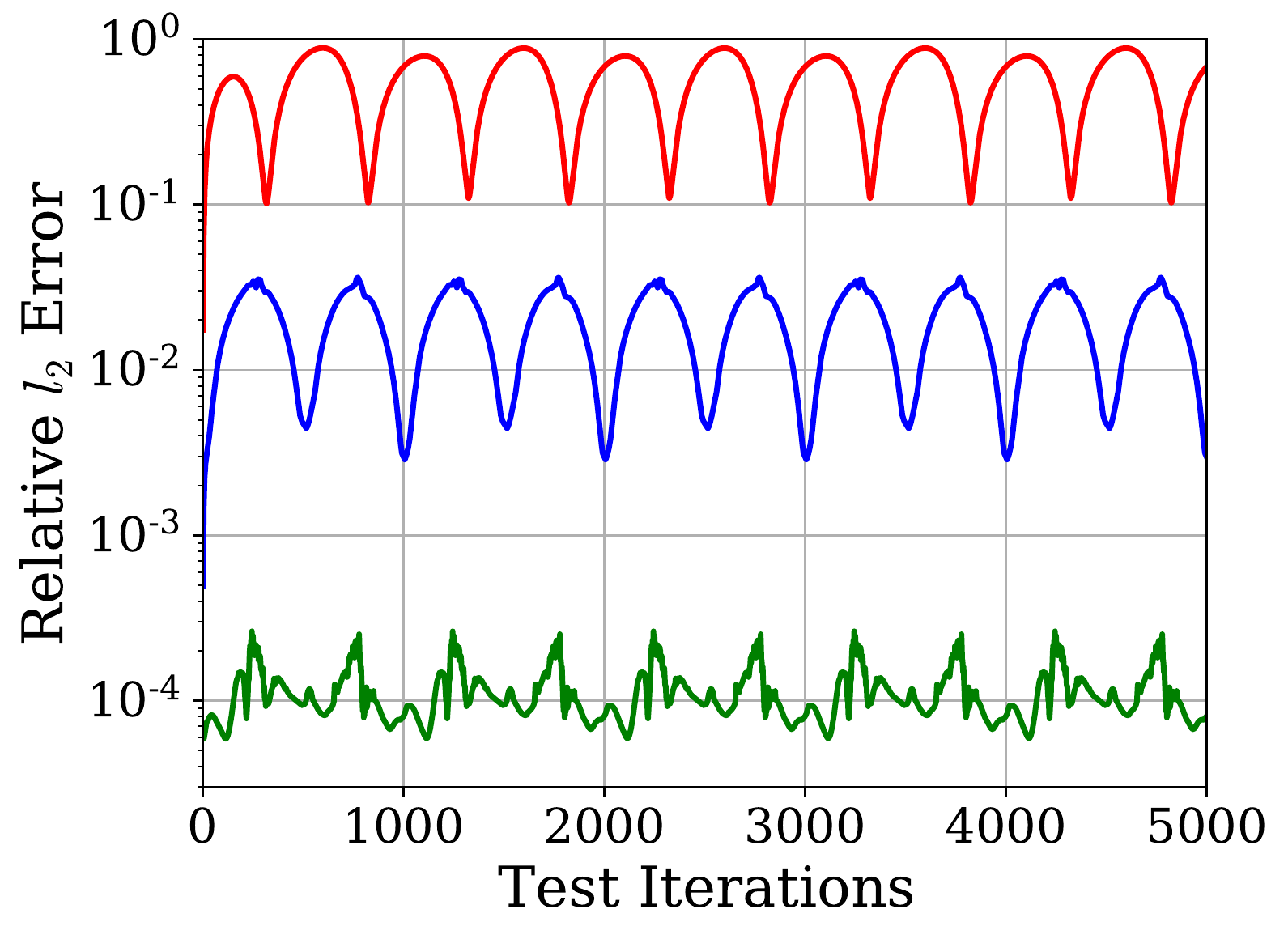}
		\caption{$\alpha = 1.5$, relative errors.}
		\label{subfig:s8_error_a1.5}
	\end{subfigure}%
	\\
	\begin{subfigure}[b]{0.5\linewidth}
		\centering 
		\includegraphics[width=\textwidth]{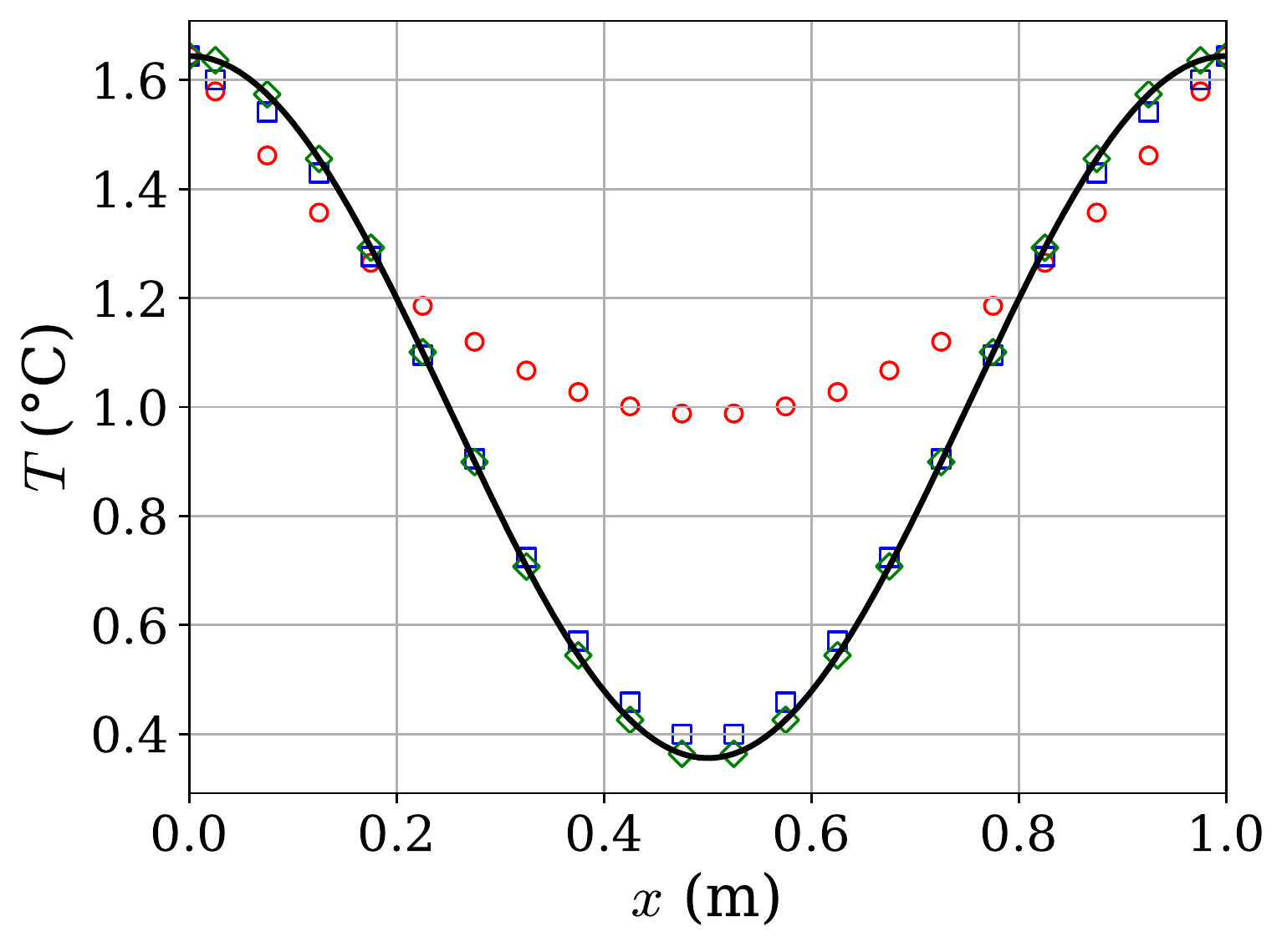}
		\caption{$\alpha = 0.7$, time level $n=5000$.}
		\label{subfig:s8_profile_a0.7_t2}
	\end{subfigure}%
	\begin{subfigure}[b]{0.5\linewidth}
		\centering 
		\includegraphics[width=\textwidth]{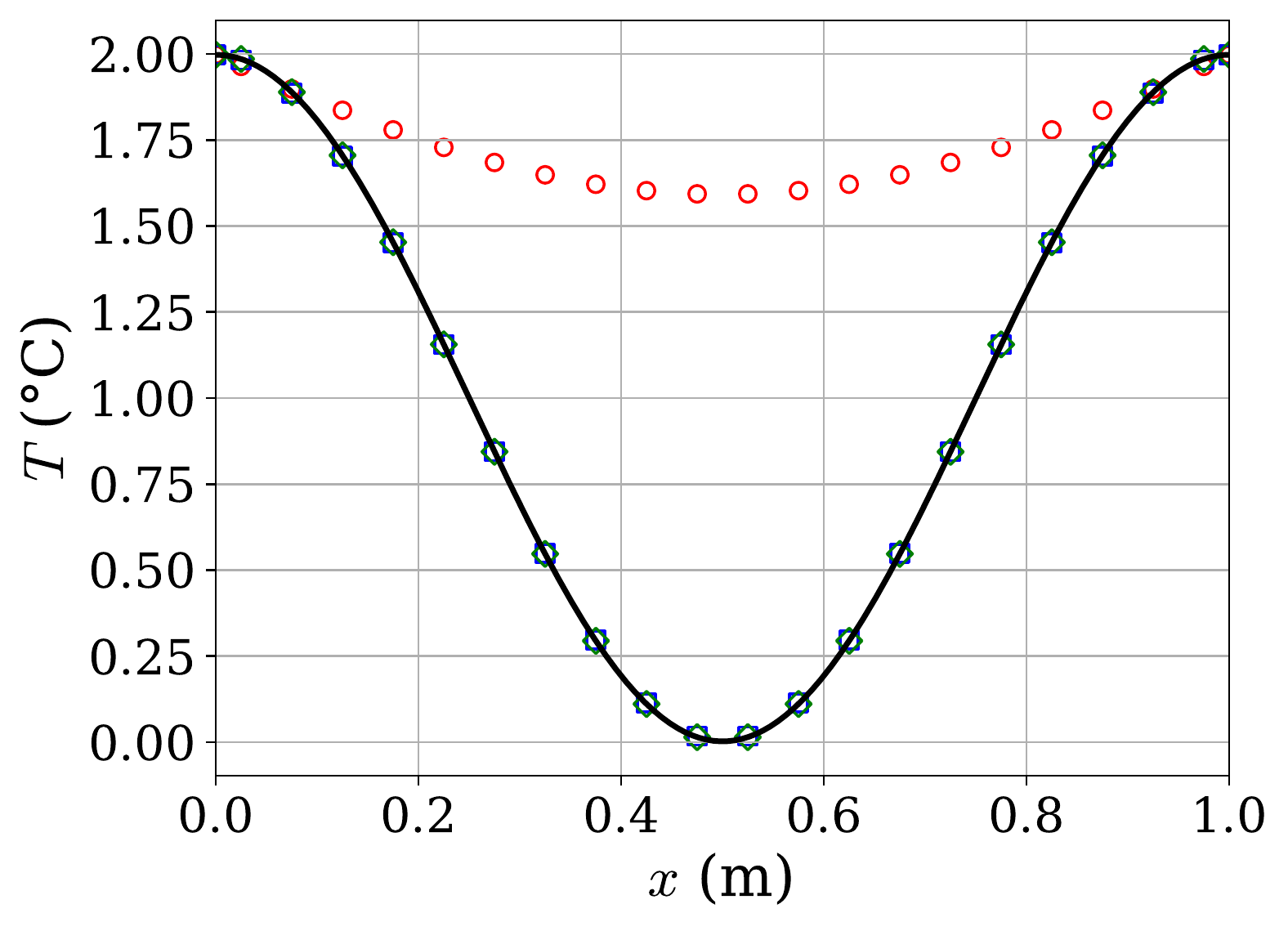}
		\caption{$\alpha = 1.5$, time level $n=5000$.}
		\label{subfig:s8_profile_a1.5_t2}
	\end{subfigure}
	\caption{Solution 4, interpolation: Comparison of relative errors and final temperature profiles for $\alpha=0.7,1.5$ (\blackline\ Exact, \raisebox{0.5pt}{\textcolor{red}{$\circ$}} PBM, \textcolor{blue}{$\square$} DDM, \raisebox{0.5pt}{\textcolor{green}{$\diamond$}} HAM). Both DDM and HAM provide reasonable predictions, while PBM does not. HAM is significantly more accurate than DDM.}
	\label{fig:s8_interp}
\end{figure}

As can be seen from the bottom halves of Figures~\ref{fig:s1_interp}--\ref{fig:s8_interp}, the predictions of CoSTA-based HAM are qualitatively correct for all interpolation scenarios. DDM also provides qualitatively correct predictions for Solutions~1,~2 and~4 (cf. Figures~\ref{fig:s1_interp}, \ref{fig:s2_interp} and~\ref{fig:s8_interp}), but its final-time-level predictions for Solution~3 deviates significantly from the reference profiles (cf. Figure~\ref{fig:s6_interp}). Furthermore, the relative errors of the HAM predictions are consistently more than one order of magnitude lower than those of the DDM predictions, as shown in the top halves of Figures~\ref{fig:s1_interp}--\ref{fig:s8_interp}; the difference in accuracy is particularly striking for Solution~2 (cf. Figure~\ref{fig:s2_interp}). Since HAM and DDM both utilize the same DNN setup and training routine, the significant performance difference can only be explained by the utilization of PBM inside HAM. As discussed previously, the combined use of a PBM and a DNN in CoSTA-based HAM allows some relevant physics to be accounted for by the PBM, thereby making the learning task of the DNN easier than in the DDM case where \emph{all} relevant physics must be learnt by the DNN. It is interesting to note that HAM clearly benefits from the PBM in spite of the PBM yielding low quality predictions on its own; the PBM predictions are qualitatively incorrect for Solutions~2 and~3 (cf. Figures~\ref{fig:s2_interp} and~\ref{fig:s6_interp}), and their errors are roughly one order of magnitude larger than the corresponding DDM errors for Solutions~1 and~4 (cf. Figures~\ref{fig:s1_interp} and~\ref{fig:s8_interp}). Another interesting observation is that the DDM and HAM error curves are less smooth than those of PBM, suggesting that the DNNs inject some pseudo-random noise into the DDM and HAM predictions. This noise could possibly be reduced by informing the DNNs of the temporal correlation between data at subsequent time levels through input (e.g. use two subsequent profiles as input, rather than just a single profile), using a different DNN architecture (e.g. LSTM architecture) and/or a different training regime (e.g. perform adversarial training with a temporal discriminator).

\subsubsection{Extrapolation scenarios}
\begin{figure}
	\begin{subfigure}[b]{0.5\linewidth}
		\centering 
		\includegraphics[width=\textwidth]{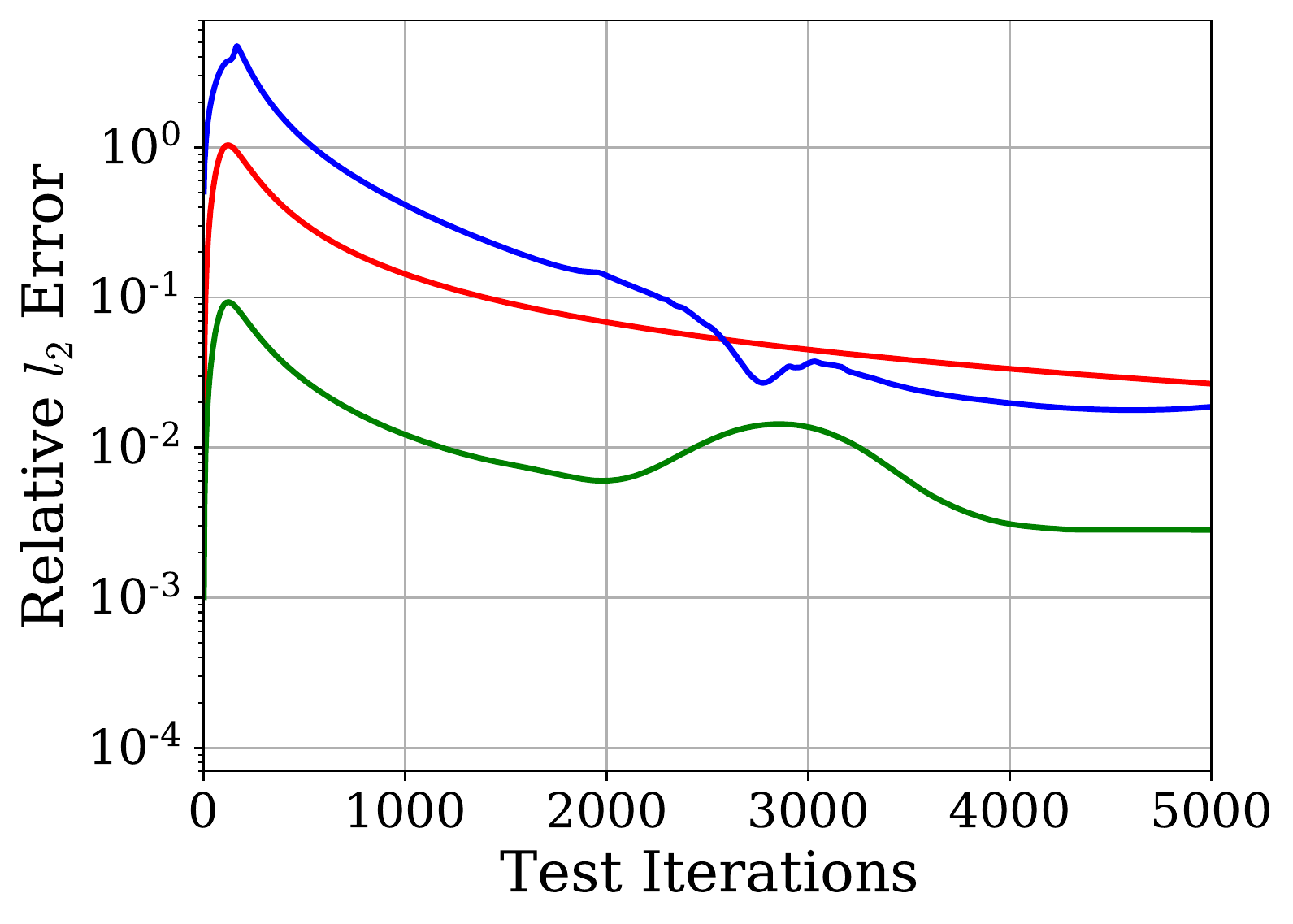}
		\caption{$\alpha = -0.5$, relative errors.}
		\label{subfig:s1_error_a-0.5}
	\end{subfigure}%
	\begin{subfigure}[b]{0.5\linewidth}
		\centering 
		\includegraphics[width=\textwidth]{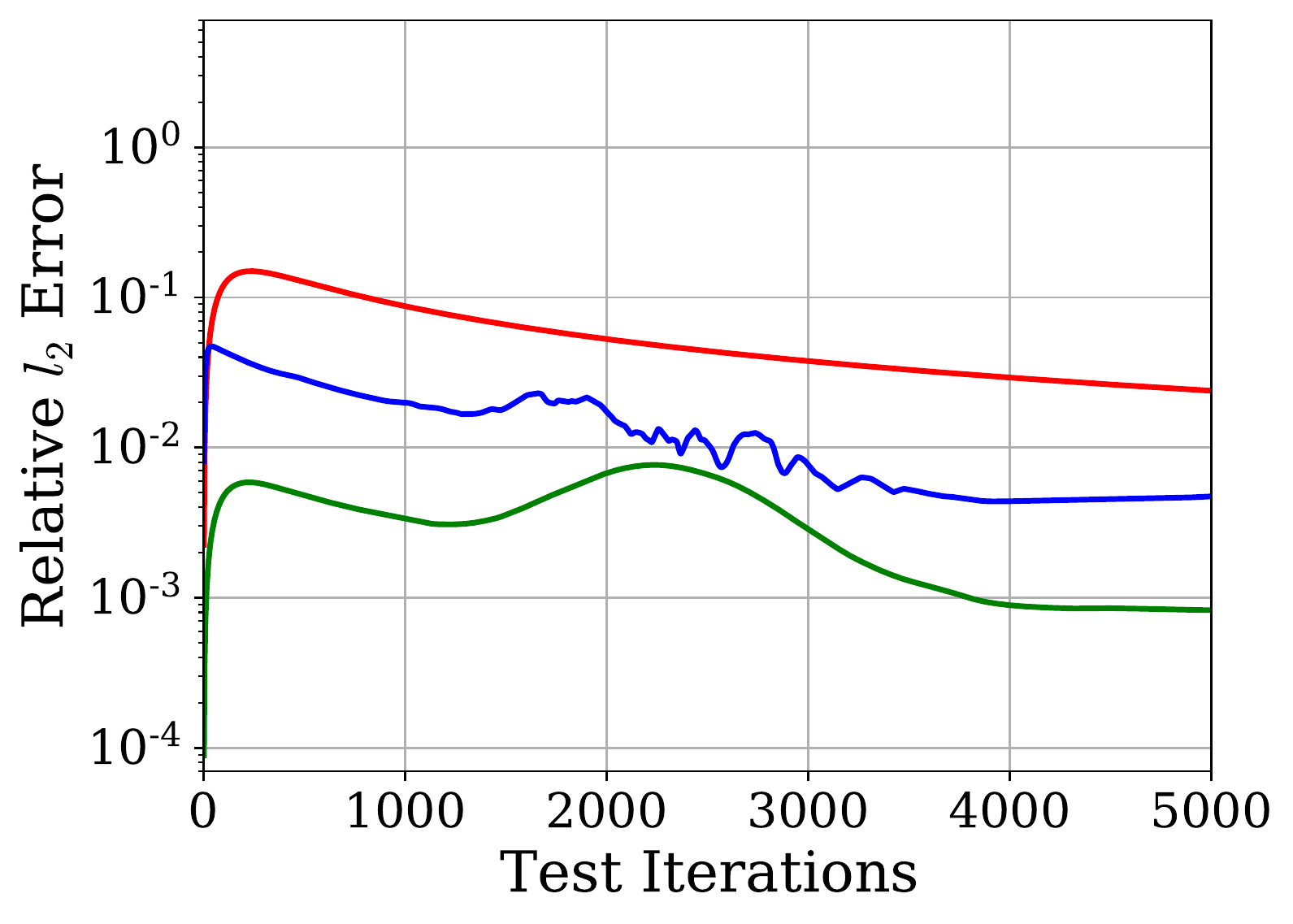}
		\caption{$\alpha = 2.5$, relative errors.}
		\label{subfig:s1_error_a2.5}
	\end{subfigure}%
	\\
	\begin{subfigure}[b]{0.5\linewidth}
		\centering 
		\includegraphics[width=\textwidth]{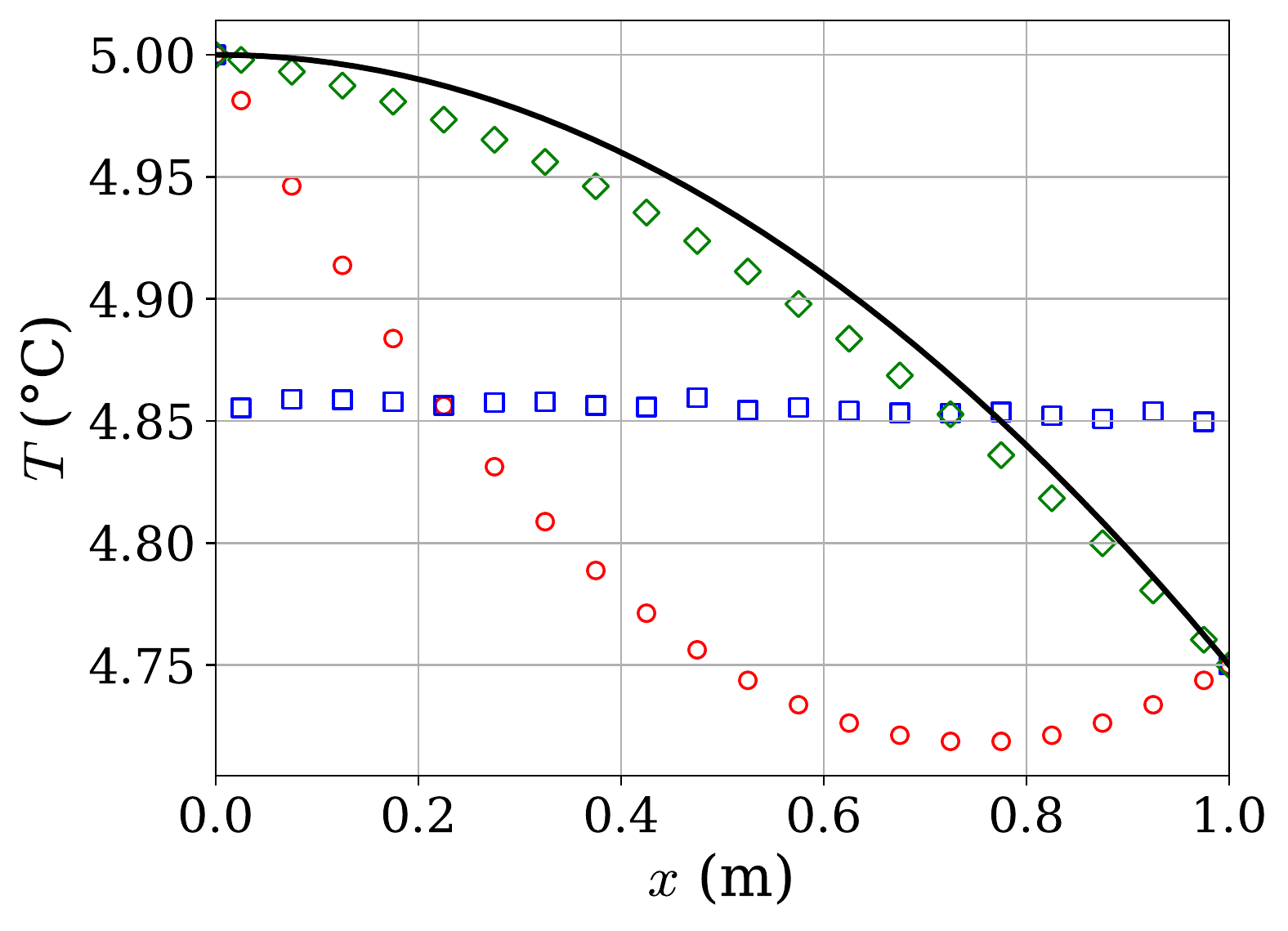}
		\caption{$\alpha = -0.5$, time level $n=5000$.}
		\label{subfig:s1_profile_a-0.5_t2}
	\end{subfigure}%
	\begin{subfigure}[b]{0.5\linewidth}
		\centering 
		\includegraphics[width=\textwidth]{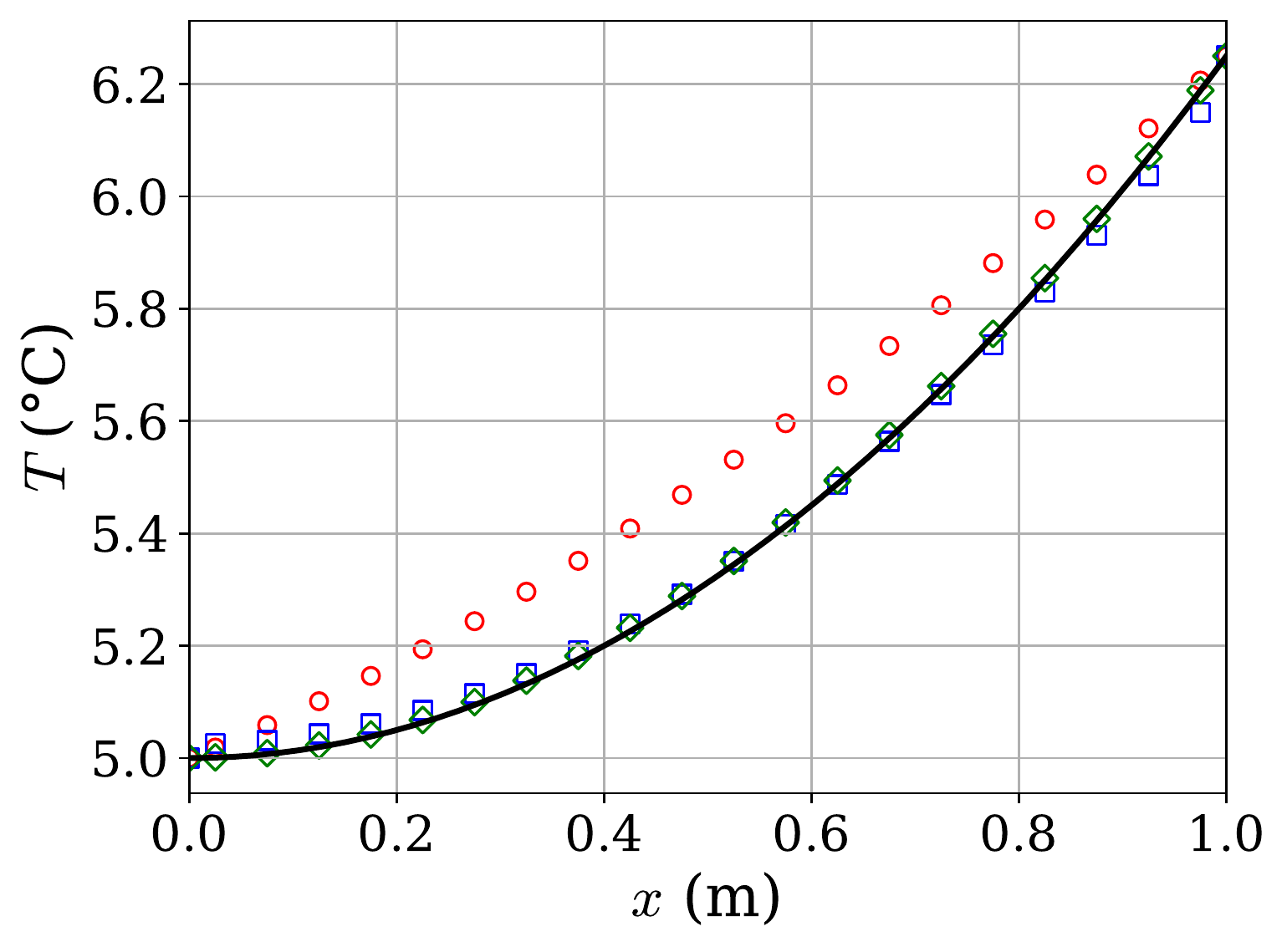}
		\caption{$\alpha = 2.5$, time level $n=5000$.}
		\label{subfig:s1_profile_a2.5_t2}
	\end{subfigure}
	\caption{Solution 1, extrapolation: Comparison of relative errors and final temperature profiles for $\alpha=-0.5,2.5$ (\blackline\ Exact, \raisebox{0.5pt}{\textcolor{red}{$\circ$}} PBM, \textcolor{blue}{$\square$} DDM, \raisebox{0.5pt}{\textcolor{green}{$\diamond$}} HAM). All models perform reasonably well for $\alpha=2.5$, but only HAM provides qualitatively correct predictions for $\alpha=-0.5$}
	\label{fig:s1_extrap} 
\end{figure}

\begin{figure}
	\begin{subfigure}[b]{0.5\linewidth}
		\centering 
		\includegraphics[width=\textwidth]{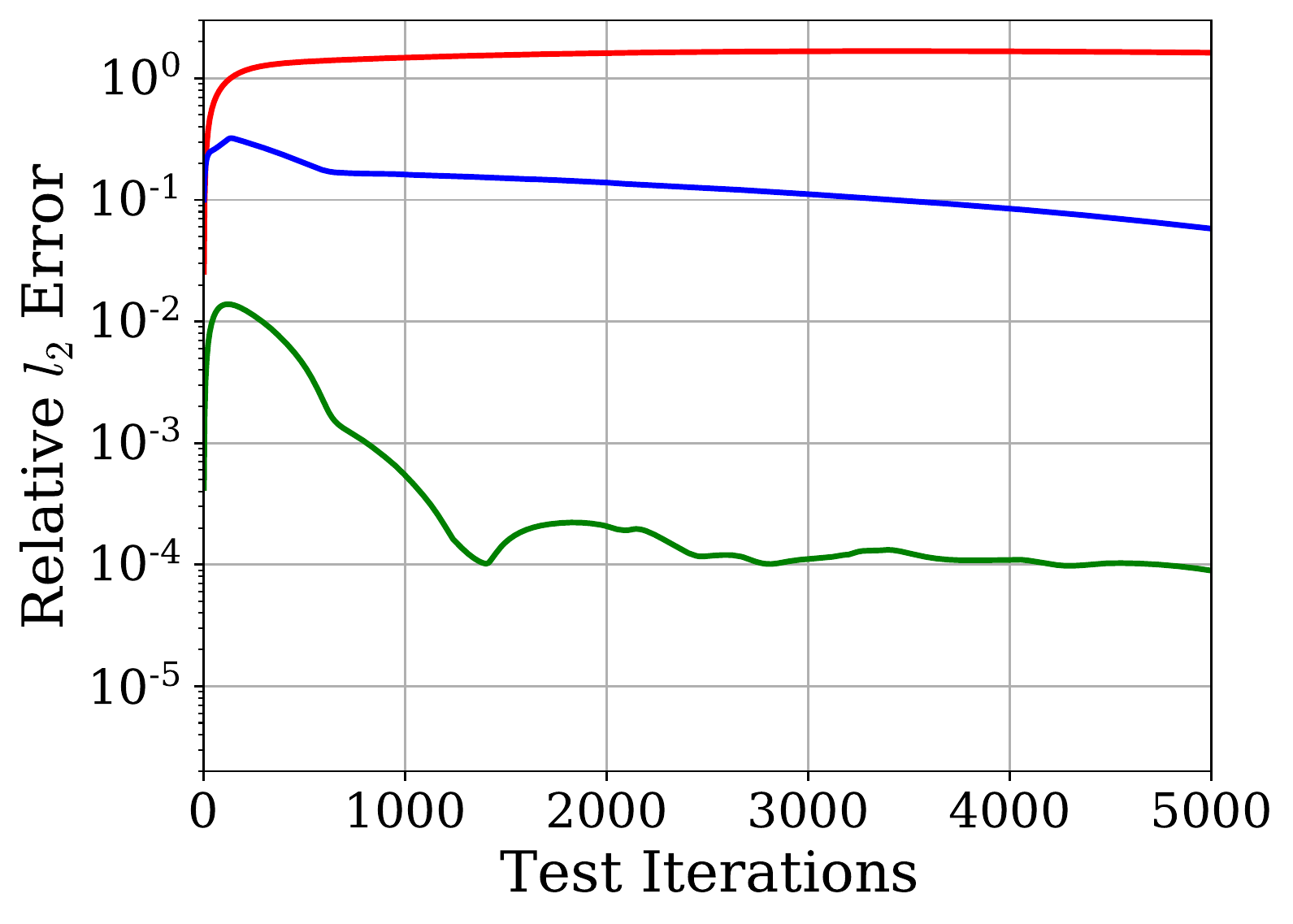}
		\caption{$\alpha = -0.5$, relative errors.}
		\label{subfig:s2_error_a-0.5}
	\end{subfigure}%
	\begin{subfigure}[b]{0.5\linewidth}
		\centering 
		\includegraphics[width=\textwidth]{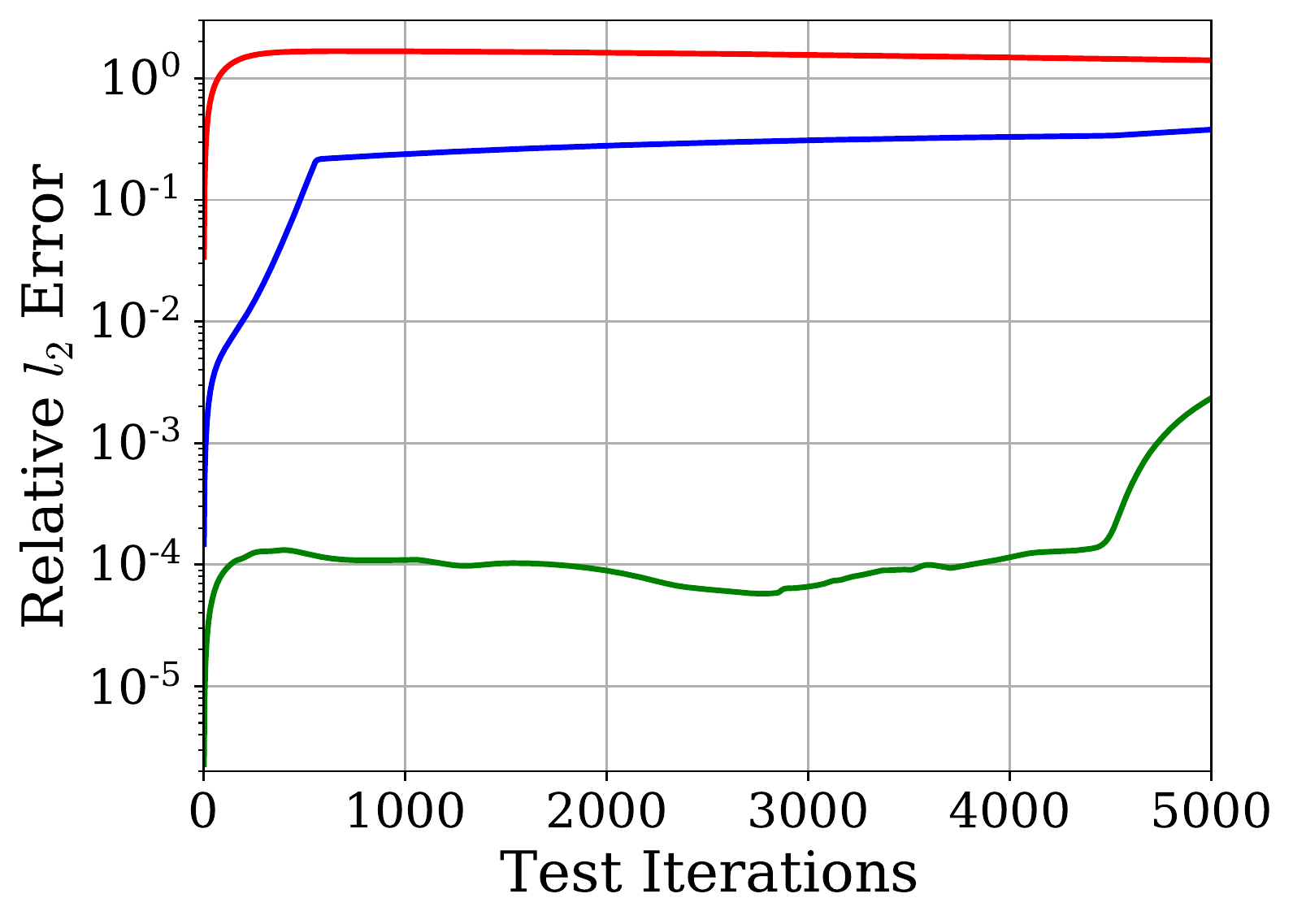}
		\caption{$\alpha = 2.5$, relative errors.}
		\label{subfig:s2_error_a2.5}
	\end{subfigure}%
	\\
	\begin{subfigure}[b]{0.5\linewidth}
		\centering 
		\includegraphics[width=\textwidth]{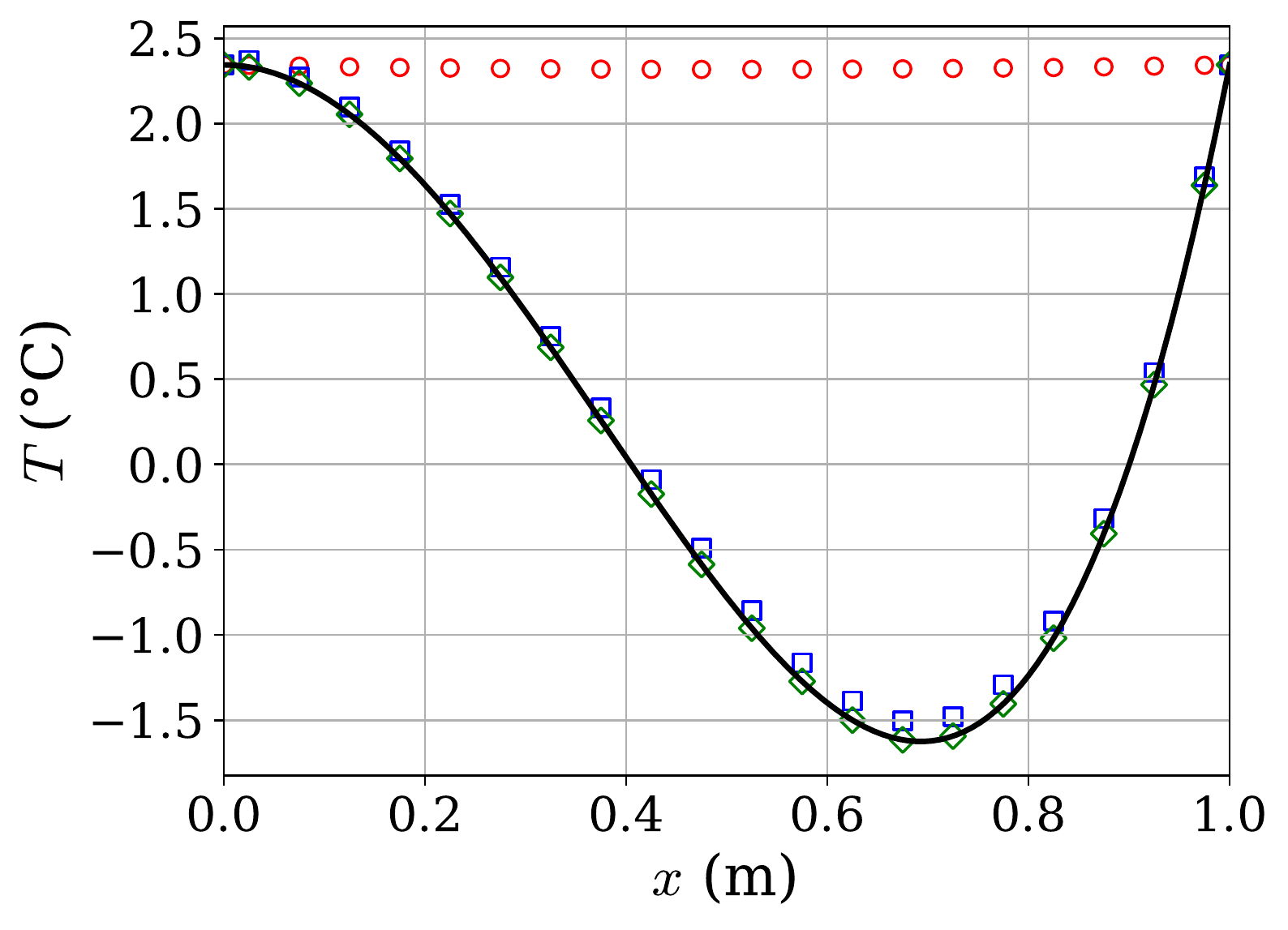}
		\caption{$\alpha = -0.5$, time level $n=5000$.}
		\label{subfig:s2_profile_a-0.5_t2}
	\end{subfigure}%
	\begin{subfigure}[b]{0.5\linewidth}
		\centering 
		\includegraphics[width=\textwidth]{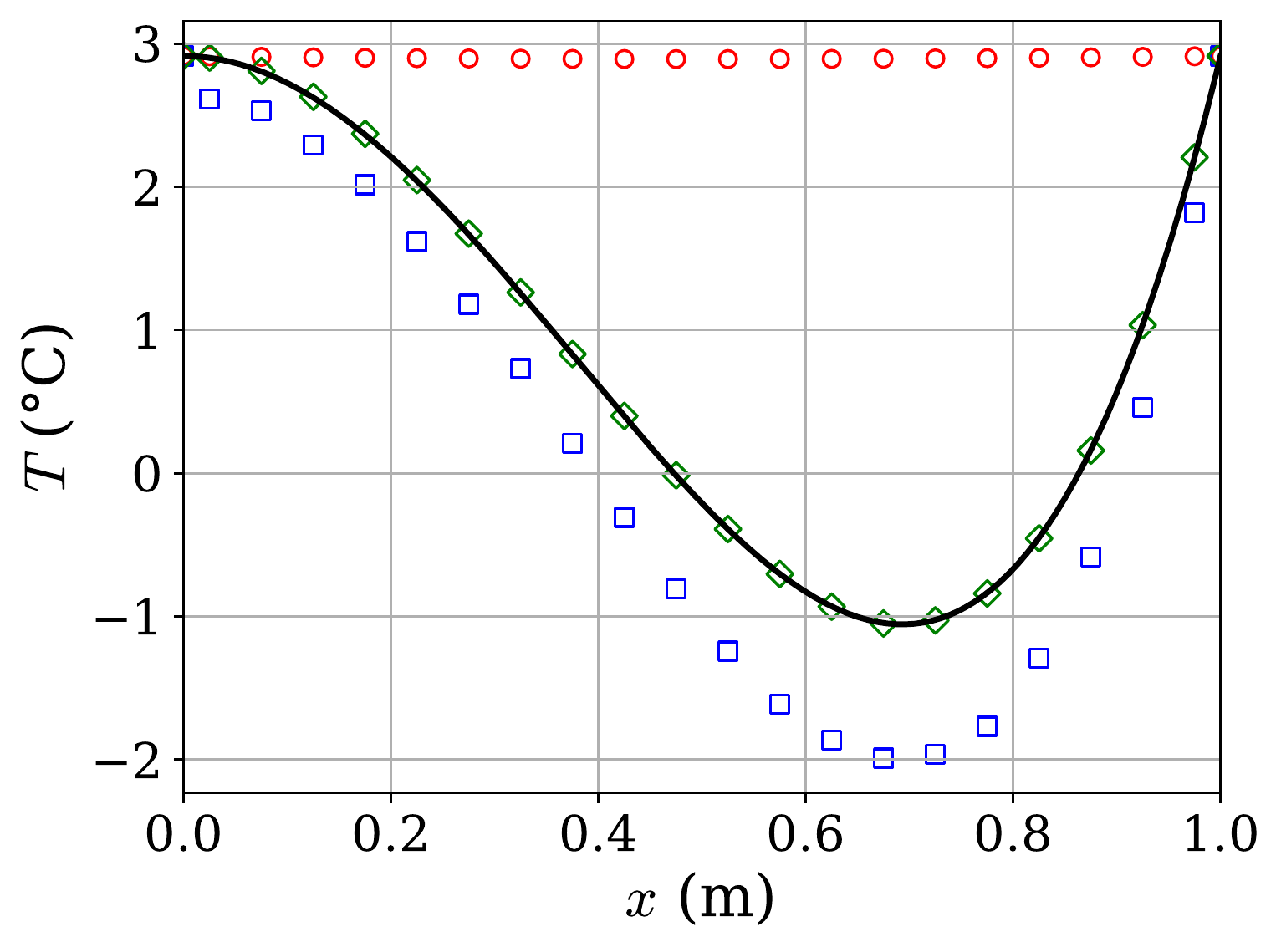}
		\caption{$\alpha = 2.5$, time level $n=5000$.}
		\label{subfig:s2_profile_a2.5_t2}
	\end{subfigure}
	\caption{Solution 2, extrapolation: Comparison of relative errors and final temperature profiles for $\alpha=-0.5,2.5$ (\blackline\ Exact, \raisebox{0.5pt}{\textcolor{red}{$\circ$}} PBM, \textcolor{blue}{$\square$} DDM, \raisebox{0.5pt}{\textcolor{green}{$\diamond$}} HAM). HAM's predictions are (mostly) three orders of magnitude more accurate than thoe of DDM. PBM's predictions are qualitatively incorrect.}
	\label{fig:s2_extrap} 
\end{figure}

\begin{figure}
	\begin{subfigure}[b]{0.5\linewidth}
		\centering 
		\includegraphics[width=\textwidth]{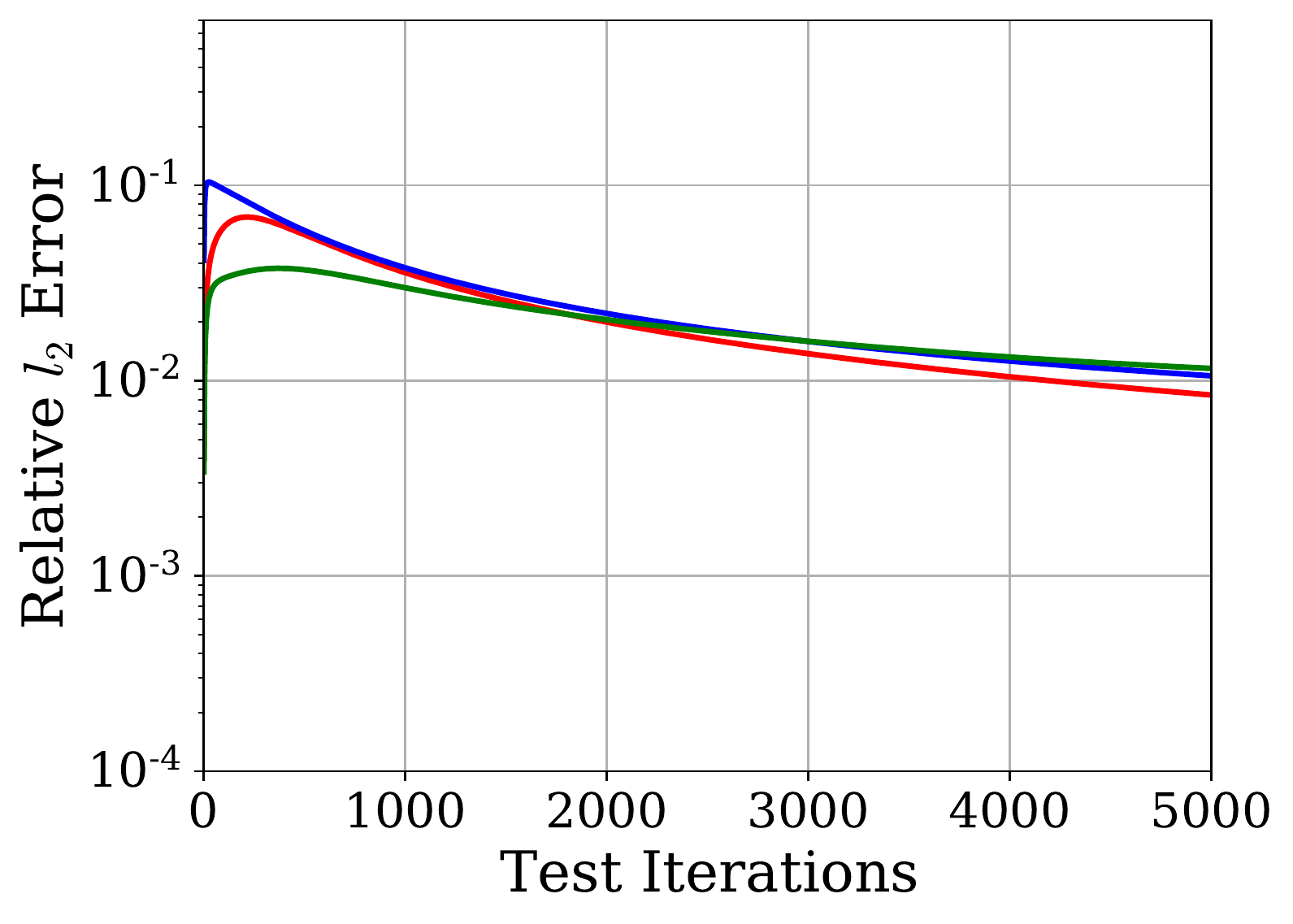}
		\caption{$\alpha = -0.5$, relative errors.}
		\label{subfig:s6_error_a-0.5}
	\end{subfigure}%
	\begin{subfigure}[b]{0.5\linewidth}
		\centering 
		\includegraphics[width=\textwidth]{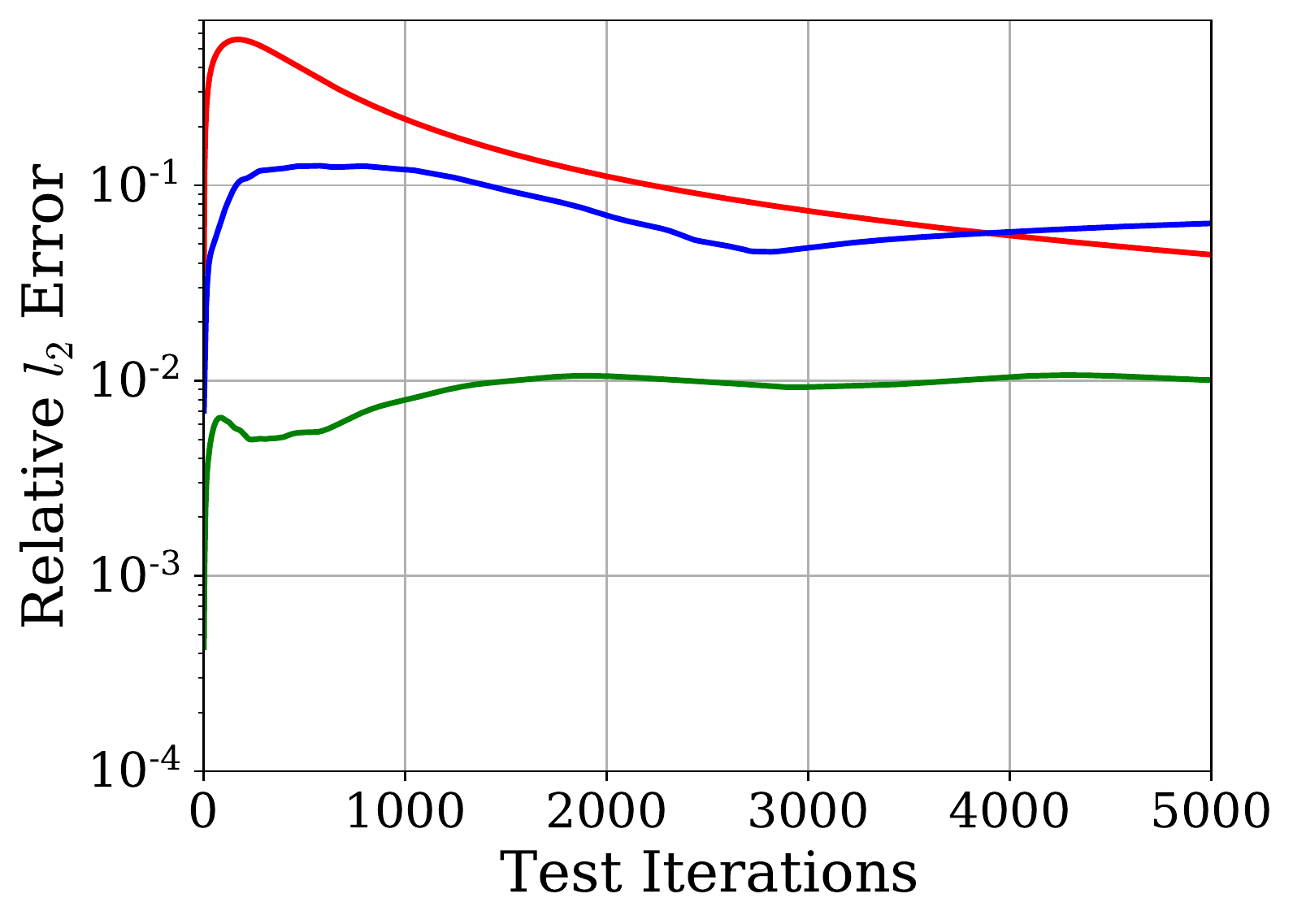}
		\caption{$\alpha = 2.5$, relative errors.}
		\label{subfig:s6_error_a2.5}
	\end{subfigure}%
	\\
	\begin{subfigure}[b]{0.5\linewidth}
		\centering 
		\includegraphics[width=\textwidth]{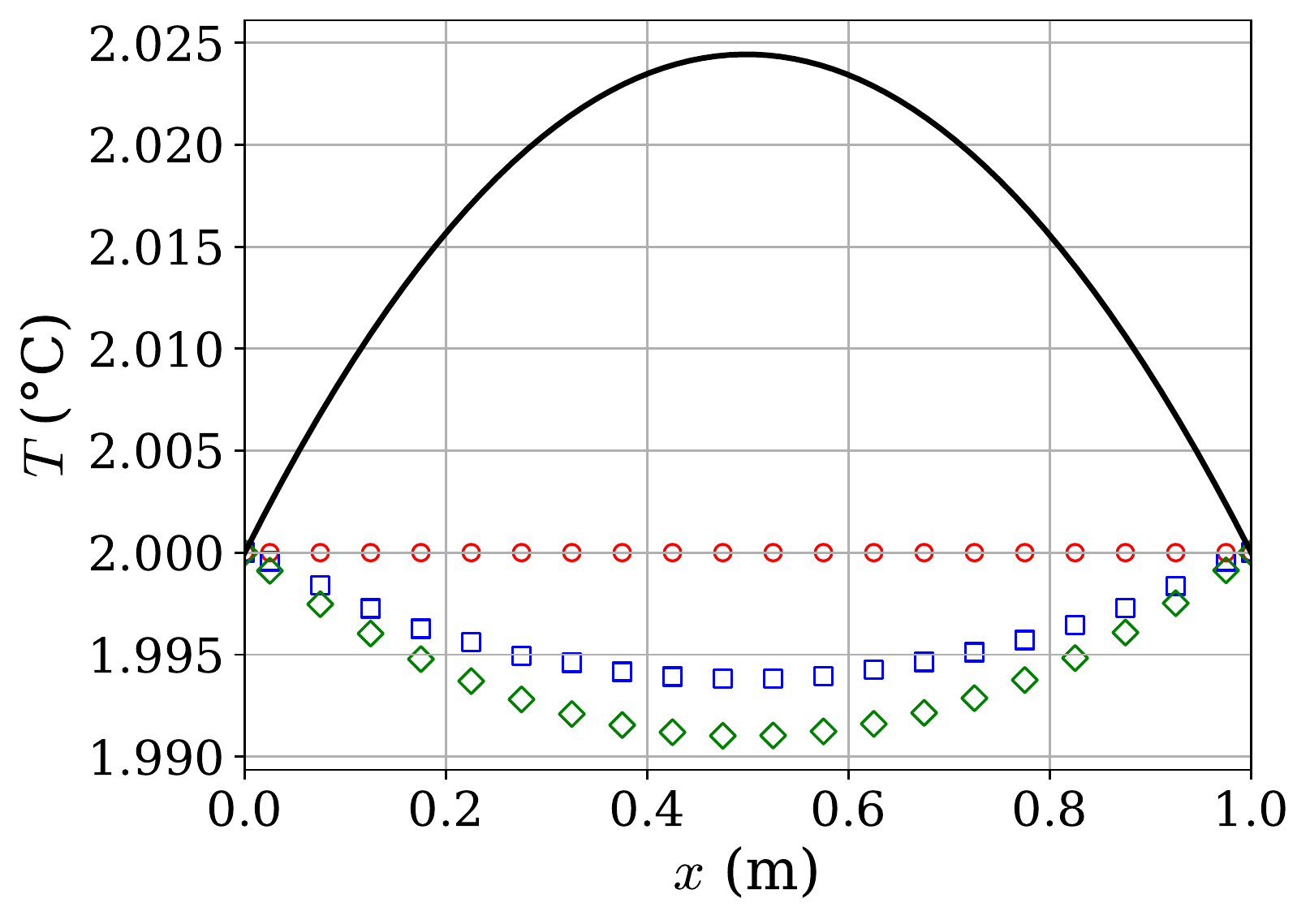}
		\caption{$\alpha = -0.5$, time level $n=5000$.}
		\label{subfig:s6_profile_a-0.5_t2}
	\end{subfigure}%
	\begin{subfigure}[b]{0.5\linewidth}
		\centering 
		\includegraphics[width=\textwidth]{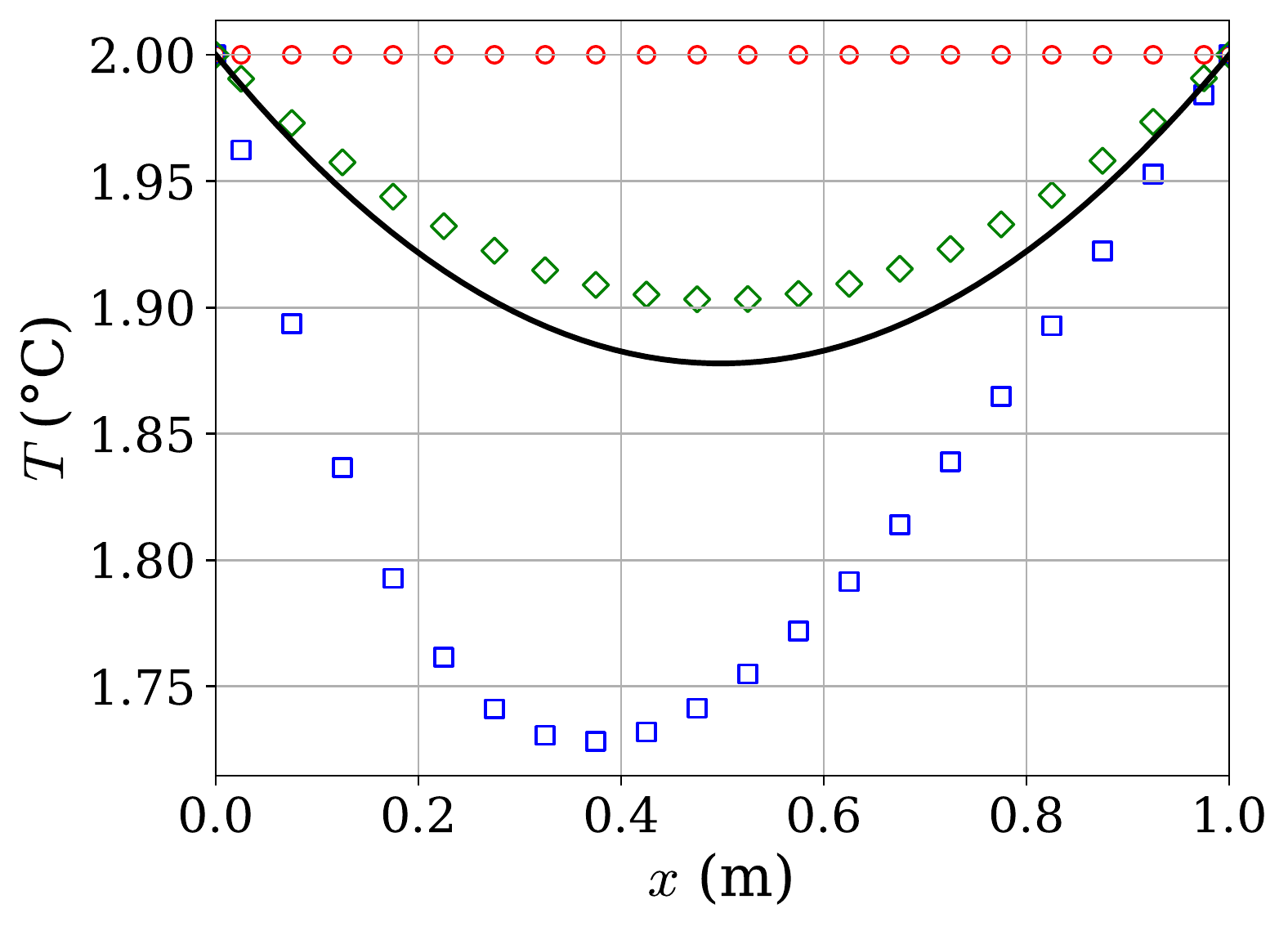}
		\caption{$\alpha = 2.5$, time level $n=5000$.}
		\label{subfig:s6_profile_a2.5_t2}
	\end{subfigure}
	\caption{Solution 3, extrapolation: Comparison of relative errors and final temperature profiles for $\alpha=-0.5,2.5$ (\blackline\ Exact, \raisebox{0.5pt}{\textcolor{red}{$\circ$}} PBM, \textcolor{blue}{$\square$} DDM, \raisebox{0.5pt}{\textcolor{green}{$\diamond$}} HAM). HAM is most accurate for $\alpha=2.5$, while all methods fail to provide qualitatively correct predictions for $\alpha=-0.5$.}
	\label{fig:s6_extrap} 
\end{figure}

\begin{figure}
	\begin{subfigure}[b]{0.5\linewidth}
		\centering 
		\includegraphics[width=\textwidth]{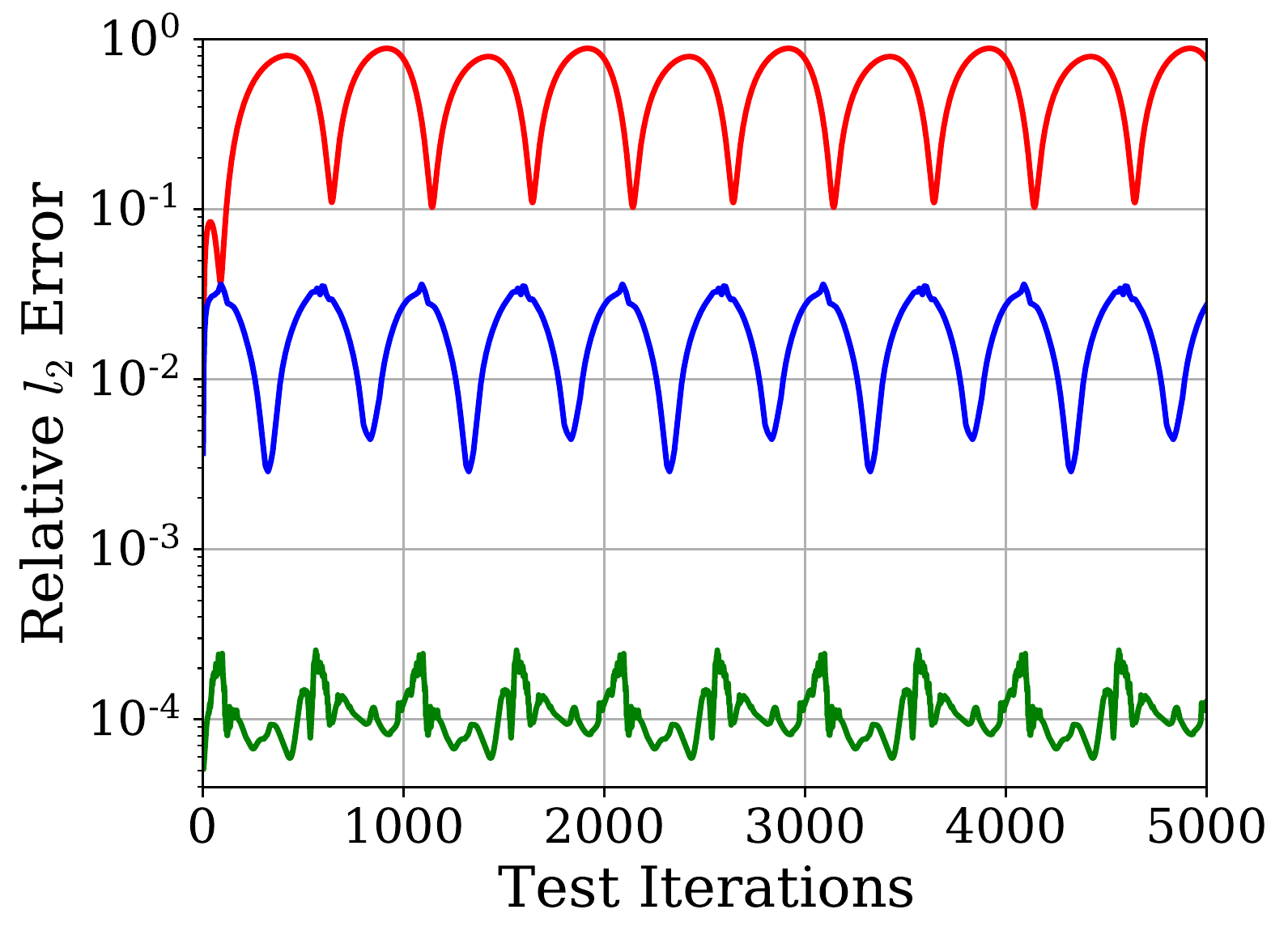}
		\caption{$\alpha = -0.5$, relative errors.}
		\label{subfig:s8_error_a-0.5}
	\end{subfigure}%
	\begin{subfigure}[b]{0.5\linewidth}
		\centering 
		\includegraphics[width=\textwidth]{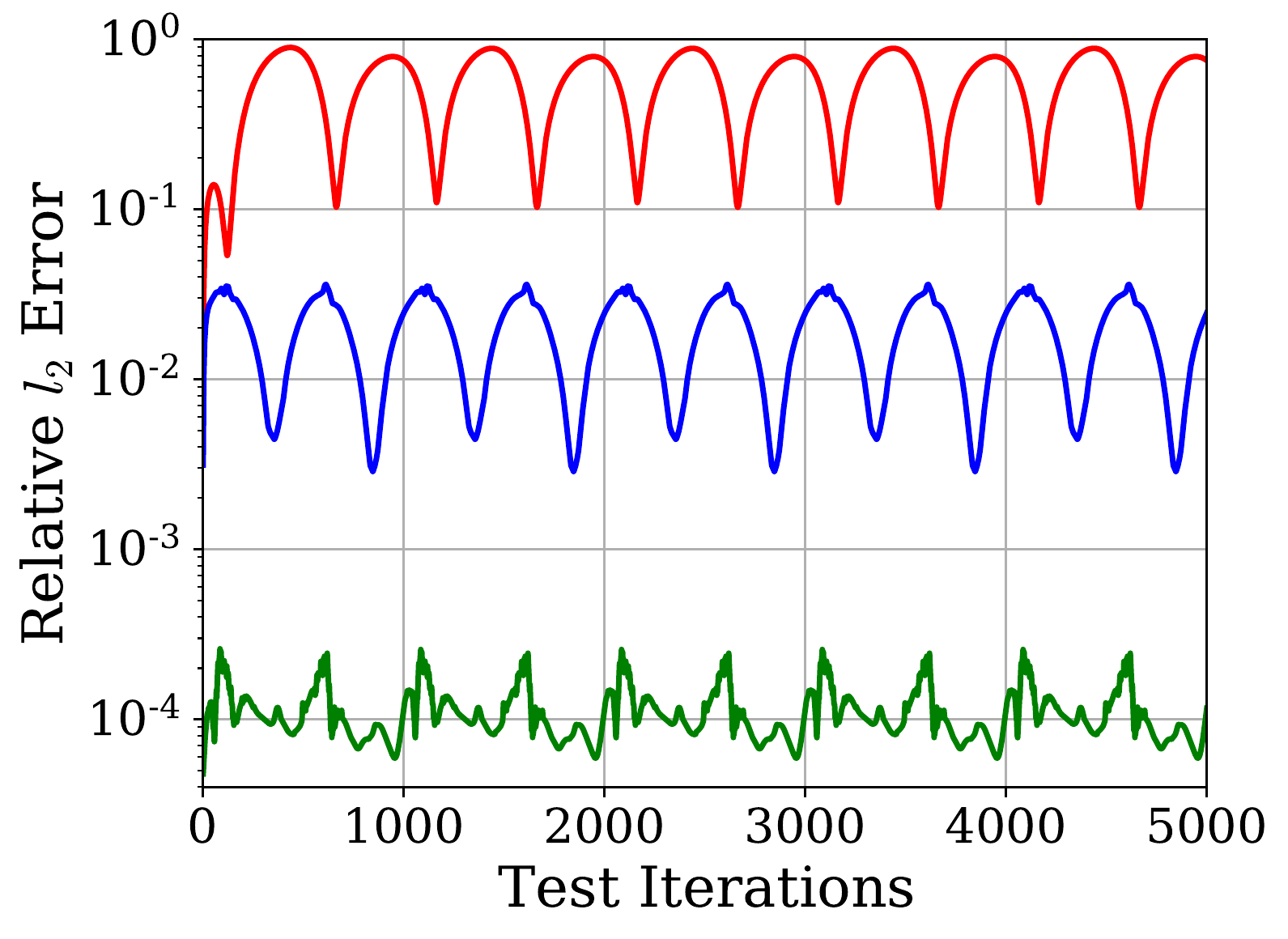}
		\caption{$\alpha = 2.5$, relative errors.}
		\label{subfig:s8_error_a2.5}
	\end{subfigure}%
	\\
	\begin{subfigure}[b]{0.5\linewidth}
		\centering 
		\includegraphics[width=\textwidth]{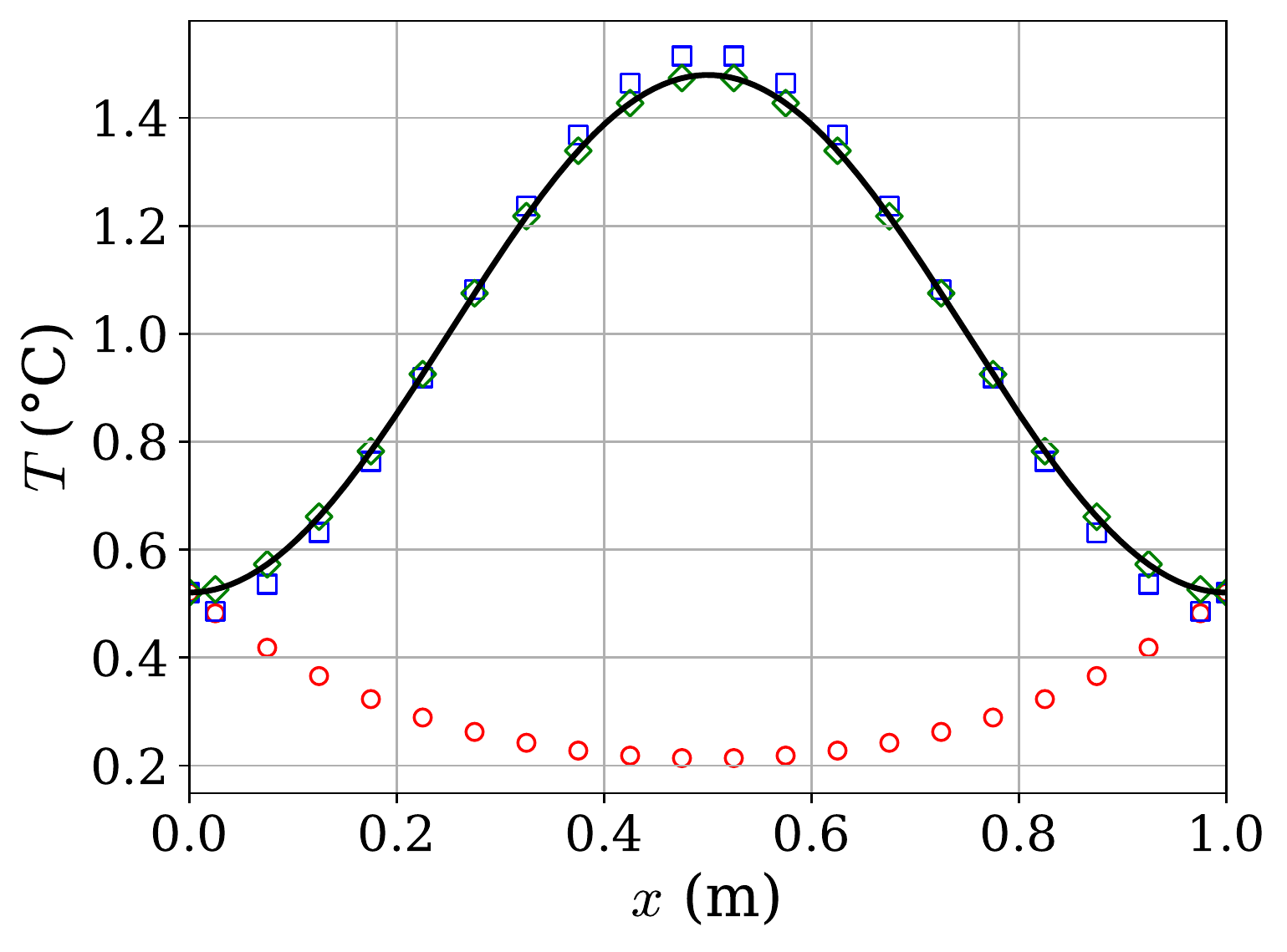}
		\caption{$\alpha = -0.5$, time level $n=5000$.}
		\label{subfig:s8_profile_a-0.5_t2}
	\end{subfigure}%
	\begin{subfigure}[b]{0.5\linewidth}
		\centering 
		\includegraphics[width=\textwidth]{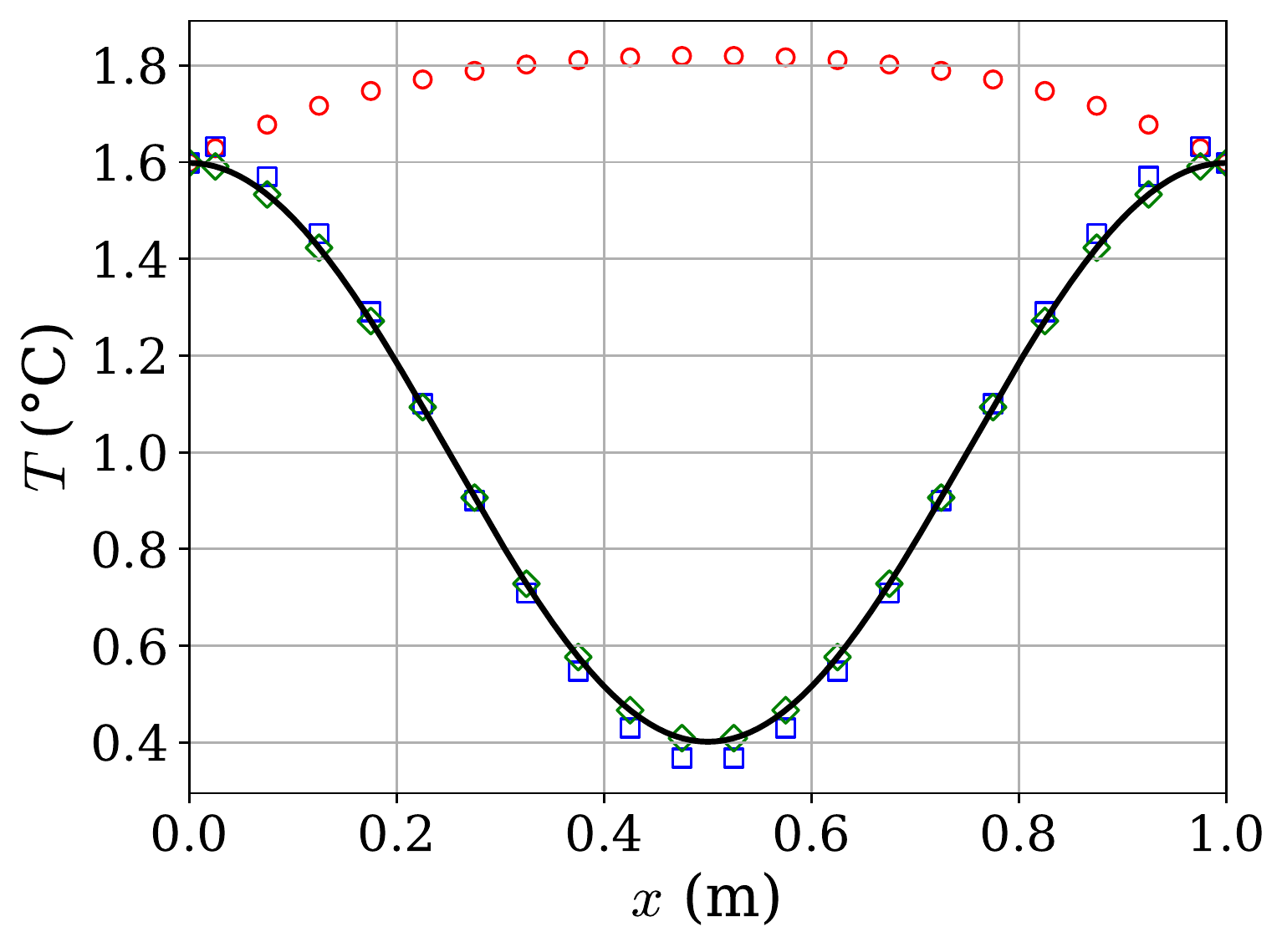}
		\caption{$\alpha = 2.5$, time level $n=5000$.}
		\label{subfig:s8_profile_a2.5_t2}
	\end{subfigure}
	\caption{System 4, extrapolation: Comparison of relative errors and final temperature profiles for $\alpha=-0.5,2.5$ (\blackline\ Exact, \raisebox{0.5pt}{\textcolor{red}{$\circ$}} PBM, \textcolor{blue}{$\square$} DDM, \raisebox{0.5pt}{\textcolor{green}{$\diamond$}} HAM). Both DDM and HAM provide qualitatively correct predictions, while PBM does not. HAM is significantly more accurate than DDM.}
	\label{fig:s8_extrap} 
\end{figure}

Owing to its well-known generalizability, the PBM performs similarly in the extrapolation scenarios as in the interpolation scenarios. However, as the accuracy of PBM was poor in the interpolation scenarios due to significant modeling error, the PBM results shown in Figures~\ref{fig:s1_extrap}--\ref{fig:s8_extrap} are also poor in general. DDM performs better than PBM, providing qualitatively correct predictions for Solutions~2 and~4 (cf. Figures~\ref{fig:s2_extrap} and~\ref{fig:s8_extrap}). The better performance of DDM in comparison to the PBM can be attributed to the fact that DDM, though generally poor in generalization, still learns about the nature of the unknown source term from the data. However, in terms of accuracy, HAM significantly outperforms DDM for both these manufactured solutions. HAM also outperforms DDM for Solution~1, where the DDM prediction for $\alpha=-0.5$ is qualitatively incorrect while the HAM prediction is not (cf. Figure~\ref{subfig:s1_profile_a-0.5_t2}). Furthermore, HAM also outperforms both PBM and DDM for Solution~3 with $\alpha=2.5$ (cf. Figure~\ref{subfig:s6_error_a2.5}). The remaining scenario -- Solution~3 with $\alpha=-0.5$ -- proves to be the most difficult scenario considered, as all three models fail to provide qualitatively correct predictions for this scenario (cf. Figure~\ref{subfig:s6_profile_a-0.5_t2}). Thus, while CoSTA-based HAM generally performs well in our experiments, the results for Solution~3 with $\alpha=-0.5$ show that significant improvements are still possible and desirable. In addition to ensuring temporal coherence, as previously discussed, utilizing problem-specific data augmentation techniques might be one way of boosting HAM's performance. Another possibility is to use a more accurate PBM as a basis for the full CoSTA model which will result in more relevant physics being accounted for by the PBM, thereby simplifying the learning task of the DNN. 

\section{Conclusion and future work}
\label{sec:conclusionandfuturework}
This paper has introduced the Corrective Source Term Approach (CoSTA) to hybrid analysis and modeling (HAM). The method utilizes a deep neural network (DNN) to generate a corrective source term that augments the discretized governing equation of a physics-based model (PBM). The purpose of the corrective source term is to account for physics that is unresolved by the PBM. In a series of numerical experiments on one-dimensional heat diffusion problems, we have compared the performance of CoSTA-based HAM with that of the uncorrected PBM and that of a comparable data-driven model (DDM). The most important conclusions from the study are as follows:

\begin{itemize}
    \item In the absence of any modeling error, i.e., when the full physics were known, the PBM was more accurate than the DDM. The PBM also generalized well to the extrapolation scenarios, while the DDM did not. On the other hand, HAM improved on the accuracy of the PBM in the interpolation scenarios, while still performing better or very close to the PBM in the extrapolation scenarios. The accuracy improvements of HAM can be attributed to its ability to compensate for the numerical discretization error. This ability can be exploited to accelerate numerical solvers by enabling the use of coarse meshes while retaining accuracy, as unresolved subgrid-scale physics can be accounted for by the corrective source term. This feature of HAM will be extremely useful in digital twin applications as accurate and real-time performance of the models is of utmost importance.  
    \item In the presence of modeling error i.e., when some relevant physics was unknown, 
    the PBM failed to make reasonable predictions. The DDM generally outperformed the PBM for both interpolation and extrapolation scenarios, since the DDM could learn about the unknown physics from data observations. However, HAM consistently performed better than both PBM and DDM, often decreasing the relative error of the predictions by several orders of magnitude. The improvement in predictions compared to DDM was even more pronounced for extrapolation scenarios; in other words, HAM was found to be more generalizable to new, unseen scenarios. In the context of digital twin applications, it is foreseen that new scenarios never witnessed before could occur from time to time, and HAM will be able to deal with such situation better than PBM and DDM.
\end{itemize}

The work presented herein clearly illustrates the great potential that can be unlocked by combining PBM and DDM in a HAM framework like CoSTA. While we have only considered a single PBM and a single DNN architecture for 
modelling a single class of physical systems,
CoSTA can easily be used with other PBMs and DNNs to
model other systems as well -- including multidimensional systems and systems with multiple governing equations -- due its solid but flexible theoretical foundation. Thus, the numerous benefits of employing a HAM framework like CoSTA, which we identify below, can greatly increase the applicability of data-driven techniques in a variety of applications, including digital twins and also high-stakes engineering applications currently reserved for pure PBM. Notable examples include improving the dynamic model of a system (like a vehicle or a process), improving the accuracy of building energy simulation models using regular measurements, modelling complex physical phenomena like turbulence whose full physical understanding is still limited, accelerating high fidelity numerical solvers, and stabilizing ROMs based on significant dimensionality reduction.

One important benefit is that HAM aims to utilize existing knowledge to the greatest extent possible, such that, for any given problem, HAM requires less complex data-driven techniques than pure DDM. This can improve the predictability and interpretability of the DDM, thereby making the model more trustworthy. Another related benefit is that the DNN-generated corrective source term of CoSTA can be explained and analyzed within the framework of a PBM based on known first principles. This allows for an inbuilt sanity check for the data-driven part of the HAM model. For example, in the case of heat diffusion modeling, if there exists known bounds for the energy transferred to a system, then the DNN-generated source term must also be bounded, and any violation of these bounds can be automatically detected. The lack of such automatic DNN-misbehaviour detection in pure DDM is one of the reasons why the black-box nature of DNNs has inhibited the acceptance of data-driven techniques in high-stakes applications. As for digital twins, CoSTA facilitates the use of coarse meshes in numerical solvers through its ability to correct both modeling errors and discretization errors, thereby improving computational efficiency while retaining accuracy. Furthermore, since the learning of the source term can continue even after model deployment, CoSTA allows for any new phenomena encountered by the digital twin to be learned automatically. Thus, CoSTA has the potential to develop models which will be generalizable, trustworthy, accurate and computationally efficient, and self-evolving.
However, while our results are promising, more research is required with regards to the accuracy, stability and interpretability of the learned source term, such as to ensure that the findings of this paper generalize to other physical systems and processes. Making optimal choices for PBMs, as well as for DNN architectures and training regimes, are other interesting research directions. Some of the outlined improvements and extensions will be the topic of future research by the authors, but we hope that this paper will inspire other scientists to pursue the outlined research directions as well.

\section*{Acknowledgments}
The second and third author are grateful for the support received by the Research Council of Norway and the industrial partners of the following projects: EXAIGON--{\em Explainable AI systems for gradual industry adoption\/} (grant no. 304843), {\em Hole cleaning monitoring in drilling with distributed sensors and hybrid methods\/} (grant no. 308823), and RaPiD--{\em Reciprocal Physics and Data-driven models\/} (grant no. 313909). The fourth author gratefully acknowledges the Early Career Research Program (ECRP) support of the U.S. Department of Energy, Office of Science, Office of Advanced Scientific Computing Research under Award Number DE-SC0019290.

\bibliography{references}
\end{document}